%% file: main.tex
\documentclass[10pt,twocolumn]{article}
\usepackage[margin=0.75in,columnsep=0.25in,headheight=26pt,headsep=8pt]{geometry}
\usepackage[T1]{fontenc}
\usepackage{amsmath,amssymb,amsthm,mathtools}
\usepackage{mathpazo}           % Palatino text and math, with sans-serif headings, as in LeJEPA
\usepackage{tgpagella,tgheros,microtype}
\usepackage{graphicx,booktabs}
\usepackage[numbers,sort&compress]{natbib}
\usepackage{xcolor}
\usepackage{tikz}
\usepackage{tcolorbox}
\tcbuselibrary{skins,breakable}
\usepackage{titlesec,fancyhdr}
\usepackage[font=footnotesize,labelfont=bf,labelsep=period]{caption}
\usetikzlibrary{arrows.meta,decorations.pathreplacing,calc,positioning,shapes.geometric}
\usepackage[colorlinks=true,linkcolor=blue!60!black,citecolor=blue!60!black,urlcolor=blue!60!black]{hyperref}
\usepackage[capitalize,noabbrev]{cleveref}

\definecolor{good}{HTML}{1F77B4}\definecolor{bad}{HTML}{D62728}\definecolor{ntp}{HTML}{FF7F0E}\definecolor{aligned}{HTML}{2CA02C}
\definecolor{rule}{HTML}{8EB9D9}

\titleformat{\section}{\sffamily\bfseries\Large\raggedright}{\thesection}{0.6em}{}
\titleformat{\subsection}{\sffamily\bfseries\large\raggedright}{\thesubsection}{0.6em}{}
\titlespacing*{\section}{0pt}{10pt plus 3pt minus 2pt}{5pt plus 1pt}
\titlespacing*{\subsection}{0pt}{8pt plus 2pt minus 2pt}{4pt plus 1pt}
\newif\ifappendixpart
\newcommand{\roadmap}[3]{\hyperref[#3]{\ifnum\value{section}=#1\textbf{Sec #1: #2}\else Sec #1: #2\fi}}
\fancypagestyle{plain}{\fancyhf{}\fancyfoot[C]{\thepage}}

\tcbset{result/.style={enhanced,breakable,arc=2pt,boxrule=0.6pt,left=4pt,right=4pt,top=2pt,bottom=3pt,toptitle=1pt,bottomtitle=1pt,
  fonttitle=\bfseries,coltitle=white,before skip=6pt,after skip=6pt}}
\newcounter{theorem}\newcounter{proposition}\newcounter{lemma}
\NewDocumentEnvironment{theorem}{o}{\refstepcounter{theorem}%
  \begin{tcolorbox}[result,colframe=red!45!black,colback=red!2,title={Theorem~\thetheorem\IfValueT{#1}{: #1}}]}{\end{tcolorbox}}
\NewDocumentEnvironment{proposition}{o}{\refstepcounter{proposition}%
  \begin{tcolorbox}[result,colframe=blue!50!black,colback=blue!2,title={Proposition~\theproposition\IfValueT{#1}{: #1}}]}{\end{tcolorbox}}
\NewDocumentEnvironment{lemma}{o}{\refstepcounter{lemma}%
  \begin{tcolorbox}[result,colframe=green!35!black,colback=green!2,title={Lemma~\thelemma\IfValueT{#1}{: #1}}]}{\end{tcolorbox}}

\newcommand{\R}{\mathbb{R}}
\newcommand{\E}{\mathbb{E}}
\newcommand{\Z}{\mathcal{Z}}
\newcommand{\X}{\mathcal{X}}
\newcommand{\Oset}{\mathrm{O}(k)}
\newcommand{\mQ}{\boldsymbol{Q}}\newcommand{\mK}{\boldsymbol{K}}\newcommand{\mD}{\boldsymbol{D}}
\newcommand{\mC}{\boldsymbol{C}}\newcommand{\mI}{\boldsymbol{I}}\newcommand{\vPhi}{\boldsymbol{\Phi}}
\title{\bfseries Proving and Improving the Platonic\\Representation Hypothesis with JEPAs}
\author{Randall Balestriero$^{1,3}$ \qquad Yann LeCun$^{2,3}$\\[1pt]
  \normalsize $^1$Brown University \quad $^2$New York University \quad $^3$AMI Labs \quad
  Correspondence: \href{mailto:rbalestr@brown.edu}{\texttt{rbalestr@brown.edu}}\\[2pt]
  \normalsize\raisebox{-.2em}{\includegraphics[height=1.1em]{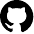}}\,\href{https://github.com/galilai-group/prh-proof}{\textbf{Code}}
  \qquad\raisebox{-.25em}{\includegraphics[height=1.2em]{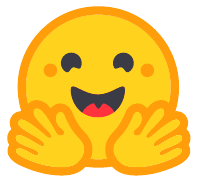}}\,\href{https://huggingface.co/spaces/galilai-group/prh_demo}{\textbf{Live demo}}}
\date{}

\begin{document}
\twocolumn[{%
\vspace*{-2.5em}
\maketitle
\thispagestyle{plain}
\begin{tcolorbox}[enhanced,colback=rule!18,boxrule=0pt,frame hidden,sharp corners,left=8pt,right=8pt,top=6pt,bottom=4pt]
Whether models trained on different modalities converge to the same representation of the world is central to AI. The Platonic Representation Hypothesis (PRH) conjectures that they do as they scale, but without a proof it offers neither a guarantee nor a recipe. We turn the hypothesis into a theorem. First, we prove a \emph{Platonic Representation Theorem}: Joint-Embedding Predictive Architectures (JEPAs) trained separately on any number of modalities learn the same representation up to one rotation whenever their positive pairs are two observations of one scene--\emph{good views}--and their misalignment is bounded otherwise. Second, we show that next-token prediction is a JEPA with \emph{bad views}, parts of a single document, which explains why language models align with vision only through scale. Third, we validate the theory from 145k captions to the texts of 940M web images, and uncover a scaling law of alignment that predicts unseen runs within seed noise. Good views need 42 times fewer training samples than next-token prediction on the same corpus to reach the same alignment, and a 109M-parameter text JEPA aligns with DINOv2 and CLIP's image encoders better than every language model we evaluated, up to 72B parameters, and than CLIP's text encoder: next-token prediction is a remarkably inefficient route to platonic representations. As a showcase, one rotation turns separately trained encoders into an image captioner (CIDEr .838 on COCO). Even that rotation needs no pair: on the latest pair-free benchmark, our text JEPA beats every zero-shot accuracy reported with language models up to 8B parameters, e.g., 64\% against at most 56\% top-5 on ImageNet-100. Positive views, and not scale alone, are thus the lever to build platonic representations.

\medskip
\centering\includegraphics{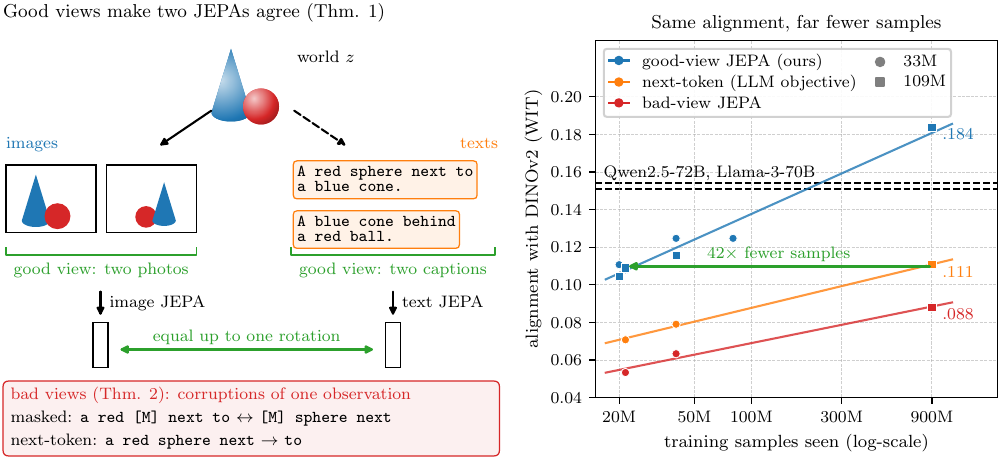}
\captionof{figure}{\textbf{Good views make the PRH a theorem, and alignment 42 times more sample-efficient. Left:} the PRH picture \citep{huh2024platonic}, one world $z$ observed as images and as texts. When a JEPA's positive pairs are two observations of one scene (\emph{good views}: two photographs, two captions), the encoders of every modality learn the same embedding up to one rotation (\cref{thm:exact}); bad views, two corruptions of one observation, and next-token prediction, a prefix and its continuation, align only partly (\cref{prop:goodbad,prop:ntp}). \textbf{Right:} alignment with DINOv2 on the PRH paper's benchmark (WIT-1024, mutual 10-NN) of text models trained on the texts of web images, against the samples they saw (means of three seeds, one for the 109M bad-view JEPA; lines: log-linear trends). With 21.6M samples a good-view JEPA matches next-token prediction on the same corpus at 900M samples (.109 and .111; 14 times fewer text tokens), and with 900M it reaches .184, above Qwen2.5-72B and Llama-3-70B, trained on 15--18T tokens (dashed). This demonstrates that \emph{with good views, alignment needs far less scale}.}
\label{fig:teaser}
\end{tcolorbox}
\vspace{0.2em}
}]

\section{Introduction}\label{sec:intro}

Whether different observers of one world end up with the same picture of it is a question with roots dating back millennia, from the shadows of Plato's cave to perception as unconscious inference \citep{plato_republic,vonhelmholtz1867} and the reduction of redundancy in sensory messages \citep{barlow1961possible}. Deep networks revived that question. Word embeddings trained separately on different languages turned out to be related by a linear map \citep{mikolov2013exploiting}, even by a rotation that can be found without a single translated pair, enough to translate words and then sentences with no parallel data \citep{conneau2018word,lample2018unsupervised}. Across modalities, \citet{huh2024platonic} observed that the representations of vision and language models become increasingly similar as models scale, and hypothesized that they converge to a shared statistical model of reality--coined the \emph{Platonic Representation Hypothesis} (PRH).

Alas, the PRH remains an observation. Its supporting argument is an idealized contrastive learner whose kernel converges to the pointwise mutual information of co-occurring world events \citep{huh2024platonic}, the closest proof we know of covers embedded deep linear networks trained by SGD \citep{ziyin2025perfect}, and a recent theory explains why translations share representations inside multilingual language models \citep{doshi2026theory}. For the self-supervised objectives used in practice, and in particular for Joint-Embedding Predictive Architectures (JEPAs) \citep{lecun2022path,balestriero2025lejepa}, there is no proof, no prediction of when the hypothesis holds, and hence no recipe to improve alignment. As a result, scaling models and data remains the main lever used so far.

Our study proposes to turn that observation into a theorem by asking the following question: \emph{when do models, trained separately on different modalities, provably learn the same representation?} We answer it for JEPAs with a short argument. A JEPA with whitened second-order anti-collapse learns the spectral embedding of its \emph{positive pairs}, the pairs of inputs it declares to be views of each other \citep{haochen2021provable,balestriero2022contrastive}. If the modalities render one world and their positive pairs relate the same world states, their training problems are one problem written in different coordinate systems, and their solutions coincide up to a rotation--which we coin the \emph{Platonic Representation Theorem} (\cref{thm:exact}). Nothing in this argument is specific to images and text: it holds for any modalities that observe one world, e.g., images, video, text, audio or actions, and for any number of them at once. We therefore obtain the following key insight. \emph{The PRH is a statement about positive views: a JEPA whose positives are two observations of the same scene learns a representation that belongs to the world, while a JEPA whose positives are corruptions of one observation learns one that belongs to its modality} (\cref{fig:teaser}). We coin the former \emph{good-view} and the latter \emph{bad-view} JEPAs, and summarize our contributions below.

\textbf{Contribution 1: We turn the PRH into a theorem.} In an idealized model, any number of JEPAs trained separately on renderings of one world learn the same embedding up to one rotation each when their positive pairs relate the same world states and their minimizers are unique up to rotation, as we prove for whitened JEPAs under a spectral gap (\cref{thm:exact,prop:spectral}); otherwise, their misalignment is bounded by the mismatch of their positive pairs (\cref{thm:robust}). Positives that corrupt a single observation make an encoder spend dimensions on its modality's nuisance whenever they keep more of it than of the world's $k$-th direction, which no rotation can align, while two observations of one scene spend none (\cref{thm:nuisance,prop:goodbad}), and views and data together set what a JEPA learns (\cref{thm:coverage}).

\textbf{Contribution 2: We show that next-token prediction is a bad-view JEPA.} Its positives are a prefix and its continuation within one document, and its anti-collapse term is the softmax log-partition (\cref{prop:ntp}). This suggests why language models align with vision only partly, and mostly through scale.

\textbf{Contribution 3: We validate good views and uncover a scaling law of alignment.} With identical texts and encoders, good views align best in all but a few of the comparisons we ran (\cref{app:results}): from 145k Flickr captions to the texts of 940M web images, with image encoders trained from scratch or frozen, and on video (\cref{sec:exp}). Alignment follows a scaling law in the fraction of good views and the samples seen, which predicts 85 new runs, before they are trained, as accurately as rerunning them with another seed (\cref{sec:law}).

\textbf{Contribution 4: We improve alignment beyond 70B language models, which scale poorly in comparison.} Good views help in three ways (\cref{sec:scaling}). \emph{They scale with data:} trained on the texts of 50M web images, a 33M-parameter text JEPA reaches .125 on the PRH paper's benchmark, against .034 on Flickr captions, level with Qwen2.5-3B and -7B. \emph{They need fewer samples:} on the same corpus, a good-view JEPA reaches with 21.6M training samples the alignment that next-token prediction reaches with 900M, 42 times more (\cref{fig:teaser}). \emph{They beat larger models:} trained on the texts of 940M web images, a 109M-parameter text JEPA reaches .184, above every language model we evaluated, up to Qwen2.5-72B (.154), and above CLIP's text encoder (.124), on two benchmarks and two metrics; it also leads the language models with MAE, I-JEPA and CLIP's image encoders, e.g., .230 against .210 with CLIP ViT-H. For alignment, next-token prediction is thus a remarkably inefficient use of scale: our 109M encoder, which read 116B text tokens, beats models 660 times larger trained on over 100 times more.

\textbf{Contribution 5: We caption images without training a network on image--caption pairs.} One linear map with orthonormal rows, the only part fitted on pairs, joins frozen DINOv2 and DINOv3 image encoders to a language model fine-tuned on text alone, reaching CIDEr .838 on COCO (\cref{sec:caption}). With no pair at all, the rotation can be found from the geometry of unpaired images and captions: on the benchmark of \citet{schnaus2026rosetta}, which aligns a vision and a language encoder from unpaired images and captions and scores the result by zero-shot classification, our 109M text JEPA beats every zero-shot accuracy they report, over 21 pairs of vision and language models and five settings, and it beats their strongest language model, Qwen3-8B, rerun under identical conditions, by 17 points of zero-shot ImageNet-100 accuracy (\cref{sec:pairfree}). This demonstrates that the PRH--with JEPAs--is no longer a hypothesis but a theorem and a tool.

\section{Shared Positives Provably Make Representations Platonic}\label{sec:theory}

We start by formalizing a world, the modalities that observe it and their positive pairs, then prove that JEPAs with shared positives agree up to one rotation (\cref{thm:exact}), bound their misalignment otherwise (\cref{thm:robust}), and show which positives make an encoder spend its embedding on its modality's nuisance (\cref{thm:nuisance,prop:goodbad,thm:coverage}), of which next-token prediction is an instance (\cref{prop:ntp}).

\subsection{The Platonic Representation Theorem}\label{sec:thm}
\paragraph{World, modalities, and positive pairs.}
Let a world be a finite set of states $z\in\Z$ with probabilities $p(z)>0$, and let $M\ge2$ modalities observe it, modality $m$ through an injective \emph{rendering} $x=g_m(z)$, e.g., the pixels of a photograph or the tokens of a caption. A JEPA of modality $m$ is trained on positive pairs $(g_m(z),g_m(z'))$ whose world states are drawn from a symmetric \emph{positive-pair distribution} $(z,z')\sim\pi_m$ with both marginals equal to $p$. Its encoder $f:\X_m\to\R^k$ minimizes
\begin{equation}\label{eq:jepa}
\mathcal L_m(f)=\E_{(z,z')\sim\pi_m}\,\ell\big(f(g_m(z)),f(g_m(z'))\big)+\Omega(f),
\end{equation}
a prediction term $\ell$ plus an anti-collapse term $\Omega$ that depends on $f$ only through the distribution of its embeddings (possibly a hard constraint, $\Omega\in\{0,\infty\}$). As in our experiments, the predictor is the identity: $\ell$ compares the two embeddings directly, whereas JEPAs used as world models predict one embedding from the other through a learned predictor, often conditioned on an action, which our analysis does not cover. Both $\ell$ and $\Omega$ are invariant to rotations $\mQ\in\Oset$: $\ell(\mQ a,\mQ b)=\ell(a,b)$, and rotating every embedding leaves $\Omega$ unchanged. Spectral contrastive learning and whitened MSE fit this form \citep{haochen2021provable,ermolov2021whitening}, VICReg and Barlow Twins in a whitened limit \citep{bardes2022vicreg,zbontar2021barlow,balestriero2022contrastive}, and LeJEPA's SIGReg, which pushes the embeddings toward an isotropic Gaussian, in expectation over its directions \citep{balestriero2025lejepa} (\cref{app:proofs}). The key observation is that an injective rendering only changes coordinates: $\mathcal L_m(f)=\mathcal L_{\pi_m}(f\circ g_m)$, the same objective written on world states, $\mathcal L_\pi(h)=\E_{(z,z')\sim\pi}\ell(h(z),h(z'))+\Omega(h)$ for $h:\Z\to\R^k$ (\cref{lem:cov}). We are now able to state when modalities learn the same representation.

\begin{theorem}[Platonic Representation Theorem]\label{thm:exact}
Let $M\ge2$ modalities share one positive-pair distribution, $\pi_1=\dots=\pi_M=\pi$, and let the minimizers of $\mathcal L_\pi$ be unique up to rotation. Then there is one embedding of the world $h^\star:\Z\to\R^k$ such that every minimizer $f_m^\star$ of $\mathcal L_m$, for every modality $m$, satisfies
\begin{equation}
f_m^\star\circ g_m=\mQ_m\,h^\star\quad\text{for some }\mQ_m\in\Oset,
\end{equation}
so any two modalities agree up to the single rotation $\mQ_m\mQ_{m'}^\top$. (Proof in \cref{app:proofs}.)
\end{theorem}

In words, encoders that never see each other's modality, and whose renderings need not be related beyond describing the same world, learn the same embedding up to one rotation each: in this idealized JEPA setting--one finite world rendered injectively, shared positives, population optima, and minimizers unique up to rotation--the PRH becomes a theorem, for images and texts as for any other set of modalities. The proof is direct: every $f_m^\star\circ g_m$ minimizes $\mathcal L_\pi$. The key assumption is uniqueness up to rotation, which holds for JEPAs with second-order anti-collapse under a spectral gap, as formalized below. Write $\boldsymbol\Pi$ for the matrix of $\pi$ ($\Pi_{zz'}=\pi(z,z')$), $\mD=\mathrm{diag}(p)$, and $\mK_\pi=\mD^{-1/2}\boldsymbol\Pi\mD^{-1/2}$ for the normalized adjacency of the graph that positives draw on the world, with eigenvalues $1=\mu_0\ge\mu_1\ge\dots$ and orthonormal eigenvectors $\boldsymbol v_0=\mD^{1/2}\mathbf 1,\boldsymbol v_1,\dots$.

\begin{proposition}[Spectral JEPAs Are Unique Up to Rotation]\label{prop:spectral}
Let $\ell(a,b)=\|a-b\|^2$, let $\Omega$ whiten the embeddings ($\E_p h=0$, $\E_p hh^\top=\mI_k$), and let $1\le k<|\Z|$. Then $\min\mathcal L_\pi=2\sum_{i=1}^k(1-\mu_i)$, and if $\mu_k>\mu_{k+1}$ (with $\mu_{|\Z|}:=-\infty$) the minimizers are exactly $\mQ\vPhi_\pi$, $\mQ\in\Oset$, with the \emph{spectral embedding} $\vPhi_\pi(z)=\big(v_1(z),\dots,v_k(z)\big)/\sqrt{p(z)}$. (Proof in \cref{app:proofs}.)
\end{proposition}

In words, a JEPA with second-order anti-collapse learns the top eigenvectors of the graph its positives draw, the population form of Laplacian eigenmaps \citep{belkin2003laplacian,balestriero2022contrastive,haochen2021provable}, unique up to rotation when the spectrum has a gap. LeJEPA's SIGReg target $\mathcal N(0,\mI_k)$ is rotation invariant and has the same second moments; \cref{prop:spectral} does not cover uniqueness under its full-distribution constraint, which we leave as an assumption.

\subsection{When Positives Differ}\label{sec:differ}
Alas, in practice the modalities' positives are not the same pairs of world states.

\begin{figure}[t]
\centering
\includegraphics{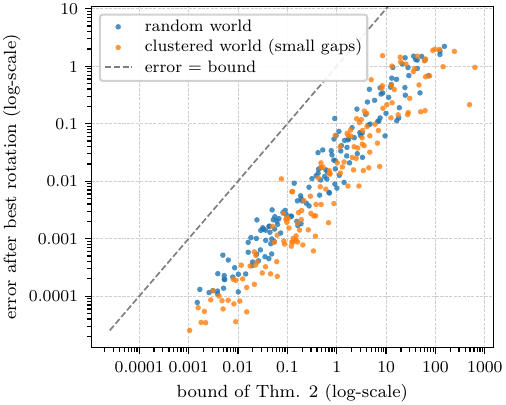}
\caption{\textbf{Kernel mismatch bounds misalignment (\cref{thm:robust}).} Each point is one randomly generated finite world (8--39 states, random marginal, embedding dimension $k$ from 1 to 5) with a pair of kernels on it: the world's kernel, and a second kernel of the same marginal that mixes in an unrelated kernel with weight $10^{-4}$ to $1$, as if a second modality had partly different positives. For each pair we compute the two spectral embeddings (\cref{prop:spectral}) and plot their distance after the best rotation (vertical) against the bound of \cref{thm:robust} (horizontal); of the 300 worlds, half are clustered, with small spectral gaps. We clearly observe that \emph{the error lies below the bound in every world} (dashed: equality), at 18\% of it at most.}
\label{fig:dk}
\end{figure}

\begin{theorem}[Kernel Mismatch Bounds Misalignment]\label{thm:robust}
Let $\pi_1,\pi_2$ share the marginal $p$, let $\vPhi_1,\vPhi_2$ be their spectral embeddings (any orthonormal top-$k$ eigenvectors if $\mK_{\pi_1}$ has ties), and let $\delta=\mu_k-\mu_{k+1}>0$ be the eigengap of $\mK_{\pi_2}$. Then
\begin{equation}
\begin{split}
&\min_{\mQ\in\Oset}\Big(\E_{p}\big\|\vPhi_1(z)-\mQ\vPhi_2(z)\big\|^2\Big)^{\!1/2}\\
&\quad\le\frac{2\sqrt2}{\delta}\min\big(\sqrt k\,\|\mK_{\pi_1}{-}\mK_{\pi_2}\|_2,\,\|\mK_{\pi_1}{-}\mK_{\pi_2}\|_F\big).
\end{split}
\end{equation}
(Proof in \cref{app:proofs}, empirical validation in \cref{fig:dk}.)
\end{theorem}

In words, for the objective of \cref{prop:spectral} the left side is the root-mean-square distance between the embeddings of paired observations after the best single rotation, and only two quantities matter: how different the graphs drawn by the two modalities' positives are, and how separated the top of their spectrum is. Crucially, the same argument applied to a block of eigenvectors, with gaps on both sides of the block \citep{yu2015useful}, suggests where alignment lives: directions separated from the rest of the spectrum by large gaps are protected, while directions near the end of the spectrum need not align at all. \Cref{fig:dk} validates the bound on randomly generated finite worlds, each point being one world with a pair of kernels that differ by a controlled amount: in all 300 worlds the error lies below the bound, at 18\% of it at most.

\paragraph{What pairs are for.}
\Cref{thm:exact} says that the embeddings agree up to a rotation, but not which one. As formalized below, the first two moments of unpaired data cannot find it, and $k$ pairs suffice.

\begin{proposition}[$k$ Pairs Identify the Rotation]\label{prop:ident}
In the setting of \cref{thm:exact,prop:spectral}, (i) every rotation of a whitened embedding is whitened, so no statistic of the first two moments of unpaired samples from two modalities identifies the rotation between them; (ii) if the embeddings of $n\ge k$ world states span $\R^k$, the rotation $\mQ$ with $f_1^\star\circ g_1=\mQ f_2^\star\circ g_2$ is the unique minimizer over $\boldsymbol R\in\Oset$ of $\sum_{i\le n}\|f_1^\star(g_1(z_i))-\boldsymbol R f_2^\star(g_2(z_i))\|^2$ (orthogonal Procrustes). (Proof in \cref{app:proofs}.)
\end{proposition}

In words, pairs enter only through one $k\times k$ orthogonal matrix, which $k$ pairs in general position determine; when the positives differ, more pairs average out the residual misalignment of \cref{thm:robust}. Higher-order structure of a finite world can reveal the rotation without any pair, as recent work shows by aligning encoders from their shared geometry alone \citep{schnaus2026rosetta}; with good views, whose geometry the theory ties to the world's, this works best (\cref{sec:pairfree}). When the two embedding dimensions differ (768 for DINOv2, 384 for our text JEPA) we fit a linear map with orthonormal rows, a case the theory does not cover; such a map between dimensions $d_a\ge d_b$ has $d_ad_b-d_b(d_b+1)/2$ free parameters, 0.22M for DINOv2 ViT-B and 1.9M for the 5{,}120-dimensional features of our best captioner, all fitted on pairs.

\subsection{Good-View and Bad-View JEPAs}\label{sec:goodbad}
\Cref{thm:robust} singles out what a JEPA must get right for the PRH: its positive pairs. Call a JEPA \emph{good-view} when its positives are two observations of world states that the world itself relates, e.g., two photographs or two descriptions of one scene, and \emph{bad-view} when they are corruptions of a single observation, e.g., two crops of one photograph or two maskings of one sentence (\cref{fig:teaser}). Injective renderings cannot express this distinction: there, a state has a single observation. We therefore let each observation carry a \emph{nuisance} $\nu$ of its own modality--which words a caption happens to use, from which viewpoint a photograph was taken--drawn independently of the state. It is a nuisance only for alignment: a world model that acts through a camera must track its viewpoint, but no caption determines it, so no rotation can match it across modalities. We render $x=g_m(z,\nu)$ injectively, so that a state has many observations. A good-view JEPA's positives are $(g_m(z,\nu),g_m(z',\nu'))$ with $(z,z')\sim\pi$ and independent nuisances; a bad-view JEPA's are two corruptions of one observation, which share its nuisance. For an encoder $f$ centered and whitened on its observations, let $\mC_f=\mathrm{Cov}\big(\E[f(x)\mid z]\big)$ be the covariance of the part of its embedding that the world state explains, and $\rho(f)=\mathrm{tr}\,\mC_f^{1/2}/k\in[0,1]$ its \emph{scene-explained fraction}. The cost of encoding nuisance is formalized below.

\begin{theorem}[Nuisance Costs Alignment]\label{thm:nuisance}
Let $f_1$ and $f_2$ be centered, whitened encoders of two modalities into $\R^k$. Over paired observations $(x_1,x_2)$ of one world state with independent nuisances,
\begin{equation}
\min_{\mQ\in\Oset}\E\big\|\mQ f_1(x_1)-f_2(x_2)\big\|^2\ge2k\big(1-\min_m\rho(f_m)\big),
\end{equation}
and $\rho(f)=1$ if and only if $f(x)$ is a function of the world state alone. (Proof in \cref{app:proofs}.)
\end{theorem}

In words, every direction an encoder spends on its own modality's nuisance is lost for alignment, whatever whitened embedding the other modality learns. Which JEPAs spend directions on nuisance is formalized below.

\begin{proposition}[Good Views Ignore Nuisance, Bad Views Capture It]\label{prop:goodbad}
Take the objective of \cref{prop:spectral}. (i) If $\mu_k>\max(\mu_{k+1},0)$, a good-view JEPA learns a function of the world state alone ($\rho=1$): good-view JEPAs of any number of modalities with the same world kernel $\pi$ agree up to one rotation each on every observation, although every state has many. (ii) Let a bad-view JEPA corrupt state and nuisance independently, let $\alpha_k$ be the correlation that two corruptions of one observation keep along the $k$-th direction of the state, and $\beta_1$ the largest they keep along a direction of the nuisance. If its kernel has a gap after its $k$-th eigenvalue, it spends $r$ of its $k$ directions on nuisance, so that its alignment error with any centered, whitened encoder of another modality is at least $2r$, and $r\ge1$ if and only if $\beta_1>\alpha_k$. (Formal statement and proof in \cref{app:proofs}.)
\end{proposition}

In words, a corruption that preserves the nuisance more than the world's $k$-th direction--spans of one caption keep its word choice, crops of one photograph its viewpoint and lighting--makes the JEPA spend embedding directions on that nuisance, which no rotation can align with another modality; two observations of one scene share only the scene, so a good-view JEPA spends none. The model still assumes that every observation determines its state, while a caption sees only part of a photograph; \cref{thm:robust} covers that departure only through the kernel mismatch. By the same definition our image encoders, and DINOv2, are bad-view JEPAs: their positives are crops of one photograph. Our experiments mostly vary the text side against a frozen DINOv2, the vision model with which \citet{huh2024platonic} measured the PRH, so that our text encoders compare directly with the language models measured there; DINOv3 \citep{simeoni2025dinov3}, released after that benchmark, serves in our captioner (\cref{sec:caption}), and other vision families rank the text side alike (\cref{tab:vision_families}). Good image views need several observations of each scene, e.g., the frames of a video: MSR-VTT tests them (\cref{sec:both}), but an image JEPA trained on its 7k clips is far weaker than DINOv2, and scaling good image views is left to future work.

\begin{figure*}[t]
\centering
\input{fig_positives}
\caption{\textbf{What counts as a positive, and which observations it connects.} Eight captions of two scenes (dots), joined as each objective's positives join them. \textbf{(a)} Good views join every description of a scene and never another, so the world kernel they induce is the world's own (\cref{thm:coverage}). \textbf{(b)} Weak corruptions mostly pair a caption with itself, keeping its wording, and join other captions only where their corruptions overlap, which takes dense data. \textbf{(c)} Strong corruptions also coincide across scenes, so the induced kernel mixes scenes whatever the amount of data. \textbf{(d)} Next-token prediction pairs a prefix with its own continuation, and implicitly contexts followed by the same token (\cref{prop:ntp}): narrow links within scenes, plus spill through frequent tokens. That is, \emph{good views fix the world kernel by design, while bad views and next-token prediction get it from the data}.}
\label{fig:positives}
\end{figure*}

\paragraph{Views and data together.}
\Cref{prop:goodbad} lets a corruption act on state and nuisance separately. In general, whether the positives of two different observations coincide depends on the data: masking two captions of one scene can leave the same words. In the augmentation-graph analyses of contrastive learning, such overlaps are what connect samples of one class \citep{haochen2021provable,wang2022chaos}, with the function class deciding what overlaps \citep{saunshi2022understanding,kiani2022joint,cabannes2023ssl}. We now state an analogue with world states in place of classes, and its cross-modal consequence (\cref{fig:positives}). Given any law $W_m$ of positive pairs over modality $m$'s finitely many observations--good views, corruptions, or a mix--call its \emph{induced world kernel} $\pi_{W_m}$ the probability that a positive pair joins an observation of $z$ to one of $z'$.

\begin{theorem}[Views and Data Set the World Kernel]\label{thm:coverage}
Take the objective of \cref{prop:spectral} on the observations of each modality $m$, and let the $k$-th eigenvalue of $\mK_{\pi_{W_m}}$ exceed both the next one and the share of nuisance its positives keep. (i) Every spectral solution $f_m$ is within $\tau_m$ of a rotation of the spectral embedding of $\pi_{W_m}$, where $\tau_m$ grows with how much positive pairs mix world state with nuisance and vanishes when they do not. (ii) Over paired observations of two modalities,
\begin{equation}
\Big|\min_{\mQ\in\Oset}\Big(\E\big\|f_1(x_1)-\mQ f_2(x_2)\big\|^2\Big)^{\!1/2}-d\Big|\le\tau_1+\tau_2,
\end{equation}
with $d$ the misalignment of the spectral embeddings of $\pi_{W_1}$ and $\pi_{W_2}$, which \cref{thm:robust} bounds. (Formal statement and proof in \cref{app:proofs}.)
\end{theorem}

In words, a spectral JEPA learns, up to $\tau_m$, the spectral solution of the world kernel that its views and its data induce, and two such JEPAs of different modalities agree, up to $\tau_1+\tau_2$, exactly as well as those kernels' top eigenvectors do. Good views induce the world's own kernel with $\tau_m=0$, which gives \cref{thm:exact} with many observations per state. Crucially, a bad-view JEPA does not choose its world kernel: $\pi_{W_m}$ records which states its positives connect, which depends on how the data cover each state and its neighbors. Bad-view JEPAs of two modalities therefore agree only insofar as their data make the positives connect the same states, while good views fix the kernel by design.

\begin{figure}[t]
\centering
\includegraphics{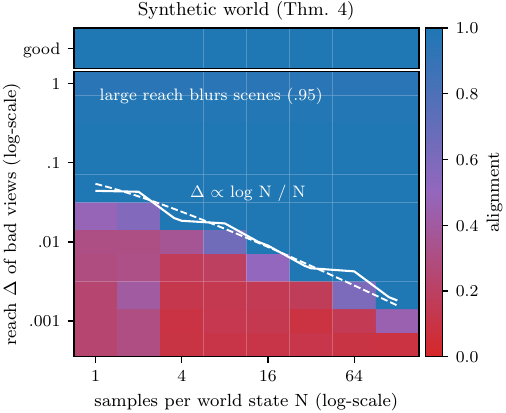}
\caption{\textbf{Positives that join less of a scene need more data (\cref{thm:coverage}).} A synthetic world of five classes of six states (literal data, as in \cref{fig:coverage}): alignment of a text JEPA with the image side against the samples per state $N$ and the reach $\Delta$ of its bad views (means over three samplings; a state has width $1/6$). White: the 0.9 contour, and $\Delta\propto\log N/N$ (dashed, one fitted constant), the scale of the largest gap between a state's samples. Reach large enough to cross classes caps alignment (.95), and good views (top strip) align exactly at any $N$. We observe that \emph{weaker positives need more data for the same alignment}.}
\label{fig:phase}
\end{figure}

\paragraph{How much data, and which.} Read on finitely many samples, with positives that join two samples when their corruptions overlap--the empirical augmentation graph--weak views join few samples, so each level set of the world splits into small pieces that a spectral JEPA separates, and more samples chain the overlaps through each level set. Good views join all observations of related states in one step and need no such density. The kind of data, in contrast, sets the kernel itself: chains of overlapping positives stay within a scene for literal descriptions, but cross into states that share only a nuisance for alt-texts sharing a brand name. \Cref{fig:phase} maps this trade-off on a synthetic world: the reach that weak bad views need falls as the samples per state grow, along $\Delta\propto\log N/N$, the scale of the largest gap between a state's samples; reach large enough to cross scenes caps alignment at any sample size; and good views align exactly from one sample per state. Mixed data, whose classes overlap, cap every bad-view JEPA lower still (\cref{fig:coverage}). Our runs show the same trade-off: a good-view JEPA reaches the alignment of next-token prediction on the same corpus with 42 times fewer samples (\cref{fig:teaser}), and alignment follows a scaling law in the fraction of good views and the samples seen (\cref{sec:law}). Real data agree with this reading as well: with the same bad views, a text JEPA trained on LLaVA recaptions, literal descriptions of what is visible, aligns with DINOv2 better than one trained on alt-texts (\cref{tab:sources}); changing the data changes how positives mix state and nuisance as well as the kernel, so this comparison cannot single out the kernel.

\subsection{Next-Token Prediction is a Bad-View JEPA}\label{sec:ntp}

Language models are the usual text side of the PRH \citep{huh2024platonic}. We now show that their training objective is a JEPA in disguise, as formalized below.

\begin{proposition}[Next-Token Prediction Has the Form of a JEPA]\label{prop:ntp}
Let a context $c$ be embedded by $h(c)\in\R^k$ and tokens by a table $\{w_v\}_{v\in V}$. The next-token loss with a linear softmax head is
\begin{equation}\label{eq:ntp}
\begin{split}
-\log\frac{e^{\langle h(c),w_y\rangle}}{\sum_v e^{\langle h(c),w_v\rangle}}={}&\underbrace{-\langle h(c),w_y\rangle}_{\text{prediction}}\\
&+\underbrace{\log\textstyle\sum_v e^{\langle h(c),w_v\rangle}}_{\text{anti-collapse }A(h(c))}.
\end{split}
\end{equation}
(i) This has the structure of a JEPA, generalizing \eqref{eq:jepa} to two encoders and an anti-collapse term that depends on the target encoder: context encoder $h$, target encoder $y\mapsto w_y$, inner-product prediction, positives $(c,y)$ drawn from within one sequence. (ii) $A$ does not depend on the target token $y$, and $tA(h/t)\to\max_v\langle h,w_v\rangle$ as $t\to0$, so the temperature-scaled loss approaches the margin $\max_v\langle h,w_v\rangle-\langle h,w_y\rangle\ge0$, which vanishes iff the target attains the maximal logit. (iii) With full-support conditionals and $k$ large enough, minimizers satisfy $\langle h(c),w_v\rangle=\log p(v\mid c)+b(c)$, a set invariant under $\mathrm{GL}(k)$ ($h\mapsto \boldsymbol Gh$, $w\mapsto \boldsymbol G^{-\top}w$), so $h$ is identified at best up to an invertible linear map, not a rotation. (Proof in \cref{app:proofs}.)
\end{proposition}

In words, next-token prediction is a JEPA whose positives are a prefix and its continuation, and whose anti-collapse term is the log-partition. Part (iii) is the matrix-factorization view of word embeddings \citep{levy2014neural}, and (ii) plays the role of the uniformity term of contrastive learning \citep{wang2020understanding}, much as masked autoencoders implicitly align the two parts that a mask creates \citep{zhang2022mask}. Crucially, (i) tells us what next-token prediction's positives are: a prefix and its continuation inside one document. Its kernel therefore relates texts through shared wording, like masked-span JEPAs, and not through shared world states. Heuristically, \cref{prop:goodbad} then applies to it: a prefix and its continuation share their document's nuisance (its author's wording and style), so next-token prediction should spend directions on that nuisance and align with vision worse than a good-view JEPA. \Cref{prop:goodbad} assumes one encoder and a kernel on one space, so this is a prediction to test rather than a corollary--which we do next.

Read through \cref{thm:coverage}, again heuristically since next-token prediction is not a spectral JEPA and is identified only up to $\mathrm{GL}(k)$, it has a second kind of positive. At the optimum (iii), $\langle h(c),w_y\rangle=\log p(y\mid c)+b(c)$, and contexts followed by the same token gather around that token's output vector, the neural collapse that language models approach with scale \citep{papyan2020prevalence,wu2024linguistic}. Contexts thus become implicit positives of each other when they share a continuation: very weak views, since one token rarely spans a scene, so joining the descriptions of one scene takes long chains of shared continuations, i.e., very dense data, while frequent continuations (function words, a document's style) also join contexts across scenes. Good-view JEPA views instead join all descriptions of a scene at once and never leave it. This is our reading of why language models align with vision mostly through scale, and why good views need far less of it.

\section{Good Views Improve Alignment}\label{sec:exp}

We now empirically validate the predictions of \cref{sec:theory} and \cref{prop:ntp}: good-view JEPAs should align across modalities better than bad-view JEPAs and next-token prediction trained on identical data, whatever the data, the encoders and the modality.

\paragraph{Setup.} We strive for simplicity and thus adopt a unified recipe: image and text encoders are trained \emph{separately}, each with the same LeJEPA objective \citep{balestriero2025lejepa} (invariance to the mean of two global views, plus SIGReg with $\lambda=0.02$), projector, optimizer and schedule, using stable-pretraining. Data is Flickr30k \citep{young2014flickr} with the Karpathy split \citep{karpathy2015deep}: 29k training images with five captions each. The image encoder is a ViT-S/16 \citep{dosovitskiy2021vit} trained on 2 global ($224^2$) and 6 local ($96^2$) crops. The text encoder is a 12-layer BERT of the same width (384, 33M parameters; BERT-S below, with BERT-T, -B and -L of 11M, 109M and 335M parameters for scaling) \citep{devlin2019bert}, trained from scratch in three variants that differ only in their positives:
\textbf{masked spans} (bad-view JEPA): 8 views of one caption, each trimmed to a random contiguous span, with 15\% of tokens masked;
\textbf{caption views} (good-view JEPA): each of the 8 views is drawn from a different caption of the same image, then trimmed and masked the same way;
\textbf{next-token prediction}: the same BERT made causal, with a tied softmax head, on one caption per image per epoch; as a control, we also train it on pairs of captions of the same image.
No encoder ever sees the other modality. The runs of \cref{tab:main} use 200 epochs, batch 256 and AdamW (lr $5\!\cdot\!10^{-4}$, weight decay 0.05); longer and larger runs use a lower learning rate with gradient clipping. We train three seeds of every comparison and report their mean unless stated otherwise. Details are in \cref{app:details}.

\paragraph{Metrics.} We embed the 1{,}014 validation and 1{,}000 test image--caption pairs (first caption). \emph{Mutual 10-NN} \citep{huh2024platonic} measures the overlap of the 10-nearest-neighbor sets in the two spaces without fitting anything; chance is $10/N=.01$. To test equality \emph{up to rotation} directly, we fit one orthogonal map on the validation pairs \citep{schonemann1966generalized} and report image$\to$caption retrieval R@1 and R@5 among the 1{,}000 test captions (chance 0.1\%).

\begin{table}[t]
\centering
\small
\setlength{\tabcolsep}{4pt}
\caption{\textbf{Good views align across modalities.} Image JEPA vs.\ each text encoder (epoch 199) on 1{,}000 Flickr test pairs (backbone features; mean $\pm$ half-range over three seeds): good-view JEPA (caption views), bad-view JEPA (masked spans), next-token prediction, and next-token prediction on pairs of captions of one image; R@1 and R@5 use a single rotation fitted on validation pairs. We clearly observe that \emph{caption views beat masked spans and next-token prediction on every metric}, even when next-token prediction is trained on pairs of captions of the same image.}
\label{tab:main}
\begin{tabular}{lccc}
\toprule
Text encoder & mutual 10-NN & R@1 & R@5 \\
\midrule
Good-view JEPA & \textbf{.062}{\scriptsize$\pm$.003} & \textbf{.044}{\scriptsize$\pm$.002} & \textbf{.133}{\scriptsize$\pm$.004} \\
Bad-view JEPA & .047{\scriptsize$\pm$.001} & .027{\scriptsize$\pm$.002} & .095{\scriptsize$\pm$.003} \\
Next-token & .042{\scriptsize$\pm$.001} & .021{\scriptsize$\pm$.002} & .077{\scriptsize$\pm$.005} \\
Next-token, pairs & .043{\scriptsize$\pm$.001} & .023{\scriptsize$\pm$.003} & .083{\scriptsize$\pm$.008} \\
\midrule
Random init & .010 & .001 & .011 \\
Chance & .010 & .001 & .005 \\
\bottomrule
\end{tabular}
\end{table}

\begin{figure*}[t]
\centering
\includegraphics{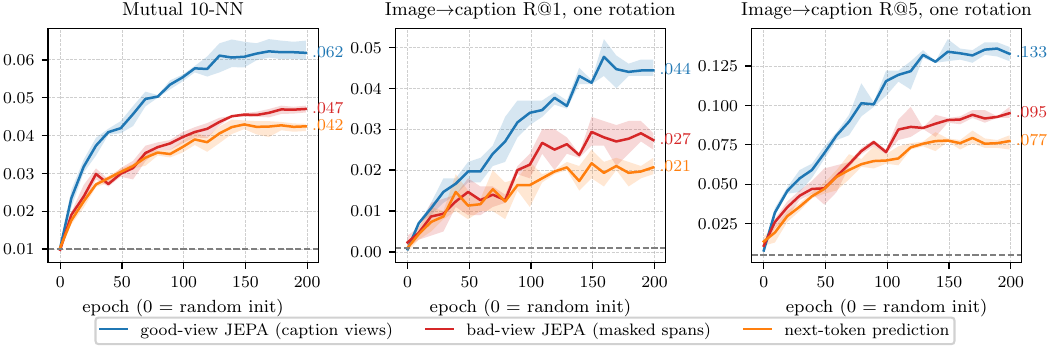}
\caption{\textbf{Good views align, bad views and next-token prediction do not.} Alignment between a separately trained image JEPA and three text encoders on Flickr, at matched epochs (mean of three seeds, band: range; dashed: chance). All start at chance. We clearly observe that \emph{the good-view JEPA (caption views) aligns markedly better than the bad-view JEPA (masked spans) and next-token prediction, which nearly coincide}, as predicted by \cref{thm:robust,prop:ntp}.}
\label{fig:compare}
\end{figure*}

\subsection{Good Views Align Across Modalities}
\begin{figure}[t]
\centering
\includegraphics{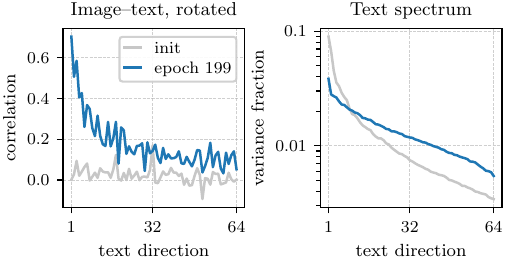}
\caption{\textbf{Alignment lives in the leading spectral directions.} Left: after the best rotation, correlation between image and text embeddings along the text embedding's principal directions (good-view JEPA, at initialization and at epoch 199). Right: variance of the text embedding along the same directions. This suggests that \emph{alignment concentrates where the spectrum is well separated}, consistent with \cref{thm:robust}, although noise in low-variance directions would produce the same trend.}
\label{fig:spectrum}
\end{figure}

\Cref{fig:compare,tab:main} report the alignment of a separately trained image JEPA with each text encoder, and we obtain a few key observations. First, cross-modal alignment rises from chance throughout training for every text encoder, although none ever sees the other modality. Second, with caption views the true caption is retrieved first for 4.4\% of test images through one rotation, 44 times chance, while masked spans and next-token prediction, trained on the same captions with the same encoder, nearly coincide and stay well below--a ranking that holds in every seed, with gaps several times the seed spread. Third, the gain comes from the objective and not from the information that two captions describe the same image: next-token prediction trained on two captions of the same image, \texttt{[CLS]}\,$a$\,\texttt{[SEP]}\,$b$\,\texttt{[SEP]}, barely moves (mutual 10-NN .042 to .043, R@1 .021 to .023), i.e., the same positives help next-token prediction far less than they help as JEPA views, as per \cref{prop:ntp}. Fourth, alignment concentrates in the leading spectral directions, as the block version of \cref{thm:robust} suggests: after the best rotation, the correlation between image and text embeddings is 0.70 on the first principal direction of the text space and decays to about 0.1 by the 60th, while it is near zero everywhere at initialization (\cref{fig:spectrum}). Finally, the nuisance mechanism of \cref{thm:nuisance} can be measured directly, since every Flickr image (a world state) has five captions (five nuisances): from all captions of the 2{,}014 validation and test images, we estimate $\mC_f$ for each text encoder, restricted to its top-$k$ principal directions and whitened, with the mean of five captions corrected for its within-image spread. For $k=32$, the scene explains a fraction $\rho$ of .804 of the good-view JEPA's embedding, against .655 for the bad-view JEPA and .608 for next-token prediction (each within $\pm$.005 across seeds), so \cref{thm:nuisance} bounds their best-rotation error against DINOv2 below by 12.6, 22.1 and 25.1 (from the unrounded fractions), which the measured errors (33.5, 39.4 and 40.3) respect and follow in order; $k=128$ gives the same ordering (\cref{tab:nuisance}). That is, \emph{bad views and next-token prediction spend more of their embedding on the wording of a particular caption, as \cref{prop:goodbad} predicts, and align less}.

\subsection{Trimming Makes Good Views, the Rest Barely Matters}
\begin{figure}[t]
\centering
\includegraphics{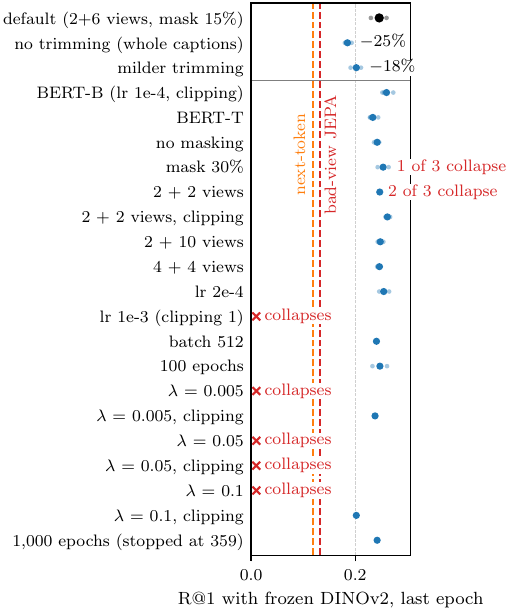}
\caption{\textbf{Trimming is the knob that matters.} One change at a time to the good-view text JEPA on Flickr: R@1 with frozen DINOv2 through one rotation, last epoch, three seeds where they exist (faint dots; large dot: mean of the seeds that do not collapse; black: default). Top: the two changes to trimming, with the change of their mean from the default's; whole captions as views lose a quarter of the alignment. Dashed: the bad-view JEPA and next-token prediction with the default recipe. That is, \emph{trimming matters most, and every variant of the good-view JEPA that does not collapse stays far above both baselines}; gradient clipping prevents most collapses.}
\label{fig:ablations}
\end{figure}

Recall that a good view should keep only what two observations share. To identify which knob of the good-view JEPA matters, \cref{fig:ablations} changes one setting at a time and scores the text JEPA against frozen DINOv2 (R@1 through a rotation fitted on validation pairs, last epoch, no selection on test pairs). With three seeds for every setting except the SIGReg weights, the larger learning rate and the 1{,}000-epoch run, every run that does not collapse lands between .18 and .28, far above the bad-view JEPA (.13) and next-token prediction (.12), which use the same recipe and were not tuned. Trimming matters most: whole captions as views lose a quarter of the alignment (.185 against .246), as trimming is what makes two descriptions of a scene share only their common content. Masking, the number of views, batch size, width and schedule move the mean by at most .016, about the spread of the default's own seeds (.230--.260). Stability is another matter. Without gradient clipping, 3 of the 36 runs with the default SIGReg weight collapse abruptly between epochs 20 and 60 of 200 (two of three seeds with 2 + 2 views, one with mask 30\%), and so does every other SIGReg weight we tried. With clipping, 2 + 2 views trains for all three seeds (.262) and so do $\lambda=0.005$ and $0.1$, while a larger learning rate and $\lambda=0.05$ still collapse; none of our longer runs, all trained with clipping and a lower learning rate, collapses (\cref{tab:sweep}). \emph{We thus recommend views drawn from different observations of the same world state, each trimmed (cropped, for images) and lightly masked, trained with gradient clipping, as the starting point for the modalities we tested.}

\subsection{Good Views Help on Both Sides}\label{sec:both}
\begin{figure}[t]
\centering
\includegraphics{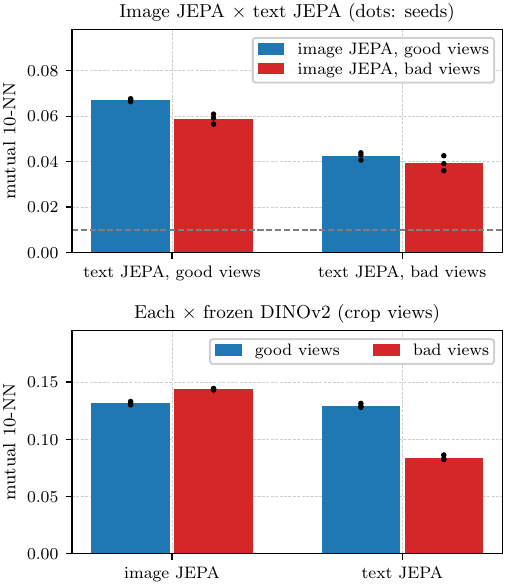}
\caption{\textbf{Good views help on both sides.} MSR-VTT, 1{,}000 test clips. Top: image JEPAs trained on different frames of a clip (good) or crops of one frame (bad), against text JEPAs trained on different captions (good) or spans of one caption (bad); dots are the three seeds, dashed is chance. Bottom: alignment of each with frozen DINOv2, whose own positives are crops. This demonstrates that \emph{good views on both sides align best in mutual 10-NN, and that DINOv2 prefers the image JEPA whose positives match its own.}}
\label{fig:both_sides}
\end{figure}

To test the image half of the prediction we need a world state observed several times in both modalities, which video provides. On MSR-VTT \citep{xu2016msrvtt} (7k training clips, 20 captions each) we train the same image and text JEPAs with views from different frames (images) or captions (text) of a clip (good), or from a single frame or caption (bad), and evaluate on the 1{,}000 test clips (middle frame and one caption; \cref{app:details}). Good views raise mutual 10-NN on each side, in every seed, the text side much more than the image side: .067 when both sides have good views, .059 with good text views only, .043 with good image views only and .039 with neither (same ordering in each seed; chance .01; \cref{fig:both_sides,tab:msrvtt}). Conversely, frozen DINOv2, whose positives are crops, agrees more with the image JEPA trained on crops than with the one trained on frames (.144 vs.\ .132, in every seed). That is, \emph{alignment follows agreement of positives}, as per \cref{thm:robust}.

\begin{figure*}[t]
\centering
\includegraphics{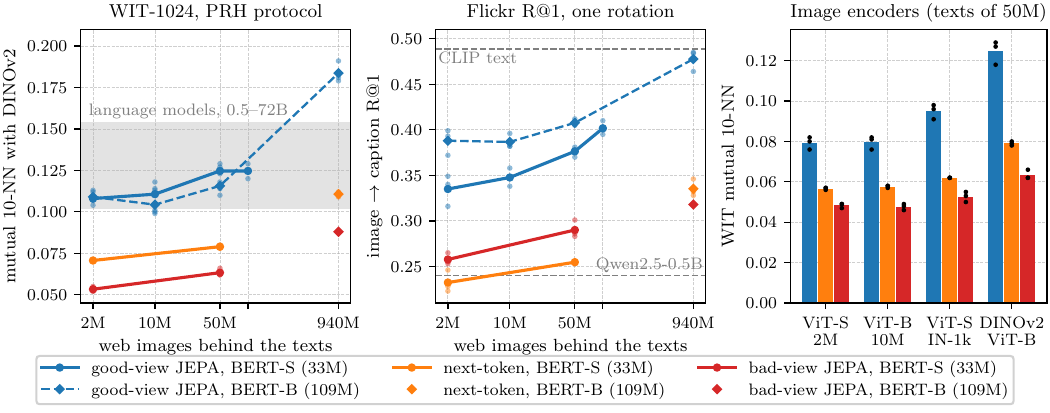}
\caption{\textbf{Good views scale.} Left and middle: text JEPAs trained on the texts of 2.16M to 940M web images (Recap-DataComp), never on a Flickr or COCO caption, against frozen DINOv2 ViT-B on WIT-1024 (PRH-paper protocol) and on Flickr (R@1 through one rotation fitted on 30k pairs). Solid lines and circles: BERT-S (33M); dashed lines and diamonds: BERT-B (109M), trained with next-token prediction and bad views on the texts of 940M images only, hence single diamonds there. Lines are means over seeds and faint dots the seeds (one seed for the bad-view JEPA at 940M). Shaded: the language models we evaluated (Qwen2.5 0.5--72B, Llama-3-8B and -70B, Mistral-7B; left); grey dashed: CLIP's text encoder and Qwen2.5-0.5B under the same protocol (middle). Right: image JEPAs trained from scratch (ViT-S on 2.16M and ViT-B on 10M Recap images, ViT-S on ImageNet-1k) and frozen DINOv2, each against the good-view, next-token and bad-view text JEPAs trained on the same texts of 50M web images (WIT; means of three seeds, dots: seeds). We clearly observe that \emph{good views beat bad views and next-token prediction at every scale and with every image encoder we compared, and that a 109M good-view JEPA on the texts of 940M images surpasses every language model we evaluated.}}
\label{fig:scaling}
\end{figure*}

\subsection{Good-View JEPAs Beat a Pretrained Language Model and Approach Pair-Trained Encoders}
Our absolute numbers so far are limited by the image encoder, a ViT-S trained from scratch on 29k images: training it five times longer raises its mutual 10-NN with the good-view text JEPA from .065 to .077 (seed 0), and with DINOv2 from .101 to .121. To isolate the text side, we pair each text encoder with a frozen DINOv2 ViT-B/14 \citep{oquab2024dinov2} under the same protocol (\cref{tab:calib}). With final checkpoints and no selection on test data, the good-view JEPA retrieves the true caption first for 24.6\% of test images, about twice the bad-view JEPA (.133), our next-token model (.118, or .137 with caption pairs) and the pretrained Qwen2.5-0.5B language model \citep{qwen2024qwen25} (.135, last layer). Text encoders trained on vastly more data reach .300 (all-mpnet-base-v2 \citep{reimers2019sentencebert,song2020mpnet}, a billion sentence pairs, Flickr and COCO captions among them) and .320 (CLIP's text encoder, 400M image--text pairs, \citealp{radford2021clip}), but our good-view JEPAs have the neighborhoods that agree most with DINOv2's (mutual 10-NN .204--.257 vs.\ .189 and .167), and the one trained on the texts of 940M web images also retrieves better than both (.349, seed 0). Retrieval is limited by how well the rotation is estimated: fitting the rotation on all 30k Flickr train and validation pairs instead of 1k, still the only part fitted on pairs, raises R@1 to .346 for the good-view JEPA (.363 with BERT-B and .400 for a BERT-B trained on Flickr and COCO captions, one seed each), against .21--.25 for the bad-view JEPA and next-token prediction and .48--.49 for the two reference encoders. Crucially, a good-view JEPA trained on web text alone closes most of that last gap (.464, \cref{sec:scaling}).

\begin{table}[t]
\centering
\small
\setlength{\tabcolsep}{3.5pt}
\caption{\textbf{Good-view JEPAs agree with DINOv2 more than the pretrained models we tested.} Same protocol with a frozen DINOv2 ViT-B/14. Top two rows and bottom two: pretrained models (last layer); the others are ours (BERT-S unless named). All entries use final checkpoints, except the early-stopped row (selected on a loss over the validation and test captions, which can only favor this baseline), with no selection on test data (mean $\pm$ half-range over three seeds where shown, seed 0 elsewhere). The last column fits the single map on all 30k Flickr train and validation pairs instead of 1k (--: we kept no epoch-29 checkpoint to embed the training captions with). No network is trained on pairs in either case. For reference, our Flickr image JEPA vs.\ DINOv2 (image vs.\ image) reaches mutual 10-NN .101 and R@1 .217. The rows above the first rule are trained on Flickr (and COCO) captions, so this comparison is in-domain for them and out-of-domain for most pretrained models (all-mpnet-base-v2, trained on a billion sentence pairs, saw Flickr and COCO captions); the Recap rows never saw a Flickr or COCO caption. CLIP's text encoder was trained with 400M images. Bold: best of ours. This demonstrates that \emph{good-view JEPAs have the neighborhoods closest to DINOv2's and, trained on web text alone, match or exceed the retrieval of text encoders trained on pairs.}}
\label{tab:calib}
\begin{tabular}{lccc}
\toprule
Text encoder, vs.\ frozen & mutual & \multicolumn{2}{c}{R@1, rotation fit on}\\
DINOv2 ViT-B/14 & 10-NN & 1k pairs & 30k pairs \\
\midrule
Qwen2.5-0.5B & .098 & .135 & .240 \\
BERT-base & .093 & .149 & .273 \\
bad-view JEPA & .132{\scriptsize$\pm$.001} & .133{\scriptsize$\pm$.007} & .207{\scriptsize$\pm$.008} \\
next-token & .122{\scriptsize$\pm$.001} & .118{\scriptsize$\pm$.002} & .213{\scriptsize$\pm$.004} \\
\quad early-stopped & .124{\scriptsize$\pm$.002} & .134{\scriptsize$\pm$.008} & -- \\
next-token, pairs & .132{\scriptsize$\pm$.002} & .137{\scriptsize$\pm$.004} & .247{\scriptsize$\pm$.016} \\
good-view JEPA & .232{\scriptsize$\pm$.006} & .246{\scriptsize$\pm$.015} & .346{\scriptsize$\pm$.007} \\
good-view JEPA, BERT-B & .245 & .273 & .363 \\
good, BERT-T, +COCO & .228 & .238 & .335 \\
good, +COCO & .228 & .256 & .367 \\
good, BERT-B, +COCO & .233 & .285 & .400 \\
good, BERT-L, +COCO & .227 & .269 & .376 \\
\midrule
\multicolumn{4}{l}{\emph{good-view JEPAs on Recap texts, no Flickr or COCO caption}}\\
texts of 2.16M images & .204{\scriptsize$\pm$.004} & .236{\scriptsize$\pm$.019} & .335{\scriptsize$\pm$.016} \\
\quad BERT-B & .214{\scriptsize$\pm$.003} & .272{\scriptsize$\pm$.010} & .388{\scriptsize$\pm$.014} \\
texts of 10M images & .205{\scriptsize$\pm$.004} & .243{\scriptsize$\pm$.004} & .348{\scriptsize$\pm$.010} \\
\quad BERT-B & .219{\scriptsize$\pm$.003} & .280{\scriptsize$\pm$.003} & .387{\scriptsize$\pm$.007} \\
texts of 50M images & .219{\scriptsize$\pm$.002} & .283{\scriptsize$\pm$.002} & .376{\scriptsize$\pm$.006} \\
\quad BERT-B & .236{\scriptsize$\pm$.002} & .300{\scriptsize$\pm$.011} & .408{\scriptsize$\pm$.003} \\
texts of 100M images & .229{\scriptsize$\pm$.001} & .309{\scriptsize$\pm$.007} & .402{\scriptsize$\pm$.007} \\
texts of 940M, BERT-B & \textbf{.257} & \textbf{.349} & \textbf{.464} \\
\midrule
all-mpnet-base-v2 & .189 & .300 & .481 \\
CLIP B/32 text & .167 & .320 & .489 \\
\bottomrule
\end{tabular}
\end{table}

\begin{figure*}[t]
\centering
\includegraphics{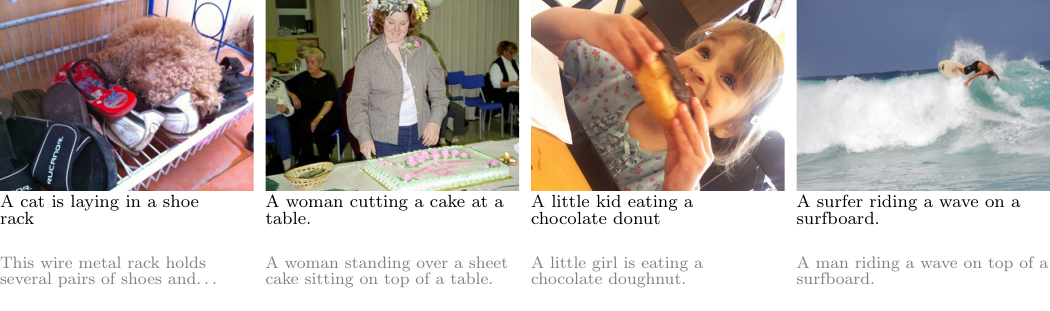}
\caption{\textbf{Captions without training a network on image--caption pairs.} Selected captions of COCO test images (first lines) with the decoder of our demo: frozen DINOv2 and DINOv3 image embeddings, one linear map fitted on the 30k Flickr pairs, and a language model fine-tuned on text only (consensus of 32 samples; \cref{tab:captioner_scale}), with one human caption (grey). That is, \emph{frozen encoders joined by one rotation describe scenes, objects and actions}, although only the rotation is fitted on image--caption pairs.}
\label{fig:captions}
\end{figure*}

\section{Next-Token Prediction Scales Poorly Compared to Good-View JEPAs}\label{sec:scaling}
Language models align with vision through scale \citep{huh2024platonic}. We now show that next-token prediction spends that scale poorly: on the same corpus, good views reach its alignment with 42 times fewer samples, and a 109M good-view JEPA that read 116B text tokens beats language models with up to 660 times more parameters, trained on 15--18T tokens, over 100 times more.

\subsection{Good Views Scale on Web Text}
Trained on more text, a text JEPA leaves the caption domain. We train the same 33M-parameter encoder on the texts of Recap-DataComp web images \citep{li2024recap}: each image's alt-text and two recaptions, written by an image captioner (LLaVA built on LLaMA-3-8B), are its views. We obtain three key observations (\cref{fig:scaling}). First, on the WIT-1024 benchmark of \citet{huh2024platonic}, on which none of our models is trained, its alignment with DINOv2 ViT-B rises from .034 when trained on Flickr captions to .108, .111 and .125 when trained on the texts of 2.16M, 10M and 50M images and stays at .125 with 100M (seeds differ by at most $\pm$.009). At 50M and 100M images this 33M-parameter encoder, trained without next-token or masked-token prediction, is level with Qwen2.5-3B and -7B (.125 and .118; intervals in \cref{app:results}). Second, views matter at both scales we tested: with the same encoder, texts and steps, the bad-view JEPA reaches .053 and next-token prediction .071, against .108 for the good-view JEPA, and with 23 times more images (1.9 times more steps) .063 and .079 against .125. Third, Flickr retrieval keeps rising with data, to .348, .376 and .402 with the texts of 10M, 50M and 100M images and .408 with BERT-B on 50M, level with our best in-domain model (.400) without a single Flickr or COCO caption: good views on web text carry most of the in-domain alignment. Width helps Flickr retrieval but not yet WIT at these scales: BERT-B reaches R@1 .388 and .408 with the texts of 2.16M and 50M images, but WIT .109 and .116, at or below BERT-S (.108 and .125), and with Flickr and COCO captions fixed, R@1 grows with the size of the text encoder up to BERT-B and then drops (.335, .367, .400 and .376 for 11M, 33M, 109M and 335M parameters; one seed).

\subsection{Good Views Need 42 Times Fewer Samples}
With the same 109M encoder and texts from the same corpus, the good-view JEPA reaches .109 on WIT after 21.6M training samples (the texts of 2.16M images), the alignment that next-token prediction reaches after 900M samples (.111, the texts of 940M images; we scored no earlier next-token checkpoint), and a higher Flickr R@1 (.388 against .335); each run uses the recipe of its scale (learning rate $10^{-4}$ and batch 256 against $2\cdot10^{-4}$ and 2{,}048). That is 42 times fewer samples, 14 times fewer text tokens since a good-view sample reads three texts, and 435 times fewer unique images (\cref{fig:teaser}, right). At matched samples the gap is larger still: .108 against .071 at 21.6M samples for the 33M encoder.

\subsection{A 109M Text JEPA Beats Language Models up to 72B Parameters}
Scaling data and model together changes the picture. We train the 109M-parameter BERT-B text JEPA with 900M samples from the texts of all 940M Recap-DataComp images (8 or 16 GPUs), and as controls next-token prediction and the bad-view JEPA (two masked copies of one text) with the same encoder, texts and samples. \Cref{tab:scale} gathers the comparison, from which we draw four conclusions.

\textbf{Good views beat every language model we evaluated.} On the PRH paper's benchmark, the good-view JEPA reaches .184 with DINOv2 ViT-B, above Qwen2.5 from 0.5B to 72B, Llama-3-8B and -70B and Mistral-7B, and it stays ahead with every DINOv2 size, with linear CKA in place of mutual 10-NN, and on the five times larger COCO benchmark (\cref{tab:llms,tab:coco}).

\textbf{The views make the gain, not the texts.} Next-token prediction and the bad-view JEPA trained on the same texts stay far below, and, as on Flickr captions, the scene explains more of the good-view embedding than of theirs (measured on the texts of 50M images, \cref{tab:nuisance}).

\textbf{The lead holds with every vision family.} Language models align best with CLIP's image encoders \citep{huh2024platonic}, yet with CLIP ViT-H/14 the good-view JEPA is ahead of Qwen2.5-72B and Llama-3-70B, and it leads every language model with MAE and I-JEPA as well (\cref{tab:vision_families}).

\textbf{It rivals text encoders trained on pairs.} It agrees with DINOv2 more than CLIP's text encoder (400M image--text pairs) on both benchmarks, more than all-mpnet-base-v2 (a billion sentence pairs, COCO captions among them) on WIT and as much on COCO, and it retrieves Flickr captions through one rotation almost as well as both. CLIP's text encoder is trained to agree with CLIP's own image encoder, not with DINOv2, so this compares agreement with an independently trained vision model.

Three caveats qualify these numbers. A fourth seed, trained with twice the effective SIGReg weight, reaches only .121 on WIT, after which we kept the first seed's weight for the others; all four seeds average .168, still above every language model (\cref{app:results}). The recaptioner that wrote most of these texts sees images through CLIP ViT-L/14 (\cref{sec:conclusion}). And at least 43 of the 1{,}024 WIT captions appear among the Recap-DataComp alt-texts, although removing them changes none of the nine text models we checked by more than .002. Still, our model has over 600 times fewer parameters than Qwen2.5-72B and never saw pixels: \emph{good views turn scale into alignment far more efficiently than next-token prediction}.

The image side matters as well. Image JEPAs trained from scratch without texts, on more images (all bad-view JEPAs: their positives are crops of one image), rank the text side the same way: with a ViT-S on the 2.16M Recap images, a ViT-B on 10M and a ViT-S on ImageNet-1k, the good-view text JEPA aligns best, next-token prediction second and the bad-view JEPA last (\cref{fig:scaling}, right). The ImageNet ViT-S, which sees the most samples, agrees most with DINOv2 and with our text JEPA, and with it two JEPAs trained separately from scratch agree about as much as the image JEPA agrees with Mistral-7B and Qwen2.5-7B (\cref{app:results}).

\begin{table}[t]
\centering
\small
\setlength{\tabcolsep}{1.8pt}
\caption{\textbf{At scale, good views beat language models up to 72B parameters.} Text encoders against frozen vision models under the PRH protocol: mutual 10-NN and unbiased linear CKA on WIT-1024 with DINOv2 ViT-B, mutual 10-NN on WIT-1024 with CLIP ViT-H/14 and on the 5{,}000 COCO test pairs with DINOv2 ViT-B, and Flickr R@1 through one rotation fitted on 30k pairs. Our three encoders are BERT-B models (109M) trained with 900M samples from the texts of 940M web images (one seed for the bad-view JEPA); --: not measured. More models in \cref{tab:llms,tab:coco,tab:vision_families}. We clearly observe that \emph{the good-view JEPA agrees with both vision models more than any other text encoder, while next-token prediction on the same texts falls far behind}.}
\label{tab:scale}
\begin{tabular}{lccccc}
\toprule
& \multicolumn{2}{c}{WIT, DINOv2-B} & WIT & COCO & Flickr\\
Text encoder & 10-NN & CKA & CLIP-H & 10-NN & R@1\\
\midrule
good-view JEPA & \textbf{.184} & \textbf{.507} & \textbf{.230} & \textbf{.161} & .478\\
next-token & .111 & -- & .146 & -- & .335\\
bad-view JEPA & .088 & -- & .116 & -- & .318\\
\midrule
Qwen2.5-72B & .154 & .418 & .210 & .108 & --\\
Llama-3-70B & .151 & .380 & .206 & .113 & --\\
\midrule
CLIP B/32 text & .124 & .339 & .159 & .132 & \textbf{.489}\\
all-mpnet-base-v2 & .118 & .390 & .159 & .156 & .481\\
\bottomrule
\end{tabular}
\end{table}

\begin{figure}[t]
\centering
\includegraphics{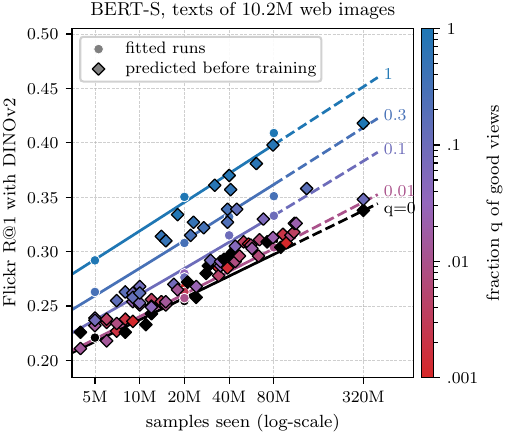}
\caption{\textbf{A scaling law of alignment in good views and samples seen.} Flickr test R@1 (DINOv2 ViT-B, 30k-pair rotation) of BERT-S text JEPAs trained on the texts of 10.2M web images, against the samples seen. Each point is a run, colored by its fraction $q$ of training items with good views (black: $q=0$, bad views only). Circles: the grid the law is fitted on; diamonds: runs predicted before training. Lines: the law at five values of $q$, dashed beyond the fitted range. We observe that \emph{R@1 grows linearly in the logarithm of the samples seen at every fraction of good views, faster with more of them}, and that the law predicts unseen runs within seed noise up to twice its range.}
\label{fig:phase_law}
\end{figure}

\subsection{A Scaling Law for the PRH}\label{sec:law}
Scaling laws turned the training of language models from trial and error into prediction; we now show that alignment admits one too. With one recipe (BERT-S on the texts of Recap-DataComp images; \cref{app:results}), each training item draws good views (two texts of its image) with probability $q$ and bad views (two corruptions of one text) otherwise, on the texts of $N$ images for $M$ samples. A bad-view floor rising with $\log M$ plus a good-view term,
\begin{equation}
\mathrm{R@1}=a+d\log M+b\log\big(1+q^{\alpha}M^{\beta}N^{\gamma}/c\big),
\end{equation}
fits the 45 cells of a grid up to 80M samples with RMSE .007 (\cref{fig:phase_law}). We then predicted 85 runs at random $(q,M)$ on 10.2M images, with $M$ from 3M to 133M samples, before training them: it predicts their Flickr R@1 with RMSE .0093, while two seeds of one run differ by .0096 (pooled over six cells and three seeds each). That is, \emph{the law predicts a run it has never seen as well as rerunning that run with another seed}. Along its iso-alignment lines, more samples substitute for good views sub-linearly ($q\propto M^{-0.30}$), and unique images enter weakly ($\gamma=.19$): without good views, WIT stays within .054--.057 from the texts of 0.41M to 50.9M images, while with them it rises from .090 to .115. The law fails further out, for good views only: at 320M samples, four times its longest fitted run and 31 passes over the images, it overestimates the runs with good views (R@1 .418 against .455 predicted at $q=1$, .348 against .389 at $q=.1$), while the run without matches (.338 against .337); it has no term for repeated data. Its exponents are poorly identified, and it is an empirical summary of one recipe, not a consequence of \cref{thm:coverage}.

\section{Platonic Representations as a Tool}\label{sec:tool}
Alignment up to one rotation lets separately trained encoders be composed. We show two uses: a captioner whose only part fitted on pairs is one rotation, and alignment with no pair at all.

\subsection{Captioning Without Training a Network on Image--Caption Pairs}\label{sec:caption}
Alignment up to one rotation makes a new kind of captioner possible. We train a caption decoder on text alone: a causal transformer reconstructs each training caption from the frozen good-view JEPA embedding of that caption, with Gaussian noise added to the embedding so that it tolerates alignment error \citep{nukrai2022capdec}. To caption an image, we embed it with the frozen DINOv2, apply the rotation and decode. That is, the rotation is the only part fitted on image--caption pairs, and no network is trained on them. Crucially, the noise level is matched to the alignment error rather than tuned: after the rotation, an image and its caption have mean cosine $c\approx.3$ (.29 on held-out validation pairs, .33 on test pairs), and Gaussian noise whose norm is $\sigma$ times the embedding's leaves an expected cosine of $(1+\sigma^2)^{-1/2}$ with it, so matching the two gives $\sigma=(c^{-2}-1)^{1/2}\approx 3$; we use $\sigma=3$, and transfer peaks there (\cref{fig:captioner}). We obtain a few key observations. First, on the 1{,}000 Flickr test images the captioner reaches CIDEr .396 (on a 0--1 scale throughout; most prior work reports it times 100) against five human captions, nine times a random rotation and 2.5 times retrieving the nearest caption through the same rotation (\cref{tab:caption}), and .431 with the rotation fitted on all 30k Flickr pairs. Second, rotations fitted on fewer pairs than dimensions are underdetermined (\cref{prop:ident}(ii)) but improve when refined on unpaired images and captions (\cref{fig:captioner}): from 30 pairs, the captioner (.169) matches caption retrieval with 1{,}014 pairs (.158). Third, the captioner transfers across datasets: on the 5{,}000 COCO Karpathy test images, with the rotation fitted on Flickr pairs only and the decoder trained on Flickr and COCO training captions, it reaches CIDEr .668 and BLEU-4 .199 with a beam of three (.630 greedy), against .038 for a random rotation and .169 for retrieving the nearest Flickr caption. Captioners built on CLIP's jointly trained encoders, and hence on 400M image--text pairs, report .345 (ZeroCap, \citealp{tewel2022zerocap}, no training), .493 (MAGIC, \citealp{su2022magic}) and .918 and .912 (CapDec and DeCap, \citealp{nukrai2022capdec,li2023decap}); unlike these, ours uses no encoder trained on pairs but fits its rotation on in-domain Flickr pairs, which they do not use. The rotation is also what makes the captioner transfer: replacing it by maps trained on the same 30k Flickr pairs (\cref{tab:captioner_scale}), an unconstrained linear map gains a little on Flickr (CIDEr .531 against .504, web-pretrained decoder, beam 3) but loses a fifth on COCO (.635 against .802), and a two-layer MLP loses on both (.498 and .585); with consensus decoding, the same maps reach .561 and .511 on Flickr but .652 and .597 on COCO, against .538 and .832 for the rotation. That is, \emph{a map free to fit the pairs fits Flickr, while a rotation, which can only turn one space onto the other, carries over to new images}, as expected if the two spaces agree up to a rotation.

Lastly, the captioner improves further without pairs (\cref{tab:captioner_scale}): with a pretrained language model fine-tuned on text alone as the decoder (Qwen2.5-1.5B, \citealp{qwen2024qwen25}), the text JEPA trained on the texts of 100M web images, DINOv2 ViT-g and DINOv3 ViT-L \citep{simeoni2025dinov3} as frozen image encoders, and the consensus of 32 sampled captions \citep{kumar2004mbr}, a single map fitted on the same 30k Flickr pairs gives CIDEr .519 on Flickr and .838 on COCO, and .538 and .832 with the decoder of our demo, first trained on web sentences (\cref{fig:captions}). \emph{We thus recommend matching the decoder's noise to the residual cosine of the rotation, $\sigma=(c^{-2}-1)^{1/2}$, rather than tuning it.}

\textbf{From photos to videos.} Neither the encoders nor the decoder ever saw a video, yet the same captioner describes one from the mean of its frames' rotated embeddings. On the 2{,}990 MSR-VTT test clips \citep{xu2016msrvtt}, with the rotation fitted on the 30k Flickr pairs and the mean of eight frames, it reaches CIDEr .246 against the 20 human captions of each clip, above the zero-shot captioners whose decoder is trained on image captions or general text alone, all built on CLIP's jointly trained encoders (.058 for ZeroCap, .074 for MAGIC and .150--.186 for DeCap, \citealp{li2023decap}), though below video--language systems such as VideoCoCa (.271, 144M video--text pairs, \citealp{yan2022videococa}) and TOPA (.329, CLIP features and LLM-written textual videos, \citealp{li2024topa}), and below decoders trained on MSR-VTT captions (.348). Averaging frames helps both captions and alignment, as \cref{thm:nuisance} suggests: the frames of a clip are several observations of one scene, and their mean averages out what is particular to each frame. On the 1{,}000 clips of the 1k-A split, it raises the fit-free mutual 10-NN between clips and captions from .148 (middle frame) to .167, and the same rotation, fitted on photos, retrieves the right clip first for 16.5\% of captions, against 31.2\% for CLIP ViT-B/32 trained on 400M image--text pairs \citep{luo2022clip4clip} (\cref{tab:video}). That is, \emph{the rotation fitted on photos carries over to video}.

\subsection{Alignment Without Any Pair}\label{sec:pairfree}
\Cref{prop:ident} says that the first two moments of unpaired data cannot find the rotation, but the geometry of a finite world can: \citet{schnaus2026rosetta} recover it with no pair at all, by matching clusters of the two clouds and refining an orthogonal map by optimal transport. The theory predicts that this works best when the two encoders' geometries agree, i.e., with good views. We test it with their method, code, hyperparameters and five seeds unchanged, on their protocol: DINOv2 ViT-B/14 against a text encoder, the map fitted on disjoint unpaired halves of COCO train2014 (images from one half, captions from the other), and scored on the 40{,}504 val2014 pairs by FOSCTTM (the fraction of samples closer than the true match; 0 is perfect, .5 chance) and by zero-shot classification (\cref{tab:pairfree}). Our setup reproduces their numbers for all-mpnet-base-v2. Our 109M good-view JEPA (three training seeds, five method seeds each) reaches FOSCTTM .018 and zero-shot accuracies of 86.0\% (CIFAR-10 top-1), 67.2\% (CIFAR-100 top-5) and 64.4\% (ImageNet-100 top-5): above every zero-shot accuracy of their paper, over 21 pairs of seven vision and three language models, up to DINOv3-7B and Qwen3-8B, and five settings (at most 82\%, 62\% and 56\%), and level with their best FOSCTTM on these data (.02; .01 with captions from another dataset). Rerun under identical conditions, their strongest text encoder, Qwen3-8B, reaches .018, 83.4\%, 61.7\% and 47.6\%: our 70 times smaller JEPA matches it on FOSCTTM and CIFAR-10 and beats it by 5.5 points on CIFAR-100 and 16.8 on ImageNet-100 (Mann--Whitney $p=.003$ and $p<.001$). It even matches CLIP's text encoder, trained on 400M image--text pairs, on both CIFARs, though not on FOSCTTM (.006) or ImageNet-100 (68.7\%). Next-token prediction on the same texts fails in half of its runs (FOSCTTM .287, 25.5\% on CIFAR-10), while our bad-view JEPA aligns well too (.026), so on COCO the contrast is between JEPAs and next-token prediction. On Flickr, whose unpaired pools are smaller (15{,}007 images and the captions of 15{,}007 others), only good-view JEPAs leave chance: with no pair, the 109M good-view JEPA retrieves the right caption first for 16.4\% $\pm$ 10.4\% of test images (median 16.1\%; chance 0.1\%), against at most 0.3\% for CLIP's text encoder, all-mpnet-base-v2, BERT, Qwen2.5-0.5B, and our bad-view JEPA and next-token models trained on the same texts (\cref{app:results}). The method's own objective on the unpaired pools, which uses no test data, picks the best of its five seeds in all six good-view JEPA cases (R@1 13.5--36.2\%). Through such a map, a captioner fitted with zero pairs reaches CIDEr .158 $\pm$ .042 on Flickr (.208 for the map the objective selects), against .430 with the 30k-pair rotation and .028 for a random map. That is, \emph{good views make two modalities alignable from their geometry alone}, which next-token prediction does not.

\begin{table}[t]
\centering
\small
\setlength{\tabcolsep}{1.7pt}
\caption{\textbf{A 109M good-view JEPA beats every pair-free zero-shot accuracy of \citet{schnaus2026rosetta}.} The protocol and method of \citet{schnaus2026rosetta}, unchanged: DINOv2 ViT-B/14 and a text encoder aligned with zero pairs on unpaired halves of COCO train2014, scored on the 40{,}504 val2014 pairs (FOSCTTM, lower is better) and by zero-shot top-1 on CIFAR-10 and top-5 on CIFAR-100 and ImageNet-100 (\%). Means over training seeds times the method's five seeds (std in \cref{app:results}). Their best: the best number of their paper for each metric over its 21 vision--language pairs, on these data and in any of its five settings (\texttt{paper/data/schnaus2026\_reported.csv}); our reruns of their MPNet pair (all-mpnet-base-v2) reproduce their reported .04, 70, 46 and 45. Last row: the ceiling of Procrustes on all training pairs, range over all encoders. Our good-view JEPAs are trained on the texts of 940M and 100M web images. We clearly observe that \emph{a 109M good-view JEPA beats every zero-shot accuracy of \citet{schnaus2026rosetta}, matches their best FOSCTTM on these data, and matches or significantly beats their 8B language model under identical conditions}, while next-token prediction on the same texts fails; only CLIP's text encoder, trained on 400M image--text pairs, aligns better on FOSCTTM and ImageNet-100.}
\label{tab:pairfree}
\begin{tabular}{lcccc}
\toprule
Text encoder (runs) & FOSCTTM & C-10 & C-100 & IN-100\\
\midrule
good-view JEPA, 109M (15) & \textbf{.018} & \textbf{86.0} & \textbf{67.2} & \textbf{64.4}\\
good-view JEPA, 33M (15) & .036 & 83.2 & 60.9 & 57.9\\
bad-view JEPA, 109M (5) & .026 & 78.1 & 55.3 & 46.7\\
next-token, 109M (15) & .287 & 25.5 & 19.9 & 20.3\\
\midrule
Qwen3-8B, rerun (5) & \textbf{.018} & 83.4 & 61.7 & 47.6\\
MPNet, rerun (5) & .043 & 70.2 & 45.6 & 45.4\\
their best, these data & .02 & 80 & 58 & 46\\
their best, any setting & .01 & 82 & 62 & 56\\
\midrule
CLIP B/32 text (5) & .006 & 84.0 & 67.0 & 68.7\\
82{,}783 pairs & .002--.005 & 83--92 & 65--85 & 57--88\\
\bottomrule
\end{tabular}
\end{table}

\section{Related Work}\label{sec:related}
The PRH was formulated by \citet{huh2024platonic}, whose contrastive argument assumes bijective observations and co-occurrence positives. They note that it fails for lossy observations, which remain open; \cref{thm:robust} addresses a different departure, mismatched positives. \citet{ziyin2025perfect} prove perfect alignment for deep linear networks trained by SGD, and \citet{doshi2026theory} explain why translations align inside multilingual language models by a hierarchical structure that languages share beneath surface features of their own, a within-modality analogue of our world states and nuisances. Identifiability results show that contrastive and augmentation-based learning recover latent variables up to simple transformations \citep{zimmermann2021contrastive,vonkugelgen2021self}, including across modalities when trained jointly \citep{daunhawer2023identifiability}; we instead train modalities separately and characterize when they agree. The spectral view of self-supervised learning \citep{haochen2021provable,balestriero2022contrastive,belkin2003laplacian} underlies \cref{prop:spectral}. How views and data shape what one encoder learns has been analyzed through augmentation graphs and their overlaps \citep{haochen2021provable,wang2022chaos}, the inductive bias of the function class \citep{saunshi2022understanding,kiani2022joint,cabannes2023ssl}, the eigenfunctions of the positive-pair kernel \citep{johnson2023contrastive} and multi-view redundancy \citep{tosh2021contrastive}; \cref{thm:coverage} is an analogue with world states in place of classes, and its part (ii) the cross-modal consequence. The density view of JEPA anti-collapse \citep{balestriero2025gaussian,balestriero2025lejepa} underlies our reading of next-token prediction. Aligning separately trained embedding spaces with few or no pairs goes back to word translation, through orthogonal Procrustes and unsupervised matching \citep{conneau2018word,alvarezmelis2018gromov}, and extends to relative representations \citep{moschella2023relative}, training-free multimodal coupling from a few anchor pairs \citep{norelli2023asif}, translation between text-embedding spaces without pairs \citep{jha2025vec2vec}, vision--language matching without parallel data \citep{schnaus2025blind} and, most recently, cross-modal alignment with no pairs at all from shared geometry \citep{schnaus2026rosetta}; unimodal vision and language encoders can be matched with little or no paired data \citep{maniparambil2024vision}. These works align given encoders; we ask which training objective makes encoders alignable, and our Procrustes map is the simplest of these tools. Text-only captioners decode the embeddings of jointly trained encoders \citep{nukrai2022capdec,li2023decap,tewel2022zerocap,su2022magic}, a linear map trained on pairs can feed image features to a frozen language model \citep{merullo2023linearly}, and unpaired captioning trains adversarially or through shared detector concepts \citep{feng2019unsupervised,laina2019towards}; ours needs no joint training and uses pairs only to fit one map with orthonormal rows.

\section{Conclusion}\label{sec:conclusion}
We have turned the Platonic Representation Hypothesis into a Platonic Representation Theorem for JEPAs which, for population optima on injective renderings of one world and objectives whose minimizers are unique up to rotation, reduces it to one requirement for any number of modalities: they must agree on which world states are positives. We further proved that positives which corrupt a single observation make an encoder spend dimensions on its modality's nuisance, which no rotation can align, and showed that next-token prediction has the form of such a bad-view JEPA. Empirically, good views align best across data, encoders and modalities, follow a scaling law that predicts unseen runs within seed noise, need 42 times fewer samples than next-token prediction on the same corpus, and give a 109M text JEPA that beats every language model we evaluated, up to 72B parameters, on the PRH paper's own benchmark, and a captioner whose only parameters fitted on image--caption pairs are those of one rotation, which unpaired data alone can find well enough to beat every pair-free zero-shot classification reported with language models up to 8B parameters.

\textbf{Known limitations.} Our theorems are population statements over a finite world with injective renderings and matched marginals; real modalities are lossy, and finite-sample training relies on the network's inductive bias. Uniqueness up to rotation is proved for second-order anti-collapse, not for LeJEPA's full-distribution constraint, which a finite world cannot meet exactly; \cref{thm:robust} is proved for spectral solutions only, and no experiment meets the hypotheses of \cref{thm:exact} exactly, since our runs use LeJEPA and the image side mostly uses crops as positives. The main caveat is where good views come from: they require knowing which texts describe the same scene, the grouping of human captions by image on Flickr and COCO, while on Recap-DataComp two of each image's three texts are generated by an image captioner (LLaVA built on LLaMA-3-8B, whose vision encoder is CLIP ViT-L/14, \citealp{li2024recap}) and carry information distilled from image--text training. Without any image captioner, good views on the alt-text and a text-only paraphrase of it reach .060 on WIT against .107 with the recaptions (\cref{tab:sources}), so the comparisons with language models and CLIP's text encoder at scale rely on these texts, while the comparisons between views, on identical texts, do not. Paired information thus enters through the definition of positives, one fitted linear map and, on Recap, captioner-generated texts, never through a cross-modal loss. The bad-view JEPA and next-token prediction see exactly the same texts as the good-view JEPA. Outside MSR-VTT our image encoders are bad-view JEPAs by our own definition, and hyperparameters were tuned only for the good-view JEPA.

Critically, our study provides what the PRH has long needed: a provable sufficient condition that directly informs how to train for it, and evidence that next-token prediction, the objective behind today's largest models, is an inefficient way to get there. We hope that our study will establish positive views, and not scale alone, as the lever to build platonic representations, and open new avenues for modalities beyond images and text, e.g., audio, video and actions.

\section*{Acknowledgments}
We thank Kerda Varaku, Lukas Kuhn, Hazel \mbox{(Heejeong)} Nam and Marcin Przewi\k{e}\'zlikowski for their helpful feedback and discussions.

\clearpage  % every float of the main text before the references
\bibliographystyle{plainnat}
\bibliography{references}

\clearpage
\onecolumn
\appendix\appendixparttrue

\section{Proofs}\label{app:proofs}

We prove the results in the order of \cref{sec:theory}, one subsection per subsection of the main text. \Cref{prop:goodbad,thm:coverage} are stated informally in the main text; their formal statements precede their proofs.

\subsection{Notation and Objectives}

\paragraph{Notation.} On a finite world, the positive-pair distribution $\pi$ defines the Markov operator $(T_\pi h)(z)=\E_\pi[h(z')\mid z]$, i.e., the matrix $\mD^{-1}\boldsymbol\Pi$ acting on functions $h:\Z\to\R$. It is self-adjoint in $L^2(p)$, with $\langle h,h'\rangle_p=\sum_zp(z)h(z)h'(z)$, and $h\mapsto\mD^{1/2}h$ maps it to $\mK_\pi=\mD^{-1/2}\boldsymbol\Pi\mD^{-1/2}$: the two share the eigenvalues $\mu_i$, and the eigenfunctions $\phi_i=\mD^{-1/2}\boldsymbol v_i$ of $T_\pi$ are orthonormal in $L^2(p)$, with $\phi_0=\mathbf 1$, so that the spectral embedding of \cref{prop:spectral} is $\vPhi_\pi=(\phi_1,\dots,\phi_k)^\top$. The operator and Hilbert--Schmidt norms of $T_{\pi_1}-T_{\pi_2}$ on $L^2(p)$ are the spectral and Frobenius norms of $\mK_{\pi_1}-\mK_{\pi_2}$, with $\|\mK_{\pi_1}-\mK_{\pi_2}\|_F^2=\sum_{z,z'}\big(\pi_1(z,z')-\pi_2(z,z')\big)^2/\big(p(z)p(z')\big)$. On world states, the objective of \eqref{eq:jepa} is $\mathcal L_\pi(h)=\E_{(z,z')\sim\pi}\ell\big(h(z),h(z')\big)+\Omega(h)$, where $\Omega(h)$ depends on $h$ only through the distribution of $h(z)$, $z\sim p$.

\paragraph{Objectives of the form \eqref{eq:jepa}.} Spectral contrastive learning, whose minimizers are rescaled eigenfunctions, and whitened MSE fit this form \citep{haochen2021provable,ermolov2021whitening}; VICReg (with the mean fixed) and Barlow Twins (with $\lambda=1$) do only in a whitened hard-constraint limit, where their minimizers are those of \cref{prop:spectral} and, both being translation invariant, agree only after centering \citep{bardes2022vicreg,zbontar2021barlow,balestriero2022contrastive}; LeJEPA's SIGReg term is rotation invariant in expectation over its projection directions \citep{balestriero2025lejepa}.

\subsection{The Platonic Representation Theorem (\texorpdfstring{\cref{sec:thm}}{Section 2.1})}

\begin{lemma}[Renderings Only Change Coordinates]\label{lem:cov}
If $g_m$ is injective, $f\mapsto f\circ g_m$ is a bijection between encoders on $g_m(\Z)$ and functions $\Z\to\R^k$, and $\mathcal L_m(f)=\mathcal L_{\pi_m}(f\circ g_m)$.
\end{lemma}

\begin{proof}[Proof of \cref{lem:cov}]
Since $g_m$ is injective it has a left inverse $g_m^{-1}$ on $g_m(\Z)$, and $h\mapsto h\circ g_m^{-1}$ inverts $f\mapsto f\circ g_m$. For the objective, the pair $(g_m(z),g_m(z'))$ with $(z,z')\sim\pi_m$ is mapped by $f$ to $(h(z),h(z'))$ with $h=f\circ g_m$, and the embeddings $f(x)$ of the data $x=g_m(z)$, $z\sim p$, have the distribution of $h(z)$. Hence $\mathcal L_m(f)=\mathcal L_{\pi_m}(f\circ g_m)$.
\end{proof}

\begin{proof}[Proof of \cref{thm:exact}]
By \cref{lem:cov}, $f$ minimizes $\mathcal L_m$ if and only if $f\circ g_m$ minimizes $\mathcal L_\pi$, for every modality $m$. The minimizers of $\mathcal L_\pi$ lie in one rotation orbit $\{\mQ h^\star:\mQ\in\Oset\}$; hence $f_m^\star\circ g_m=\mQ_mh^\star$ for some $\mQ_m\in\Oset$, for every $m$, and $f_m^\star(g_m(z))=\mQ_m\mQ_{m'}^\top f_{m'}^\star(g_{m'}(z))$ for every $z$ and every two modalities $m,m'$.
\end{proof}

\begin{proof}[Proof of \cref{prop:spectral}]
Write $h=(h_1,\dots,h_k)$. The constraints say $\langle h_i,h_j\rangle_p=\delta_{ij}$ and $\langle h_i,\mathbf 1\rangle_p=0$. Expanding, and using that both marginals of $\pi$ equal $p$,
\[
\E_\pi\|h(z)-h(z')\|^2=\sum_{j=1}^k\big(2\E_p h_j^2-2\E_\pi h_j(z)h_j(z')\big)=2k-2\sum_{j=1}^k\langle h_j,T_\pi h_j\rangle_p .
\]
$T_\pi$ is self-adjoint on $L^2(p)$ and leaves $\mathbf 1^\perp$ invariant, with eigenpairs $(\mu_i,\phi_i)_{i\ge1}$ there. By Ky Fan's maximum principle \citep{fan1949theorem}, $\sum_j\langle h_j,T_\pi h_j\rangle_p\le\sum_{i=1}^k\mu_i$ over orthonormal families in $\mathbf 1^\perp$, with equality if and only if $\mathrm{span}(h_1,\dots,h_k)$ is spanned by eigenfunctions for the $k$ largest eigenvalues. When $\mu_k>\mu_{k+1}$ that space is $\mathrm{span}(\phi_1,\dots,\phi_k)$, and two orthonormal bases of the same $k$-dimensional space differ by an orthogonal matrix, so $h=\mQ\vPhi_\pi$. Conversely every $\mQ\vPhi_\pi$ is feasible and attains $2\sum_{i\le k}(1-\mu_i)$.
\end{proof}

\subsection{When Positives Differ (\texorpdfstring{\cref{sec:differ}}{Section 2.2})}

\begin{proof}[Proof of \cref{thm:robust}]
Write $\mK_m=\mK_{\pi_m}$. The map $h\mapsto\mD^{1/2}h$ is an isometry from $L^2(p)$ to $\R^{|\Z|}$ under which $T_{\pi_m}$ becomes $\mK_m$. Both matrices have the same top eigenvector $\mD^{1/2}\mathbf 1$ with eigenvalue 1, because the marginals coincide, and both leave its orthogonal complement $H$ invariant. On $H$, let $\boldsymbol V_m$ be the matrix whose columns are orthonormal eigenvectors of $\mK_m$ for its $k$ largest eigenvalues; then the row of $\boldsymbol V_m$ at $z$ is $\sqrt{p(z)}\,\vPhi_m(z)^\top$, and $\E_p\|\vPhi_1(z)-\mQ\vPhi_2(z)\|^2=\|\boldsymbol V_1-\boldsymbol V_2\mQ^\top\|_F^2$. The variant of the Davis--Kahan theorem of \citet[Theorem 2]{yu2015useful}, applied on $H$ with $\Sigma=\mK_2|_H$, $\hat\Sigma=\mK_1|_H$, $r=1$, $s=k$ (so the gap is $\delta=\mu_k-\mu_{k+1}$ of $\mK_2$), gives an orthogonal $\hat{\boldsymbol O}$ with
\[
\|\boldsymbol V_1\hat{\boldsymbol O}-\boldsymbol V_2\|_F\le\frac{2^{3/2}\min\big(\sqrt k\,\|\mK_1-\mK_2\|_2,\|\mK_1-\mK_2\|_F\big)}{\delta},
\]
which is the bound with $\mQ=\hat{\boldsymbol O}$, since $\|\boldsymbol V_1\hat{\boldsymbol O}-\boldsymbol V_2\|_F=\|\boldsymbol V_1-\boldsymbol V_2\hat{\boldsymbol O}^\top\|_F$. The restriction to $H$ leaves both norms unchanged, because $(\mK_1-\mK_2)\mD^{1/2}\mathbf 1=0$.
\end{proof}

\begin{proof}[Proof of \cref{prop:ident}]
Let $u=f_2^\star\circ g_2$, so that $f_1^\star\circ g_1=\mQ u$ and $\E_p uu^\top=\mI_k$. (i) $\E_p(\boldsymbol Ru)(\boldsymbol Ru)^\top=\boldsymbol R\,(\E_p uu^\top)\boldsymbol R^\top=\mI_k$ for every $\boldsymbol R\in\Oset$, and the first moments are zero for every $\boldsymbol R$. Beyond the finite world: if the embedding distribution $\kappa$ is rotation invariant, as SIGReg's target $\mathcal N(0,\mI_k)$ in the limit of a continuum of world states, then $\boldsymbol R_\#\kappa=\kappa$ for every $\boldsymbol R$ and no statistic of unpaired samples identifies $\mQ$; on a finite world, the distribution of $u(z)$ is finitely supported and cannot be rotation invariant for $k\ge2$, which is why unpaired data can refine a map in practice, and why the shared geometry of a finite world can reveal it \citep{schnaus2026rosetta}. (ii) Write $\boldsymbol X=[u(z_1),\dots,u(z_n)]$ and $\boldsymbol Y=\mQ\boldsymbol X$. The objective is $\|\boldsymbol Y-\boldsymbol R\boldsymbol X\|_F^2\ge0$, with equality at $\boldsymbol R=\mQ$. If $\boldsymbol R\boldsymbol X=\mQ\boldsymbol X$ then $(\boldsymbol R-\mQ)\boldsymbol X=0$, and since $\boldsymbol X$ has rank $k$, $\boldsymbol R=\mQ$.
\end{proof}

\subsection{Good-View and Bad-View JEPAs (\texorpdfstring{\cref{sec:goodbad}}{Section 2.3})}

\begin{proof}[Proof of \cref{thm:nuisance}]
Write $x_1=g_1(z,\nu_1)$ and $x_2=g_2(z,\nu_2)$ for a paired observation, with $\nu_1,\nu_2$ independent of each other and of $z$. Since $f_1$ and $f_2$ are centered and whitened, for every $\mQ\in\Oset$, $\E\|\mQ f_1(x_1)-f_2(x_2)\|^2=2k-2\,\mathrm{tr}(\mQ\boldsymbol\Sigma)$ with $\boldsymbol\Sigma=\E[f_1(x_1)f_2(x_2)^\top]$, and the minimum over $\mQ$ is $2k-2\|\boldsymbol\Sigma\|_*$ (von Neumann's trace inequality, attained by orthogonal Procrustes). Given $z$, $x_1$ and $x_2$ are independent, so $\boldsymbol\Sigma=\E[u(z)w(z)^\top]$ with $u=\E[f_1(x_1)\mid z]$ and $w=\E[f_2(x_2)\mid z]$, both centered, with covariances $\mC_1=\mC_{f_1}$ and $\mC_2=\mC_{f_2}$. By the law of total covariance $\mC_2=\mI_k-\E[\mathrm{Cov}(f_2(x_2)\mid z)]\preceq\mI_k$, and likewise $\mC_1\preceq\mI_k$. Let $\tilde u=(\mC_1^{1/2})^+u$ and $\tilde w=(\mC_2^{1/2})^+w$ (pseudo-inverses). Since $\E(a^\top u)^2=a^\top\mC_1a=0$ for $a\in\ker\mC_1$, $u$ lies in the range of $\mC_1$ almost surely, so $u=\mC_1^{1/2}\tilde u$, and likewise $w=\mC_2^{1/2}\tilde w$, with $\E\tilde u\tilde u^\top$ and $\E\tilde w\tilde w^\top$ the orthogonal projections onto the ranges of $\mC_1$ and $\mC_2$; then $\boldsymbol\Sigma=\mC_1^{1/2}\boldsymbol R\,\mC_2^{1/2}$ with $\boldsymbol R=\E[\tilde u\tilde w^\top]$, and $\|\boldsymbol R\|_\mathrm{op}\le1$ by Cauchy--Schwarz. By H\"older's inequality for Schatten norms, $\|\boldsymbol\Sigma\|_*\le\|\mC_1^{1/2}\|_*\,\|\boldsymbol R\|_\mathrm{op}\,\|\mC_2^{1/2}\|_\mathrm{op}\le\mathrm{tr}\,\mC_1^{1/2}=k\rho(f_1)$, which bounds the error below by $2k(1-\rho(f_1))$; exchanging the roles of $f_1$ and $f_2$ bounds it below by $2k(1-\rho(f_2))$ as well, hence by $2k(1-\min_m\rho(f_m))$. Since $0\preceq\mC_1\preceq\mI_k$, $\rho(f_1)=1$ if and only if $\mC_1=\mI_k$, i.e., $\E[\mathrm{Cov}(f_1(x_1)\mid z)]=0$, i.e., $f_1(x_1)$ is almost surely a function of $z$.
\end{proof}

\begin{tcolorbox}[result,colframe=blue!50!black,colback=blue!2,title={Proposition~\ref*{prop:goodbad} (Formal): Good Views Ignore Nuisance, Bad Views Capture It}]
Modality $m$ has a finite nuisance set $\mathcal N_m$ with distribution $q_m$, $q_m(\nu)>0$, the state and the two nuisances being mutually independent, and renders $(z,\nu)$ injectively as $x=g_m(z,\nu)$; a paired image and caption are $(g_1(z,\nu_1),g_2(z,\nu_2))$ with $\nu_1,\nu_2$ independent. A good-view JEPA's positives are $(g_m(z,\nu),g_m(z',\nu'))$ with $(z,z')\sim\pi$ and $\nu,\nu'\sim q_m$ independent; a bad-view JEPA's are two corruptions $\tilde x,\tilde x'\sim\Gamma_m(\cdot\mid x)$ of one observation $x$, which share its nuisance. Take the objective of \cref{prop:spectral}. (i) A good-view JEPA's kernel is $\boldsymbol\Pi\otimes q_mq_m^\top$; if $\mu_k>\max(\mu_{k+1},0)$, its minimizers are $\mQ\vPhi_\pi(z)$, functions of the state alone, and good-view JEPAs of any number of modalities with the same world kernel $\pi$ agree up to one rotation each on every observation: \cref{thm:exact} when every state has many observations. (ii) Let a bad-view JEPA corrupt state and nuisance independently, $\Gamma_m=\boldsymbol A\otimes\boldsymbol B_m$ with $\boldsymbol A$ and $\boldsymbol B_m$ row-stochastic with stationary laws $p$ and $q_m$, so that corrupted views are observations with the law of clean ones, and let $1=\alpha_0\ge\alpha_1\ge\dots\ge0$ and $1=\beta_0\ge\beta_1\ge\dots\ge0$ be the eigenvalues of the Markov operators $\boldsymbol A^*\boldsymbol A$ on $L^2(p)$ and $\boldsymbol B_m^*\boldsymbol B_m$ on $L^2(q_m)$ of two corruptions of one state and of one nuisance ($\boldsymbol A^*$ the adjoint; $\alpha_i:=-\infty$ for $i\ge|\Z|$), with eigenfunctions $\psi_i$ and $\chi_j$ ($\psi_0=\chi_0=\mathbf 1$). Its kernel then has eigenvalues $\alpha_i\beta_j$ with eigenfunctions $\psi_i(z)\chi_j(\nu)$; let $\xi_1\ge\xi_2\ge\dots$ be those with $(i,j)\neq(0,0)$, with multiplicity. If $\xi_k>\xi_{k+1}$, $r$ of the top $k$ have $j\ge1$, its minimizers $f$ have $\mC_f$ with $r$ zero eigenvalues, i.e., $\rho(f)=1-r/k$, and by \cref{thm:nuisance} their alignment error with \emph{any} centered, whitened encoder of the other modality is at least $2r$. Moreover $r\ge1$ if and only if $\beta_1>\alpha_k$.
\end{tcolorbox}

\begin{proof}[Proof of \cref{prop:goodbad}]
(i) The good kernel $\pi(z,z')q_m(\nu)q_m(\nu')$ has marginal $p\otimes q_m$, and its Markov operator on $L^2(p\otimes q_m)$ is $T_\pi\otimes J_m$, where $J_m$ projects onto the constants of $L^2(q_m)$. Its eigenpairs are $(\mu_i,\phi_i\otimes\mathbf 1)$ and the eigenvalue $0$ on $L^2(p)\otimes\mathbf 1^\perp$. If $\mu_k>\max(\mu_{k+1},0)$, the $k$ largest eigenvalues on $\mathbf 1^\perp$ are $\mu_1,\dots,\mu_k$, separated from the rest, and \cref{prop:spectral} gives the minimizers $\mQ\vPhi_\pi\otimes\mathbf 1$: $f_m^\star(g_m(z,\nu))=\mQ_m\vPhi_\pi(z)$. For two modalities with the same $\pi$, $f_m^\star(g_m(z,\nu_m))=\mQ_m\mQ_{m'}^\top f_{m'}^\star(g_{m'}(z,\nu_{m'}))$ for all $z,\nu_m,\nu_{m'}$. (ii) Two corruptions of one observation, drawn independently given it, have the joint law $\Gamma_m^\top(\mD_p\otimes\mD_{q_m})\Gamma_m$ with $\Gamma_m=\boldsymbol A\otimes\boldsymbol B_m$, which factorizes as $(\boldsymbol A^\top\mD_p\boldsymbol A)\otimes(\boldsymbol B_m^\top\mD_{q_m}\boldsymbol B_m)$. By stationarity ($p^\top\boldsymbol A=p^\top$, $q_m^\top\boldsymbol B_m=q_m^\top$) its marginal is $p\otimes q_m$, the law of clean observations, so the minimizer is centered and whitened on clean observations and \cref{thm:nuisance} applies to it; its Markov operator is $(\boldsymbol A^*\boldsymbol A)\otimes(\boldsymbol B_m^*\boldsymbol B_m)$, with $\boldsymbol A^*$ the adjoint of $\boldsymbol A$ in $L^2(p)$, and $\boldsymbol A^*\boldsymbol A$ and $\boldsymbol B_m^*\boldsymbol B_m$ are positive semidefinite contractions fixing $\mathbf 1$, so its eigenpairs are $(\alpha_i\beta_j,\psi_i\otimes\chi_j)$ with $\alpha_i,\beta_j\in[0,1]$ and $\psi_0=\chi_0=\mathbf 1$. Under the separation assumption, \cref{prop:spectral} gives the minimizers $\mQ\boldsymbol\Psi$, where $\boldsymbol\Psi$ stacks the $k$ top eigenfunctions after $\psi_0\otimes\chi_0$. Since $\E[\psi_i(z)\chi_j(\nu)\mid z]=\psi_i(z)\,\E_{q_m}\chi_j$, which is $0$ for $j\ge1$ and $\psi_i(z)$ for $j=0$, $\mC_f=\mQ\,\mathrm{diag}(c)\,\mQ^\top$ with $c$ equal to $0$ on the $r$ eigenfunctions with $j\ge1$ and $1$ on the others, so $\mathrm{tr}\,\mC_f^{1/2}=k-r$, $\rho(f)=1-r/k$, and \cref{thm:nuisance} gives the error bound $2k(1-\rho(f))=2r$. Finally, every eigenvalue with $j\ge1$ satisfies $\alpha_i\beta_j\le\beta_j\le\beta_1=\alpha_0\beta_1$, so $r\ge1$ if and only if $\beta_1$ is among $\xi_1,\dots,\xi_k$; the eigenvalues with $j=0$ that exceed $\beta_1$ are the $\alpha_i>\beta_1$, $i\ge1$, and if $\beta_1>\alpha_k$ there are at most $k-1$ of them, so $\beta_1\ge\xi_k>\xi_{k+1}$ and $r\ge1$; conversely, if $\beta_1\le\alpha_k$, then $\xi_i=\alpha_i$ for $i\le k$ and separation leaves $r=0$. A corruption that keeps the nuisance exactly ($\boldsymbol B_m=\mI$, $\beta_j=1$ for every $j$) violates separation as soon as $|\mathcal N_m|>k+1$, but there every minimizer lies in the nuisance-only eigenspace of eigenvalue 1, and $\mC_f=0$ when $\alpha_1<1$.
\end{proof}

\begin{tcolorbox}[result,colframe=red!45!black,colback=red!2,title={Theorem~\ref*{thm:coverage} (Formal): Views and Data Set the World Kernel}]
Let modality $m$ have finitely many observations $x$, each of a world state $s_m(x)$, with law $\omega_m$ ($\omega_m(x)>0$) of state marginal $p$, and let $W_m$ be any symmetric law on pairs of observations whose marginal is $\omega_m$, the law on which encoders are whitened and evaluated: good views, $W_m(x,x')=\pi(s_m(x),s_m(x'))\,\omega_m(x\mid s_m(x))\,\omega_m(x'\mid s_m(x'))$; corruptions $\Gamma_m$ of one observation that leave the law of observations invariant ($\omega_m\Gamma_m=\omega_m$, as in \cref{prop:goodbad}); or a mix. Its induced world kernel $\pi_{W_m}(z,z')$ is the probability that a positive pair joins an observation of $z$ to one of $z'$. Let $P$ be the conditional expectation onto functions of the state in $L^2(\omega_m)$, let $\eta_m$ be the largest eigenvalue of the compression $(I-P)T_{W_m}(I-P)$ to \emph{nuisance functions} ($Ph=0$), which measures how much of its own nuisance a positive keeps ($\eta_m=-\infty$ if every state has one observation), and let $\varepsilon_m$ and $\varepsilon_m^{\mathrm{HS}}$ be the operator and Hilbert--Schmidt norms of $(I-P)T_{W_m}P$, which measure how much state and nuisance mix. Take the objective of \cref{prop:spectral} with $1\le k\le|\Z|-1$, let $\mu_1\ge\mu_2\ge\dots$ be the eigenvalues of $T_{\pi_{W_m}}$ on $\mathbf 1^\perp$ ($\mu_{|\Z|}:=-\infty$), and suppose $\delta_m=\mu_k-\max(\mu_{k+1},\eta_m)>0$. (i) Every spectral solution $f_m$ of $W_m$ (orthonormal top-$k$ eigenfunctions of $T_{W_m}$ on $\mathbf 1^\perp$) is close to a spectral embedding $\vPhi^{(m)}$ of its induced world kernel,
\[
\min_{\boldsymbol O\in\Oset}\big\|f_m-\boldsymbol O\,\vPhi^{(m)}\!\circ s_m\big\|_{L^2(\omega_m)}\le\tau_m:=\frac{2\sqrt2}{\delta_m}\min\big(\sqrt k\,\varepsilon_m,\sqrt2\,\varepsilon_m^{\mathrm{HS}}\big),
\]
and $\mathrm{tr}\,\mC_{f_m}\ge k-\tau_m^2$. (ii) For two modalities of one world, over paired observations from any coupling with $z\sim p$ and $x_m\mid z\sim\omega_m(\cdot\mid z)$,
\[
\Big|\min_{\mQ\in\Oset}\Big(\E\big\|f_1(x_1)-\mQ f_2(x_2)\big\|^2\Big)^{1/2}-d\Big|\le\tau_1+\tau_2,
\]
where $d=\min_{\mQ}\|\vPhi^{(1)}-\mQ\vPhi^{(2)}\|_{L^2(p)}$ is at most $\frac{2\sqrt2}{\delta}\min\big(\sqrt k\,\|\mK_{\pi_{W_1}}-\mK_{\pi_{W_2}}\|_2,\|\mK_{\pi_{W_1}}-\mK_{\pi_{W_2}}\|_F\big)$ by \cref{thm:robust}, with $\delta\ge\delta_2$ the gap $\mu_k-\mu_{k+1}$ of $\mK_{\pi_{W_2}}$.
\end{tcolorbox}

\paragraph{Special cases and scope.} Good views have $\varepsilon_m=0$, $\pi_{W_m}=\pi$ and $\eta_m=0$ (or $-\infty$), which gives \cref{thm:exact} with many observations per state when $\mu_k>\max(\mu_{k+1},0)$. The product corruptions of \cref{prop:goodbad} have $\varepsilon_m=0$, $\eta_m=\beta_1$ and $\pi_{W_m}=\boldsymbol A^\top\mD_p\boldsymbol A$, and $\delta_m>0$ reads $\alpha_k>\max(\alpha_{k+1},\beta_1)$: exactly the separation assumption of \cref{prop:goodbad} together with $r=0$ (for $|\mathcal N_m|\ge2$). The encoder class enters only through $\tau_m$: restricting encoders to be constant on groups of observations of one state is the theorem applied to the merged observations, which leaves $\pi_{W_m}$ unchanged and can only lower $\eta_m$ and $\varepsilon_m$, while inductive biases that merge observations of different states, or act softly, are outside the model.

\begin{proof}[Proof of \cref{thm:coverage}]
(i) Drop the index $m$. In $L^2(\omega)$, let $H_S$ be the functions of the state with zero mean and $H_N=\mathrm{range}(I-P)$ the nuisance functions, so that $\mathbf 1^\perp=H_S\oplus H_N$ orthogonally (a nuisance function has zero mean because its conditional mean is zero). For $h=a\circ s$ and $h'=a'\circ s$, $\langle h,T_Wh'\rangle_\omega=\sum_{x,x'}W(x,x')a(s(x))a'(s(x'))=\langle a,T_{\pi_W}a'\rangle_p$ and $\|h\|_\omega=\|a\|_p$, because the state marginal of $\omega$ is $p$: the compression of $T_W$ to $H_S$ is unitarily equivalent to $T_{\pi_W}$ on $\mathbf 1^\perp$. On $\mathbf 1^\perp$ write $T_W=T_0+F$ with $T_0=PT_WP+(I-P)T_W(I-P)$, which is block diagonal, and $F=(I-P)T_WP+PT_W(I-P)$, whose two off-diagonal blocks are adjoint, so $\|F\|_\mathrm{op}=\varepsilon$ and $\|F\|_\mathrm{HS}=\sqrt2\,\varepsilon^{\mathrm{HS}}$. The eigenvalues of $T_0$ are the $\mu_i$, with eigenfunctions $\phi_i\circ s$, and those of the compression to $H_N$, all at most $\eta$. If $\delta>0$, the $k$ largest are $\mu_1,\dots,\mu_k$ and the next is $\max(\mu_{k+1},\eta)$, so the top-$k$ eigenspace of $T_0$ is spanned by $\vPhi\circ s$ with gap $\delta$. The variant of the Davis--Kahan theorem of \citet[Theorem 2]{yu2015useful}, applied as in the proof of \cref{thm:robust} with $\Sigma=T_0$, $\hat\Sigma=T_W$, $r=1$, $s=k$, gives the bound for any orthonormal top-$k$ eigenbasis $f$ of $T_W$ on $\mathbf 1^\perp$, with or without ties in the spectrum of $T_W$; by the equality case of Ky Fan's principle in the proof of \cref{prop:spectral}, these are the minimizers of the objective on observations. Since $(I-P)(\vPhi\circ s)=0$ and $I-P$ is a contraction, $\|(I-P)f\|\le\|f-\boldsymbol O\vPhi\circ s\|\le\tau$ for the minimizing $\boldsymbol O$, and since $f$ is centered and whitened, $\mathrm{tr}\,\mC_f=\|Pf\|^2=k-\|(I-P)f\|^2\ge k-\tau^2$. (ii) By (i), $f_m=\boldsymbol O_m\vPhi^{(m)}\circ s_m+R_m$ with $\boldsymbol O_m\in\Oset$ and $\|R_m\|_{L^2(\omega_m)}\le\tau_m$. For $\mQ\in\Oset$, $f_1(x_1)-\mQ f_2(x_2)=\big(\boldsymbol O_1\vPhi^{(1)}(z)-\mQ\boldsymbol O_2\vPhi^{(2)}(z)\big)+R_1(x_1)-\mQ R_2(x_2)$, and the marginals of the paired law are $p$, $\omega_1$ and $\omega_2$, so by Minkowski's inequality the root-mean-square of the left side lies within $\tau_1+\tau_2$ of $\|\vPhi^{(1)}-\boldsymbol O_1^\top\mQ\boldsymbol O_2\vPhi^{(2)}\|_{L^2(p)}$; minimizing over $\mQ$, which makes $\boldsymbol O_1^\top\mQ\boldsymbol O_2$ range over $\Oset$, gives (ii). Both induced world kernels have marginal $p$, so \cref{thm:robust} bounds $d$, and $\delta\ge\delta_2>0$.
\end{proof}

\subsection{Next-Token Prediction is a Bad-View JEPA (\texorpdfstring{\cref{sec:ntp}}{Section 2.4})}

\begin{proof}[Proof of \cref{prop:ntp}]
(i) and the identity \eqref{eq:ntp} are immediate from the definition of the softmax. (ii) $A(h)=\log\sum_v e^{\langle h,w_v\rangle}$ involves the whole table but not $y$; and $\max_v a_v\le t\log\sum_ve^{a_v/t}\le\max_va_v+t\log|V|$. (iii) For a fixed context with conditional $p(\cdot\mid c)$ of full support, the expected loss $-\sum_vp(v\mid c)\log\mathrm{softmax}(\boldsymbol Wh(c))_v$, with $\boldsymbol W$ the matrix whose rows are the $w_v$, is minimized exactly when $\mathrm{softmax}(\boldsymbol Wh(c))=p(\cdot\mid c)$, i.e. $\boldsymbol Wh(c)=\log p(\cdot\mid c)+b(c)\mathbf 1$, which is attainable when $k$ is at least the rank of the matrix $[\log p(v\mid c)]_{v,c}$ modulo constants. The logits $\langle h,w_v\rangle$ are unchanged by $h\mapsto\boldsymbol Gh$, $w_v\mapsto\boldsymbol G^{-\top}w_v$ for any invertible $\boldsymbol G$.
\end{proof}

\subsection{Numerical Checks}

The released tests check the statements of \cref{sec:theory} numerically on random finite worlds: \cref{lem:cov} (for the prediction term), \cref{prop:spectral} (the spectral solution attains $2\sum_{i\le k}(1-\mu_i)$, and 50 random whitened embeddings do no better), \cref{thm:exact} (two modalities that render the world through different injective maps and solve their spectral JEPAs in closed form agree up to a rotation to machine precision), \cref{thm:robust} on 300 pairs of kernels with a common marginal, half of them clustered worlds with small spectral gaps (the bound holds in every world; the largest ratio of error to bound, .18, is that of \cref{fig:dk}), \cref{prop:ident} (Procrustes on $k$ generic pairs recovers $\mQ$, and $k-1$ pairs admit another exact fit), the decomposition and $\mathrm{GL}(k)$ invariance of \cref{prop:ntp}, \cref{thm:nuisance} (the bound holds for 60 pairs of random whitened encoders of two modalities with nuisances), and \cref{prop:goodbad} (good-view solutions of two modalities with different nuisances agree up to a rotation to machine precision; bad-view solutions lose exactly the predicted $r$ directions to nuisance and stay at least $2r$ from a good encoder of the other modality; a corruption that forgets the nuisance loses none), and \cref{thm:coverage} (product corruptions give $\varepsilon=0$, $\eta=\beta_1$ and $\pi_W=\boldsymbol A^\top\mD_p\boldsymbol A$, and on 200 random ones the bound is finite exactly when \cref{prop:goodbad} finds separation and $r=0$; positives that resample the nuisance align exactly; on random worlds with one to four observations per state and positives that mix good views, pairs of an observation with itself and random pairs, both bounds always hold, and where they are informative, below the trivial $\sqrt{2k}$--2,848 cases for (i), 307 for (ii)--the error is at most 32\% of the bound in (i), median 11\%, and 10\% in (ii), median 0.8\%).

\section{Experimental details}\label{app:details}

\paragraph{Code.} Everything is built on stable-pretraining: its \texttt{LeJEPA} class (loss, SIGReg with 1024 slices, projector $384\!\to\!512\!\to\!2048\!\to\!2048\!\to\!512$ with batch normalization), \texttt{Module}, \texttt{Manager} (training loop, checkpointing, SLURM requeue), and GPU augmentation ops. The experiment-specific code is a text view generator, a text encoder wrapper, a next-token loss, and the alignment metrics, a few hundred lines.

\paragraph{Image encoder.} ViT-S/16 (timm \texttt{vit\_small\_patch16\_224}, drop path 0.1), CLS embedding. Views: 2 global crops ($224^2$, scale 0.3--1.0) and 6 local crops ($96^2$, scale 0.05--0.3), each with horizontal flip, color jitter (0.4, 0.4, 0.2, 0.1; $p=0.8$), grayscale ($p=0.2$), Gaussian blur ($p=0.5$) and solarization ($p=0.2$). Crops are made on CPU workers and photometric operations on GPU.

\paragraph{Text encoders.} BERT with hidden size 384, 12 layers, 6 heads, MLP 1536, random initialization, \texttt{bert-base-uncased} tokenizer, maximum length 64. The JEPAs use the CLS token. Views are trimmed: global views keep a contiguous span of 50--100\% of the caption, local views 15--50\%; 15\% of kept tokens are replaced by \texttt{[MASK]}. The next-token model uses causal attention, a softmax head tied to the input embeddings, and mean-pooled final hidden states as its embedding; the last token's hidden state, the other usual choice for a causal model, aligns worse with DINOv2 ViT-B (Flickr R@1 .061 against .119 for mean pooling, three seeds; \texttt{evaluate.py embed --last-token}).

\paragraph{Optimization.} AdamW, learning rate $5\cdot10^{-4}$ with 10 epochs of linear warmup and cosine decay, weight decay 0.05, batch 256, bf16, 200 epochs, seeds 0, 1 and 2 (image and text reseeded together), for the Flickr runs. Longer and larger text runs (BERT-B on Flickr, the Flickr+COCO runs, every Recap run) use learning rate $10^{-4}$ with gradient clipping at 1.0 and one seed unless stated; the MSR-VTT runs use 800 epochs. One GPU per run, except eight for the ImageNet and ViT-B image JEPAs and for the 940M-image BERT-B runs (16 for its first seed), and two for the 200M-image BERT-B.

\paragraph{MSR-VTT.} The 7{,}010 clips of the 7k split train, the other 1{,}990 clips of the 9k split (official test clips outside the JSFUSION split, as is usual with the 9k split) fit the means and the rotation, and the 1{,}000 clips of the JSFUSION split test. We keep 8 evenly spaced frames per clip (\texttt{prh.data.extract\_frames}). Good image views draw each of the 2 global and 6 local crops from its own frame of the clip; bad image views crop one frame, drawn once per clip and epoch. Text views are as on Flickr, with 20 captions per clip. All runs use the Flickr recipe for 800 epochs (21.6k steps, about as many as on Flickr), seeds 0, 1 and 2; evaluation uses the middle frame and, for test clips, the JSFUSION caption.

\paragraph{PRH protocol.} For every WIT-1024 score we follow the released code of \citet{huh2024platonic}: WIT-1024 pairs, first caption, per-layer features (vision: CLS after each block; language: mask-weighted mean of each hidden state), elementwise clipping at the 0.95 quantile of absolute values, $\ell_2$ normalization, mutual 10-NN, maximum over layer pairs.

\section{Additional results}\label{app:results}

\begin{table}[h]
\centering
\small
\caption{\textbf{Good views also make better text encoders.} Text encoders alone (caption$\to$caption retrieval among the 5{,}000 Flickr test captions: is the nearest caption one of the four others describing the same image?) and alignment measured on projector outputs (seed 0; --: no trained projector). We clearly observe that \emph{the bad-view JEPA learns little beyond word overlap (below TF-IDF), while the good-view JEPA goes well beyond it.} The image JEPA reaches 78.5\% 20-NN accuracy on Imagenette \citep{howard2019imagenette} (29.1\% at initialization).}
\label{tab:text_only}
\begin{tabular}{lcccc}
\toprule
& caption$\to$caption R@1 & mutual 10-NN (projector) & R@1 (projector) & whitened R@1 (projector)\\
\midrule
Good-view JEPA & \textbf{.606} & \textbf{.064} & \textbf{.030} & \textbf{.034}\\
Bad-view JEPA & .381 & .044 & .018 & .022\\
Next-token prediction & .331 & -- & -- & --\\
TF-IDF & .442 & -- & -- & --\\
Random init & .178 & -- & -- & --\\
\bottomrule
\end{tabular}
\end{table}

\begin{figure}[h]
\centering
\includegraphics{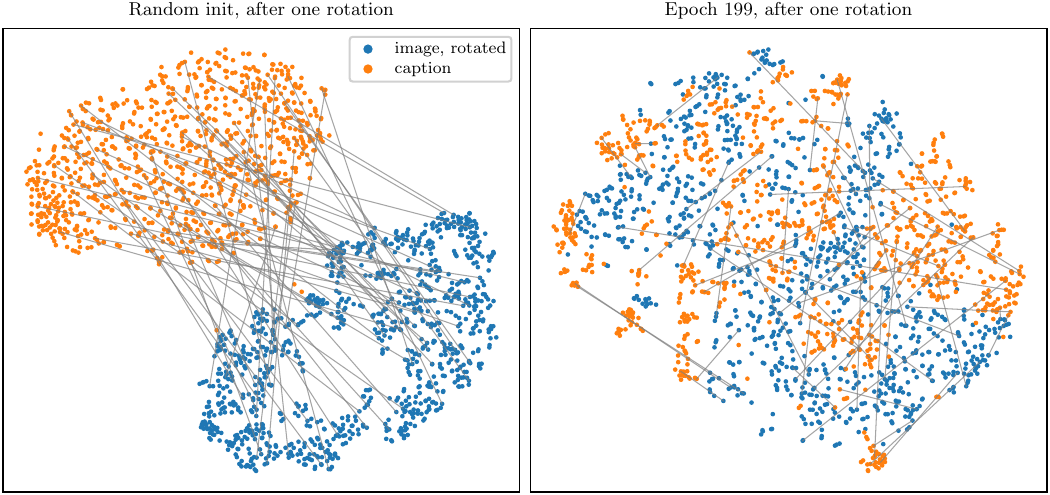}\\[4pt]
\includegraphics{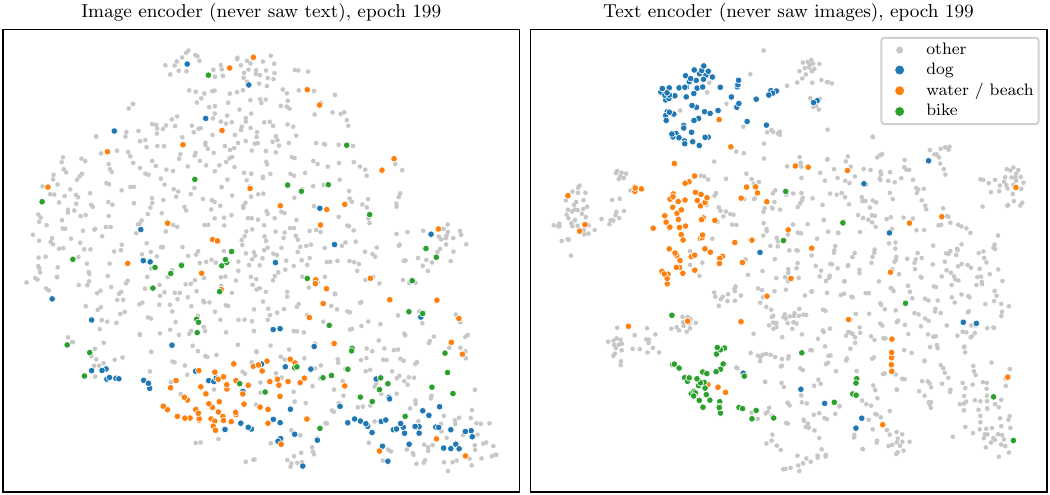}
\caption{\textbf{One rotation superimposes two separately trained encoders.} Top: joint t-SNE of test images after the best rotation (blue) and their captions (orange), at initialization and after training (good-view JEPA); grey lines join 60 random true pairs. Bottom: independent t-SNEs of the image and text embeddings, colored by caption keywords used only for coloring. We clearly observe that \emph{after training, images and their captions share one geometry}.}
\label{fig:tsne}
\end{figure}

\begin{figure}[h]
\centering
\includegraphics{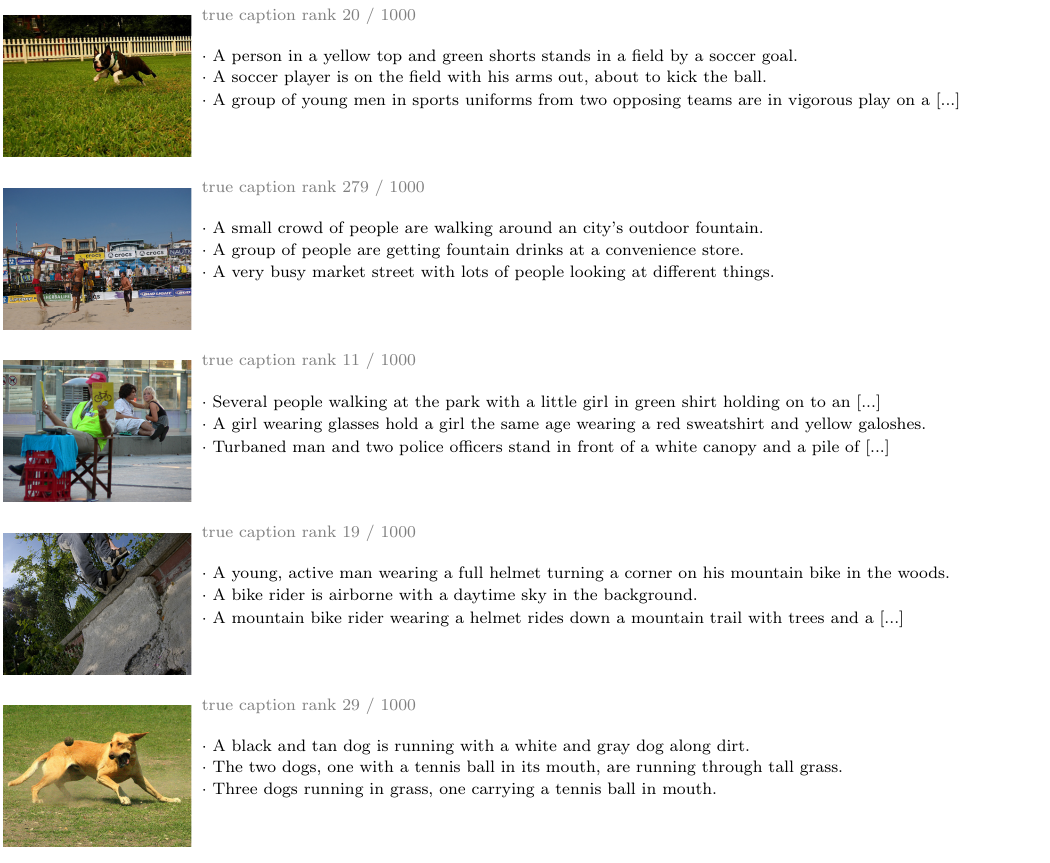}
\caption{\textbf{Retrieval through one rotation.} Image$\to$caption retrieval (good-view JEPA, epoch 199) for five Flickr test images, with the rank of the true caption among the 1{,}000 above each list ($\checkmark$ would mark it among the top three). We observe that \emph{misses are mostly semantically close}.}
\label{fig:retrieval}
\end{figure}

\paragraph{Text encoders alone.} \Cref{tab:text_only} scores each text encoder without images, and the alignment of projector outputs; \cref{fig:tsne,fig:retrieval} show the alignment of the good-view JEPA qualitatively.

\paragraph{Next-token overfitting.} On 145k short captions the next-token model's validation loss is lowest near epoch 29 (3.04) and rises to 4.49 by epoch 189. The objectives also see the captions differently: each next-token epoch reads one caption per image, while each JEPA sample draws its eight views from all five, so next-token prediction sees fewer tokens per step; more passes would not close the gap, since its validation loss is lowest at epoch 29 of 200, and the caption-pair control (\cref{tab:main}) gives it two captions of the same image per sample. All our tables report the last checkpoint of each run (\cref{tab:bigscale} also lists intermediate checkpoints of the 940M-image BERT-B); early stopping it on the validation loss instead (epoch 29 for each seed) gives R@1 .134 against DINOv2 (\cref{tab:calib}), and even selecting its best checkpoint on the test pairs would give at most .146, still far below the good-view JEPA (.246).

\paragraph{Nuisance on real encoders.} \Cref{tab:nuisance} gives the measurement of \cref{thm:nuisance} behind \cref{sec:exp}: each Flickr text encoder (epoch 199) embeds all five captions of the 2{,}014 validation and test images; we keep its top-$k$ principal directions and whiten them, estimate $\mC_f$ as the covariance of the per-image caption means minus a fifth of the mean within-image covariance (the variance the average of five captions keeps), and clip its eigenvalues to $[0,1]$. The best-rotation error pairs each image's first caption, whitened, with DINOv2 ViT-B restricted to its own top-$k$ principal directions, whitened, as $2k-2\|\boldsymbol\Sigma\|_*$ (\texttt{evaluate.py nuisance}).

\begin{table}[h]
\centering
\small
\caption{\textbf{Bad views spend their embedding on nuisance.} Scene-explained fraction $\rho(f)=\mathrm{tr}\,\mC_f^{1/2}/k$ of each text encoder's top-$k$ whitened embedding (mean $\pm$ half-range over three seeds), the lower bound $2k(1-\rho(f))$ of \cref{thm:nuisance} on the best-rotation error against DINOv2 ViT-B, and that error (means over seeds; bounds from the unrounded fractions; uncorrelated embeddings give $2k$, i.e., 64 and 256). We clearly observe that \emph{the bound holds for every encoder, and the scene-explained fraction ranks the encoders as their alignment does}.}
\label{tab:nuisance}
\begin{tabular}{lcccccc}
\toprule
& \multicolumn{3}{c}{$k=32$} & \multicolumn{3}{c}{$k=128$}\\
Text encoder (Flickr) & scene-explained & bound & error & scene-explained & bound & error\\
\midrule
Good-view JEPA & .804{\scriptsize$\pm$.003} & 12.6 & 33.5 & .654{\scriptsize$\pm$.005} & 88.6 & 157.0\\
Bad-view JEPA & .655{\scriptsize$\pm$.005} & 22.1 & 39.4 & .507{\scriptsize$\pm$.003} & 126.1 & 171.5\\
Next-token & .608{\scriptsize$\pm$.001} & 25.1 & 40.3 & .440{\scriptsize$\pm$.002} & 143.4 & 176.6\\
\midrule
\multicolumn{7}{l}{\emph{trained on Recap texts only (seed 0), measured on the same Flickr captions}}\\
Good-view JEPA, 50M images & .802 & 12.7 & 32.4 & .642 & 91.7 & 156.8\\
Bad-view JEPA, 50M images & .673 & 20.9 & 37.6 & .506 & 126.4 & 168.7\\
Next-token, 50M images & .618 & 24.4 & 40.1 & .483 & 132.3 & 171.0\\
Good-view JEPA, 100M images & .806 & 12.4 & 32.0 & .661 & 86.9 & 153.7\\
Good-view JEPA, BERT-B, 940M images & \textbf{.830} & 10.9 & \textbf{30.0} & \textbf{.722} & 71.1 & \textbf{145.3}\\
\bottomrule
\end{tabular}
\end{table}

\paragraph{Which texts give the gain.} Two of each Recap-DataComp image's three texts are written by an image captioner, so its good views carry information distilled from image--text training. \Cref{tab:sources} separates the texts from the views with one recipe per scale (BERT-S, learning rate $10^{-4}$, gradient clipping at 1, 64 tokens; 20M samples on the texts of 2.24M images, 40M on 50M; three seeds each). Without any image captioner--the alt-text paired with a paraphrase that a text-only language model (Qwen2.5-7B-Instruct) writes from the alt-text alone--good views reach .060 on WIT, against .107 with the three Recap texts on the same images and .080 with the two LLaVA recaptions alone (.096 against .125 on the texts of 50M images), which retrieve Flickr captions best at both scales (R@1 .348 and .386), plausibly because Flickr captions are literal like LLaVA's while WIT's Wikipedia captions read more like alt-texts; on the same alt-texts, good views from the paraphrase still beat bad views of the alt-text (.060 against .055 on WIT, .233 against .180 R@1, no overlap between seeds), so the views help without any image captioner; with the same bad views, the two LLaVA recaptions reach .092 and the alt-text .056 (\texttt{tables.py}). Measured on Flickr captions, the scene-explained fraction of \cref{thm:nuisance} ranks the nine encoders as WIT does but for one adjacent pair (Spearman .98). That is, \emph{the data matter even with bad views, and the views matter with any texts}, as \cref{thm:coverage} predicts: literal descriptions make positives connect related scenes, and a paraphrase of an alt-text, a meaning-preserving corruption of one text, keeps the alt-text's own nuisance.

\begin{table}[h]
\centering
\small
\caption{\textbf{The views and the data both set the alignment.} BERT-S text JEPAs, one recipe per scale, three seeds (mean and range), scored on WIT-1024 with DINOv2 ViT-B and on Flickr through one rotation fitted on 30k pairs; scene-explained: $\rho(f)$ on Flickr captions ($k=32$, seed 0, \cref{tab:nuisance}). The paraphrase is written from the alt-text alone by a text-only language model, so that row involves no image captioner. We observe that \emph{with the same bad views, LLaVA recaptions align far better than alt-texts, and with the same texts, good views double alignment} (\cref{thm:coverage}).}
\label{tab:sources}
\begin{tabular}{llcccc}
\toprule
Texts of & texts per image & views & WIT (DINOv2-B) & Flickr R@1 & scene-explained\\
\midrule
2.24M images & alt-text, two LLaVA recaptions & good & .107 (.095--.115) & .339 (.336--.344) & .793\\
2.24M images & alt-text, text-only paraphrase & good & .060 (.058--.061) & .233 (.230--.235) & .654\\
2.24M images & two LLaVA recaptions & good & .080 (.076--.082) & .348 (.346--.351) & .778\\
2.24M images & alt-text & bad & .055 (.054--.056) & .180 (.179--.180) & .606\\
\midrule
50M images & alt-text, two LLaVA recaptions & good & \textbf{.125} (.118--.129) & .376 (.370--.381) & \textbf{.802}\\
50M images & two LLaVA recaptions & good & .096 (.093--.100) & \textbf{.386} (.373--.396) & .789\\
50M images & alt-text, two LLaVA recaptions & bad & .063 (.062--.066) & .290 (.283--.301) & .673\\
50M images & two LLaVA recaptions & bad & .092 (.090--.093) & .325 (.323--.329) & .734\\
50M images & alt-text & bad & .056 (.056--.057) & .199 (.193--.203) & .618\\
\bottomrule
\end{tabular}
\end{table}

\begin{figure}[h]
\centering
\includegraphics{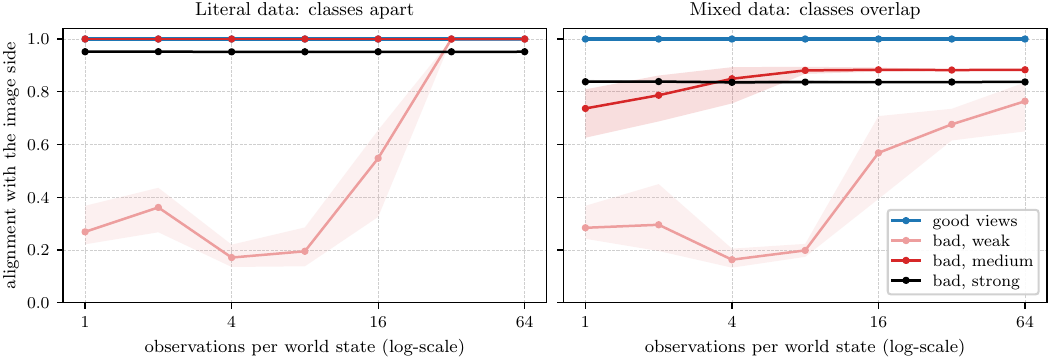}
\caption{\textbf{Weak views need many data, the kind of data sets the ceiling, good views need neither (\cref{thm:coverage}).} A synthetic world of five classes of six states; the image side is a good-view JEPA whose world kernel links states of one class. The text side observes each state through samples on a line, one block per class, with gaps between blocks (left, literal data) or without (right, mixed data); its bad views join two samples with weight $e^{-d^2/2\Delta^2}$ (weak $\Delta=.005$, medium $.08$, strong $1$; a state has width $1/6$). Alignment is the mean canonical correlation of the two spectral solutions over paired observations ($k=4$; 1: equal up to a rotation); lines are means over five samplings, bands their range.}
\label{fig:coverage}
\end{figure}

\paragraph{Data kind in the synthetic world.} \Cref{fig:coverage} follows bad views of three reaches along the number of samples per state, on literal data (classes apart in the text space) and mixed data (classes adjacent): weak views need many samples, strong ones blur the classes at any sample size, and mixed data cap every bad-view JEPA near .89, while good views align exactly (\texttt{figures.py theory}).

\paragraph{The scaling law in detail.} \Cref{fig:phase_law} (\cref{sec:law}) varies the positives and the data with one recipe (BERT-S, learning rate $10^{-4}$, clipping 1, 64 tokens, 1M samples per epoch, seed 0): each training item draws good views (two texts of its image) with probability $q\in\{0,.001,.003,.01,.03,.1,.3,1\}$ and bad views (two corruptions of one text) otherwise, on the texts of $N\in\{0.41,2.24,10.2,50.9\}$M images for $M=20$M samples, and on 10.2M images for $M\in\{5,40,80\}$M at $q\in\{0,.01,.1,.3,1\}$ (\texttt{tables.py}). Bad views gain nothing from more images: at $q=0$, WIT stays within .054--.057 from 0.41M to 50.9M images, while good views rise from .090 to .115. A fraction of good views helps Flickr retrieval before WIT (R@1 .26 at $q=0$, .28 at $q=.1$, .31--.33 at $q=.3$; WIT only from $q=.3$), and training longer raises both views (R@1 .221 to .310 from 5M to 80M samples with bad views, .292 to .409 with good views; WIT with good views .078, .099, .124 and .132 at 5M, 20M, 40M and 80M samples). The law
\[
\mathrm{R@1}=a+d\log M+b\log\!\big(1+q^{\alpha}M^{\beta}N^{\gamma}/c\big),
\]
fitted on the 33 cells at 5M and 20M samples finished when its predictions were registered (RMSE .0067), predicted the eight 40M- and 80M-sample cells before they finished with RMSE .0115: within .007 at 40M samples and within .024 at 80M, where it underestimates bad views most (.024 at $q=0$), as the bad-view floor rises faster than the $\log M$ fitted on short runs. A simpler law in the good-view samples $qM$ alone, registered first, underestimated the 80M-sample cells by .019 to .055: bad-view samples are not worthless. Refitted on all 37 cells at 5M and 20M samples (RMSE .0068), its iso-alignment lines run as $q\propto M^{-\beta/\alpha}=M^{-0.57}$ ($\alpha=1.38\pm.67$, $\beta=.79\pm.53$), so more training substitutes for good views, sub-linearly, and unique images enter weakly ($\gamma=.26\pm.15$). The exponents are poorly identified--refitted on all 45 grid cells up to 80M samples (v2, \cref{fig:phase_law}), the same lines run as $M^{-0.30}$ with an equally good fit (RMSE .0072)--and the law is an empirical summary of one recipe, not a consequence of \cref{thm:coverage}. To test it off the grid, we then trained runs at random $(q,M)$ on 10.2M images--$q$ log-uniform on $[.001,1]$ or zero, $M$ drawn on a log scale, from 3M to 133M samples--and three runs at 320M samples, after registering the predictions of both fits for every one of them. On all 85 random runs, v2 predicts R@1 with RMSE .0093: .0065 for $q\ge.1$, .0071 for $.01\le q<.1$, and .012 below, where it underestimates the bad-view floor between 20M and 60M samples (by .013 on average, at most .024). Reruns of six grid cells with two other seeds each put the seed-to-seed standard deviation of one run at .0068 (pooled, 18 runs), so two seeds of one run differ by $\sqrt2\times.0068=.0096$ in RMSE: \emph{the law predicts a run it has never seen as well as a rerun of that run with another seed}. It fails further out, for good views only: at 320M samples, four times the longest fitted run and 31 passes over the 10.2M images, the runs with good views reach R@1 .418 against .455 predicted at $q=1$ and .348 against .389 at $q=.1$ (WIT still rises at $q=1$, from .132 at 80M samples to .145), while the run without reaches .338 against .337; the law, which has no term for repeated data, overestimates very long training with good views (\cref{fig:phase_law}, diamonds).

\paragraph{Robustness of the good-view recipe.} \Cref{tab:sweep} gives the numbers behind \cref{fig:ablations}, three seeds per setting where shown, with the best checkpoint of seed 0 for reference (selected on test pairs, so not used in the text). Gains saturate near epoch 100 on Flickr's 145k captions, and a wider encoder (BERT-B) matches BERT-S at its peak while degrading less afterwards. Collapses come abruptly, after a stretch of normal training: the mask-30\% seed that collapses reaches .243 at epoch 49 and chance by epoch 59, and the two 2 + 2-view seeds collapse between epochs 19 and 39 after tracking seed 0 until then. Scored on validation pairs alone instead (one rotation fitted on 507 of the 1{,}014 validation pairs, R@1 among the other 507; \texttt{evaluate.py align --val-only --split 507}), the same runs collapse, and the two trimming changes remain the two lowest settings with several seeds (.209 and .245, against .271--.316 for every other one), while the other settings, within seed noise of each other, change order (Spearman .40 between validation and test means). That is, \emph{no conclusion of \cref{fig:ablations} rests on selection on test pairs}. Outside this sweep, four runs collapsed and feed no table or figure: the default recipe trained for 1{,}000 epochs (to chance between epochs 139 and 159), our first good and bad runs on web text at learning rate $5\cdot10^{-4}$ (by epoch 7), and BERT-L on web text, twice, even with gradient clipping (by epochs 3 to 7); hence the lower learning rate and the clipping of every longer or larger run, and no BERT-L beyond Flickr and COCO captions. The $\lambda=0.005$ run with clipping does not collapse but is partly degenerate (effective rank 33 of 384), although it scores .238.

\begin{table}[h]
\centering
\small
\caption{\textbf{Cropping matters, the rest barely moves the mean, and clipping prevents most collapses.} R@1 against frozen DINOv2 on Flickr test pairs with one rotation fitted on validation pairs (chance .001), last epoch, for three seeds where shown (one otherwise); a run collapses (\texttimes) when its last-epoch R@1 falls below .05. Default views: 2 global and 6 local spans of 50--100\% and 15--50\% of a caption. Mean: over the seeds that do not collapse (--: every seed collapses). $^\ast$Best over checkpoints on the test pairs, for reference only; the other columns involve no selection. That is, \emph{every run that does not collapse stays far above the bad-view JEPA and next-token prediction}, and without gradient clipping 3 of the 36 runs with the default SIGReg weight collapse. In stable-pretraining as we ran it, SIGReg is scaled by the number of embeddings it sees, so the batch and view rows also change its effective weight: $\times2$ for batch 512, $\times\frac12$ for 2 + 2 views and $\times1.5$ for 2 + 10 views.}
\label{tab:sweep}
\begin{tabular}{lccc}
\toprule
Good-view text JEPA, one change from the default & last-epoch R@1, seeds 0 / 1 / 2 & mean & best$^\ast$ (epoch), seed 0\\
\midrule
Default: 2 + 6 views, mask 15\%, lr $5\cdot10^{-4}$, $\lambda=0.02$ & .248 / .260 / .230 & .246 & .276 (99)\\
\quad BERT-B (768 wide; lr $10^{-4}$, clipping 1) & .273 / .252 / .255 & .260 & .275 (189)\\
\quad BERT-T (192 wide) & .244 / .229 / .228 & .234 & .254 (149)\\
\quad no masking & .236 / .246 / .244 & .242 & .259 (109)\\
\quad mask 30\% & .243 / \texttimes{} / .264 & .254 & .266 (129)\\
\quad no trimming (whole captions) & .193 / .182 / .180 & .185 & .242 (79)\\
\quad milder trimming (70--100\% / 40--80\%) & .211 / .204 / .191 & .202 & .268 (59)\\
\quad 2 + 2 views & .247 / \texttimes{} / \texttimes{} & .247 & .257 (129)\\
\quad 2 + 2 views, clipping 1 & .260 / .258 / .267 & \textbf{.262} & .272 (159)\\
\quad 2 + 10 views & .254 / .247 / .243 & .248 & .262 (119)\\
\quad 4 + 4 views & .242 / .249 / .248 & .246 & .258 (109)\\
\quad lr $2\cdot10^{-4}$ & .265 / .246 / .253 & .255 & .270 (169)\\
\quad lr $10^{-3}$ (clipping 1) & collapses & -- & --\\
\quad batch 512 & .239 / .243 / .240 & .241 & .245 (129)\\
\quad 100 epochs & .233 / .261 / .248 & .247 & .239 (59)\\
\quad $\lambda=0.005$ & collapses & -- & .196 (29)\\
\quad $\lambda=0.005$, clipping 1 & .238 & .238 & .242 (119)\\
\quad $\lambda=0.05$ & collapses & -- & .242 (89)\\
\quad $\lambda=0.05$, clipping 1 & collapses & -- & .162 (19)\\
\quad $\lambda=0.1$ & collapses & -- & .191 (79)\\
\quad $\lambda=0.1$, clipping 1 & .202 & .202 & .215 (149)\\
\quad 1{,}000 epochs, stopped at 359 (lr $10^{-4}$, clipping 1) & .242 (359) & .242 & .278 (99)\\
\bottomrule
\end{tabular}
\end{table}

\begin{table}[h]
\centering
\small
\setlength{\tabcolsep}{4pt}
\caption{\textbf{Captioning from independently trained encoders.} BLEU-4 \citep{papineni2002bleu}, CIDEr \citep{vedantam2015cider} on Flickr test images. The decoder never sees an image; the image encoder never sees text; pairs only fit one linear map. BLEU-4 rewards generic n-grams, so the random map's fluent but generic captions score above 1{,}014-pair retrieval on BLEU-4, not on CIDEr. We clearly observe that \emph{decoding through the rotation beats retrieving captions through it by a wide margin}.}
\label{tab:caption}
\begin{tabular}{lcc}
\toprule
Captioner (Flickr test) & BLEU-4 & CIDEr \\
\midrule
Oracle: decode the true caption embedding & .261 & .628 \\
\midrule
Image $\to$ rotation (30k pairs) $\to$ decode & \textbf{.185} & \textbf{.431} \\
Image $\to$ rotation (1{,}014 pairs) $\to$ decode & .173 & .396 \\
Image $\to$ rotation (300 pairs) $\to$ decode & .146 & .317 \\
Image $\to$ rotation (100 pairs) $\to$ decode & .119 & .244 \\
\quad + unpaired refinement (100 pairs) & .138 & .294 \\
\quad + unpaired refinement (30 pairs) & .099 & .169 \\
Retrieve nearest caption (30k pairs) & .065 & .188 \\
Retrieve nearest caption (1{,}014 pairs) & .054 & .158 \\
Random rotation & .059 & .043 \\
\bottomrule
\end{tabular}
\end{table}

\begin{figure}[h]
\centering
\includegraphics{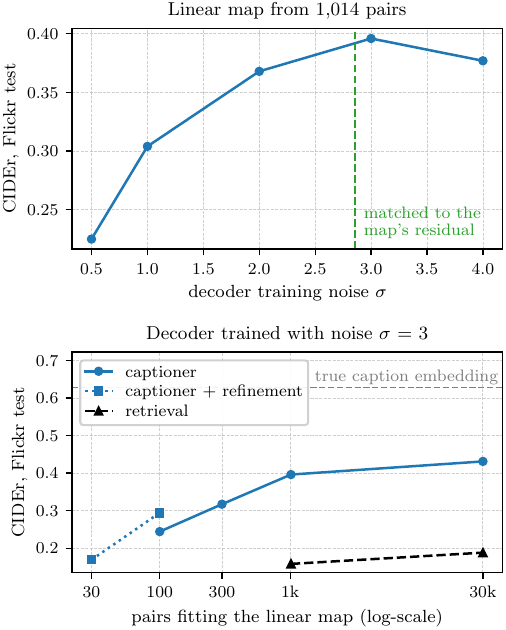}
\caption{\textbf{The captioner's noise and pairs are predicted, not tuned.} Flickr test images. Top: CIDEr against the noise the decoder is trained with; dashed, the noise matched to the residual of the rotation (rotation from 1{,}014 pairs). Bottom: CIDEr against the number of pairs that fit the one linear map, with and without unpaired refinement, against retrieving the nearest caption through the same map; dashed, decoding the true caption embedding. We clearly observe that \emph{transfer peaks at the matched noise, and that a few pairs refined on unpaired data match caption retrieval from 1{,}014 pairs.}}
\label{fig:captioner}
\end{figure}

\begin{table}[h]
\centering
\small
\setlength{\tabcolsep}{4pt}
\caption{\textbf{The captioner improves without pairs.} Building the captioner up one change at a time: Flickr test (1{,}000 images) and COCO Karpathy test (5{,}000 images), beam 3 unless stated; consensus rows choose among the greedy caption and 32 samples. The text JEPA is BERT-S trained on the texts of 100M web images; the decoders are trained on its embeddings of Flickr and COCO training captions only, with noise matched to the alignment ($\sigma=1.9$); the image encoders are frozen. The first row is the captioner of \cref{tab:caption}: on Flickr, the Flickr text JEPA and decoder ($\sigma=3$); on COCO, the Flickr and COCO ones ($\sigma=2$); it decodes greedily on Flickr and with beam 3 on COCO. Decoding the true caption embedding gives .694 and 1.305 (captions-as-written row), and .743 and 1.369 with the web-pretrained decoder. Only the rotation is fitted on image--caption pairs. Without consensus decoding, the best captioner reaches .496 and .806. This demonstrates that \emph{stronger frozen encoders, a pretrained language model and consensus decoding each add to the captioner, from CIDEr .668 to .838 on COCO}, while a larger decoder does not.}
\label{tab:captioner_scale}
\begin{tabular}{lcc}
\toprule
Captioner (CIDEr; one rotation from the 30k Flickr pairs) & Flickr & COCO \\
\midrule
Small decoder, DINOv2 ViT-B, in-domain text JEPA (\cref{tab:caption}) & .431 & .668 \\
\midrule
Qwen2.5-1.5B decoder, web text JEPA, DINOv2 ViT-g (CLS) & .466 & .720 \\
\quad + mean patch token & .480 & .756 \\
\quad + captions as written (case, punctuation) & .479 & .765 \\
\quad + DINOv3 ViT-L, concatenated & .496 & .806 \\
\quad + consensus of 32 samples & .519 & \textbf{.838} \\
\midrule
\multicolumn{3}{l}{\emph{Variants of the DINOv3 ViT-L row}}\\
\quad DINOv3 ViT-7B instead of ViT-L & .507 & .816 \\
\quad decoder first trained on 4M web sentences & .504 & .802 \\
\quad\quad map trained on the same pairs instead: any linear map & .531 & .635 \\
\quad\quad map trained on the same pairs instead: two-layer MLP & .498 & .585 \\
\quad\quad + consensus of 32 samples (the demo) & \textbf{.538} & .832 \\
\quad\quad\quad map trained on the same pairs instead: any linear map & .561 & .652 \\
\quad\quad\quad map trained on the same pairs instead: two-layer MLP & .511 & .597 \\
\quad Qwen2.5-3B decoder (captions-as-written row) & .460 & .730 \\
\bottomrule
\end{tabular}
\end{table}

\paragraph{Few pairs and no pairs.} The unpaired refinement alternates optimal transport between the rotated image cloud and the text cloud (entropic, Sinkhorn) and Procrustes on the transport plan, on the 1{,}014 validation images and captions with their pairing hidden. From 100 pairs, CIDEr rises from .244 to .294; from 30 pairs, the refined captioner reaches .169 (\cref{tab:caption}). With no pairs at all, matching these two small clouds by their internal geometry (Gromov--Wasserstein \citep{memoli2011gromov,peyre2016gromov}, then the same refinement) stays at chance (test R@1 $\approx$ .001); so does the method of \citet{schnaus2026rosetta} with this Flickr-caption JEPA (R@1 .012), whose geometry is learned from 145k short captions. With the good-view JEPAs trained on web text, that method aligns the two clouds far above chance (\cref{sec:pairfree}).

\paragraph{Video.} \Cref{tab:video} details the video results of \cref{sec:caption}: each MSR-VTT clip is embedded by the captioner's frozen image encoders (DINOv2 ViT-g and DINOv3 ViT-L) on 8 evenly spaced frames, the rotated embeddings are averaged, and the decoder of \cref{tab:captioner_scale} (web-pretrained Qwen2.5-1.5B) decodes the mean (\texttt{evaluate.py video-embed}, \texttt{video-caption} and \texttt{video-align}). Its captions are in the style of image captions and longer than MSR-VTT's (11.4 against 9.3 words), and miss what the clips' human captions often describe, actions and speech. The demo's map, fitted on the 30k Flickr pairs and 200k web image--text pairs, reaches .268.

\begin{table}[h]
\centering
\small
\caption{\textbf{Captions and retrieval for video, from photos alone.} Top: zero-shot captioning of the MSR-VTT test clips (20 references each; CIDEr on a 0--1 scale) by our captioner, whose networks never saw a video, a video--caption pair or MSR-VTT text, and by captioners built on CLIP's encoders without MSR-VTT text, with one trained on MSR-VTT captions for reference (--: not reported). Bottom: alignment (mutual 10-NN fits nothing) and retrieval through one map on the 1k-A split, against zero-shot CLIP. We observe that \emph{a rotation fitted on photos captions and retrieves videos}, above captioners built on CLIP that use no MSR-VTT text, though CLIP's jointly trained encoders retrieve better.}
\label{tab:video}
\begin{tabular}{lcccc}
\toprule
MSR-VTT test, 2{,}990 clips & CIDEr & BLEU-4 & METEOR & ROUGE-L\\
\midrule
Ours: mean of 8 frames, rotation from 30k Flickr pairs, consensus of 32 & \textbf{.246} & .166 & .206 & .444\\
\quad beam 3 & .222 & .154 & .204 & .432\\
\quad map from 30k Flickr and 200k web pairs (the demo's) & .268 & .186 & .232 & .471\\
\quad\quad middle frame instead of the mean of 8 & .242 & .172 & .219 & .457\\
\quad oracle: decode the text embedding of one human caption & .383 & .257 & .265 & .538\\
\midrule
\multicolumn{5}{l}{\emph{CLIP ViT-B/32 encoders (400M image--text pairs), no MSR-VTT text; from \citet{li2023decap}, Tab.~4}}\\
ZeroCap \citep{tewel2022zerocap} & .058 & -- & -- & --\\
MAGIC \citep{su2022magic} & .074 & -- & -- & --\\
DeCap, decoder on CC3M captions & .150 & -- & -- & --\\
DeCap, decoder on COCO captions & .186 & .147 & .204 & --\\
DeCap, decoder on MSR-VTT captions & .348 & -- & -- & --\\
\bottomrule
\end{tabular}

\medskip
\begin{tabular}{llccc}
\toprule
MSR-VTT 1k-A (1{,}000 clips) & map fitted on & mutual 10-NN & text$\to$video R@1/5/10 & video$\to$text R@1\\
\midrule
ours, mean of 8 frames & 30k Flickr pairs & \textbf{.167} & 16.5 / 38.8 / 49.1 & 18.6\\
ours, mean of 8 frames & 9k MSR-VTT clips & .167 & 19.3 / 45.6 / 57.9 & 20.5\\
ours, middle frame & 30k Flickr, 200k web pairs & .148 & 14.5 / 33.1 / 44.9 & 17.1\\
CLIP ViT-B/32 \citep{luo2022clip4clip} & -- (jointly trained) & -- & 31.2 / 53.7 / 64.2 & 27.2\\
\bottomrule
\end{tabular}
\end{table}

\paragraph{Alignment without any pair, in detail.} \Cref{tab:pairfree} runs the code of \citet{schnaus2026rosetta} unmodified (Wasserstein Procrustes with $C=30$ clusters, $S=30$ restarts, $b=10{,}000$ points, $R=100$ refinement rounds, the MPOpt solver, their five seeds) on their stored DINOv2 ViT-B/14 mean-token features, with each image's captions averaged into one text embedding, as they do; each side is centered by the mean of its own unpaired training half, and no pair, val2014 image or test label enters the fit (\texttt{evaluate.py pairfree}, \texttt{prh/pairfree.py}). Means $\pm$ std over training seeds times method seeds: FOSCTTM .018 $\pm$ .020 (median .010), CIFAR-10 86.0 $\pm$ 2.3, CIFAR-100 67.2 $\pm$ 4.4 and ImageNet-100 64.4 $\pm$ 3.6 for the 109M good-view JEPA (15 runs); .036 $\pm$ .034, 83.2 $\pm$ 7.4, 60.9 $\pm$ 8.2 and 57.9 $\pm$ 8.8 for the 33M one (15); .018 $\pm$ .002, 83.4 $\pm$ 2.9, 61.7 $\pm$ 1.4 and 47.6 $\pm$ 2.2 for Qwen3-8B with generative pooling (5); .006 $\pm$ .001, 84.0 $\pm$ 5.2, 67.0 $\pm$ 1.4 and 68.7 $\pm$ 1.8 for CLIP's text encoder (5); and .287 $\pm$ .162 and 25.5 $\pm$ 16.0 on CIFAR-10 for next-token prediction (15, half of the runs above FOSCTTM .25). Against Qwen3-8B rerun, the 109M good-view JEPA is level on FOSCTTM ($p=.053$) and CIFAR-10 ($p=.12$) and ahead on CIFAR-100 ($p=.003$) and ImageNet-100 ($p<.001$; Mann--Whitney); against CLIP's text encoder it is level on both CIFARs and on FOSCTTM ($p=.23$) and behind on ImageNet-100 ($p=.008$). These tests treat the 15 runs as independent; at the level of the three training seeds, the separation from Qwen3-8B on CIFAR-100 and ImageNet-100 is complete. Their third pair, Qwen3-Embedding-8B, reproduces their FOSCTTM but not their CIFAR-10 (64 against 74), for reasons we could not trace; their all-mpnet-base-v2 numbers reproduce on all four metrics. Their best FOSCTTM on these data is .02, with DINOv2 ViT-g, and .01 in two of their other settings (COCO images with paragraphs from SPC, and SPC alone); their best zero-shot accuracies over all five settings are 82\%, 62\% and 56\% (\texttt{paper/data/schnaus2026\_reported.csv}). An independent audit recomputed 35 fitted maps from scratch, refitted one bit for bit, and checked that shuffling each text coordinate across the caption pool, which keeps every marginal and destroys the geometry, drops the method to chance (FOSCTTM .57, CIFAR-10 11--15\%). Failures come from the initialization: the 30 cluster centers matched by CKA are mostly wrong (13--24\% correct), and a seed whose averaged initialization points away from the paired map does not recover; the method's own objective on the unpaired pools is anti-correlated with test FOSCTTM (Spearman $-.78$ over 85 COCO runs, $-.80$ over 80 Flickr runs) and flags such seeds. On Flickr (DINOv2 ViT-B CLS features, first captions, disjoint halves of the 30{,}014 non-test pairs), zero-pair R@1 is .164 $\pm$ .104 for the 109M good-view JEPA (15 runs; R@5 .349), .091 $\pm$ .052 for the 33M one (15), and at most .003 for every encoder that is not a good-view JEPA (.074 for the 50M-image good-view JEPA and .012 for the one trained on Flickr captions); given 10 known pairs, the same method reaches R@1 .32 with the good-view JEPA (one seed), against about .01 for Procrustes on those 10 pairs and .067 for CLIP's text encoder. Two disclosures: the first training seed of our 109M good-view JEPA ran on twice as many GPUs as the other two, with the SIGReg weight scaled to keep its effective value; and its positives are texts of the same web image, two of three of which are LLaVA recaptions, so that, although the text encoder never sees a pixel or a COCO pair, its views are grouped by image and partly written by a vision--language model (\cref{tab:sources}).

\paragraph{Decoder noise.} The captioner's training noise is set by matching cosines (\cref{sec:caption}), $\sigma=(c^{-2}-1)^{1/2}$ with $c$ the mean cosine between a rotated image embedding and its caption's embedding. We first measured it on test pairs; held-out validation pairs (a rotation fitted on one half of the 1{,}014 validation pairs, the cosine taken on the other half) give nearly the same noise (column \texttt{cos} of \texttt{evaluate.py align}, with \texttt{--val-only --split 507} for held-out validation pairs). With the rotation fitted on 1{,}014 pairs ($c=.29$ held out, $\sigma\approx3.3$; $c=.33$ on test pairs, $\sigma\approx2.9$), Flickr test CIDEr is .225, .304, .368, .396 and .377 for $\sigma=0.5,1,2,3,4$: lower noise decodes the true caption embedding better but transfers worse, and transfer peaks where predicted. The 30k-pair rotation leaves a smaller residual ($\sigma\approx2.1$ held out, 2.3 on test pairs) and gives .438 at $\sigma=2$ against .431 at $\sigma=3$. Differences below about .01 CIDEr are within the variation of greedy decoding across hardware (floating-point ties), and we do not interpret them. For COCO, the decoder is trained with $\sigma=2$ on the embeddings of a text JEPA trained on Flickr and COCO training captions (Karpathy train and restval splits, disjoint from its test split), and the rotation is fitted on the 30k Flickr pairs.

\paragraph{Scaling the captioner.} The language-model decoder maps the centered, unit-norm text-JEPA embedding to 10 input embeddings of Qwen2.5 through a two-layer MLP (a soft prompt, as in ClipCap, \citealp{mokady2021clipcap}) and is fine-tuned entirely (AdamW, learning rate $3\cdot10^{-5}$, batch 256, gradient clipping 1.0, bf16) on 1.4M draws of Flickr and COCO training captions (Karpathy train and restval), each embedded by the frozen text JEPA with noise $\sigma=1.9$, matched to the cosine $c=.47$ between rotated DINOv2-g embeddings and their captions. The target is the caption as written, at most 64 tokens. Two image encoders are concatenated after rescaling each embedding to unit root-mean-square per coordinate (\cref{tab:image_encoders}); otherwise the encoder with larger features dominates. Consensus decoding (minimum Bayes risk, \citealp{kumar2004mbr}) samples 32 captions besides the greedy one (temperature 0.7, top-$p$ 0.95) and keeps, of these 33, the one most similar to the other 32 under CIDEr's n-gram similarity without document frequencies, which one image's candidates cannot estimate; it uses no reference caption and no other image. With 16 samples it gains little, and 64 samples gain no more than 32 on Flickr. Keeping instead the beam candidate whose text-JEPA embedding is closest to the rotated image embedding does not help. Pretraining the decoder on 4M first sentences of Recap recaptions before the captions lets it decode true caption embeddings better (.743 and 1.369) but transfers no better on Flickr and COCO; on photos of other kinds its captions are more often right (coats of arms, a cake shaped like a sheep), and the demo uses it. Two other changes do not help: a higher input resolution raises retrieval (R@1 .440 to .462 with DINOv2-g at 448 px) but not the cosine the decoder sees, and a per-direction gain after the rotation, fitted on the same pairs, at best raises Flickr CIDEr by .017 and lowers COCO's by .053 (web-pretrained decoder, beam 3).

\paragraph{Projector outputs.} On projector outputs, where SIGReg acts, whitening each side before the rotation barely changes retrieval (R@1 .030 vs.\ .034), as expected once the embedding is nearly isotropic.

\paragraph{Transfer to COCO.} The rotation fitted on the 30k Flickr pairs (text JEPA trained on Flickr and COCO captions, BERT-S) ranks the true caption first for 11.3\% of the 5{,}000 COCO Karpathy test images (chance 0.02\%; mutual 10-NN .185), with each side centered by its own unpaired COCO mean; a rotation fitted on 2{,}500 COCO pairs and tested on the other 2{,}500 reaches 15.8\% (chance 0.04\%).

\paragraph{Frozen image encoders.} \Cref{tab:image_encoders} pairs our two strongest text JEPAs, BERT-B on the texts of 50M images and BERT-S on those of 100M, with other frozen self-supervised image encoders through the same 30k-pair rotation. Within DINOv2, larger encoders align better, and so do linear-evaluation features, the CLS token concatenated with the mean patch token. With DINOv3 ViT-7B, the true caption is retrieved first for 53.1\% of Flickr test images, through the rotation alone. Retrieval through the rotation and agreement of neighborhoods, which fits nothing, need not rank encoders alike: DINOv2 ViT-g has the most similar neighborhoods of single encoders and also gives their best captions (CIDEr .480 on Flickr and .756 on COCO with its CLS and mean patch token, against .468 and .746 for DINOv3 ViT-L and .448 and .652 for DINOv3 ViT-7B; Qwen2.5-1.5B decoder, beam 3). Concatenating two encoders helps both, and DINOv2 ViT-g with DINOv3 ViT-L gives the best captions for its cost (\cref{tab:captioner_scale}). I-JEPA, whose positives are blocks of one image and which saw only ImageNet-22k, aligns far less.

\begin{table}[h]
\centering
\small
\caption{\textbf{Linear-evaluation features and concatenated encoders align best.} Flickr test pairs, text JEPAs trained on Recap texts only (seed 0), each against a frozen self-supervised image encoder through one map with orthonormal rows fitted on the 30k Flickr train and validation pairs (chance: mutual 10-NN .01, R@1 .001). Bottom rows: the CLS + mean patch features of two encoders, each rescaled to unit root-mean-square per coordinate, concatenated. No network is trained on pairs. Notably, \emph{with DINOv3 ViT-7B, one rotation retrieves the true caption first for more than half of the test images}.}
\label{tab:image_encoders}
\begin{tabular}{llccccc}
\toprule
& & \multicolumn{3}{c}{BERT-B, texts of 50M images} & \multicolumn{2}{c}{BERT-S, 100M}\\
Frozen image encoder & features & mutual 10-NN & R@1 & R@5 & R@1 & R@5\\
\midrule
DINOv2 ViT-B/14 & CLS & .234 & .412 & .733 & .400 & .711\\
DINOv2 ViT-L/14 & CLS & .242 & .437 & .745 & .410 & .732\\
DINOv2 ViT-g/14 & CLS & .245 & .437 & .769 & .425 & .741\\
DINOv2 ViT-g/14 & CLS + mean patch & .250 & .460 & .784 & .440 & .763\\
DINOv3 ViT-L/16 & CLS & .209 & .436 & .722 & .414 & .703\\
DINOv3 ViT-L/16 & CLS + mean patch & .234 & .514 & .792 & .489 & .780\\
DINOv3 ViT-H+/16 & CLS & .181 & .353 & .625 & .342 & .607\\
DINOv3 ViT-H+/16 & CLS + mean patch & .228 & .479 & .779 & .486 & .775\\
DINOv3 ViT-7B/16 & CLS & .207 & .455 & .752 & .452 & .730\\
DINOv3 ViT-7B/16 & CLS + mean patch & .231 & .531 & .803 & \textbf{.523} & .798\\
I-JEPA ViT-H/14 (IN-22k) & mean patch & .109 & .188 & .451 & .183 & .436\\
I-JEPA ViT-g/16 (IN-22k) & mean patch & .140 & .246 & .502 & .234 & .491\\
\midrule
DINOv2-g + DINOv3-L & both, concatenated & .255 & .505 & .806 & .488 & .791\\
DINOv2-g + DINOv3-7B & both, concatenated & \textbf{.256} & \textbf{.535} & \textbf{.824} & .511 & \textbf{.801}\\
\bottomrule
\end{tabular}
\end{table}

\begin{table}[h]
\centering
\small
\caption{\textbf{Good views help on both sides (numbers behind \cref{fig:both_sides}).} MSR-VTT, 1{,}000 test clips, final checkpoints, mean $\pm$ half-range over three seeds (chance: mutual 10-NN .01, R@1 .001). R@1 and R@5 use one rotation fitted on the 1{,}990 validation clips. Every seed ranks the four combinations the same way in mutual 10-NN; the image-side effect is small and reverses on R@5.}
\label{tab:msrvtt}
\begin{tabular}{llccc}
\toprule
Image JEPA & Text JEPA & mutual 10-NN & R@1 & R@5\\
\midrule
good views (frames) & good views (captions) & \textbf{.067}{\scriptsize$\pm$.001} & \textbf{.032}{\scriptsize$\pm$.006} & .099{\scriptsize$\pm$.009}\\
bad views (crops) & good views (captions) & .059{\scriptsize$\pm$.002} & .024{\scriptsize$\pm$.002} & .105{\scriptsize$\pm$.008}\\
good views (frames) & bad views (spans) & .043{\scriptsize$\pm$.002} & .020{\scriptsize$\pm$.005} & .063{\scriptsize$\pm$.002}\\
bad views (crops) & bad views (spans) & .039{\scriptsize$\pm$.003} & .017{\scriptsize$\pm$.002} & .072{\scriptsize$\pm$.009}\\
\midrule
frozen DINOv2 ViT-B & good views (captions) & .129{\scriptsize$\pm$.002} & .085{\scriptsize$\pm$.006} & .258{\scriptsize$\pm$.011}\\
frozen DINOv2 ViT-B & bad views (spans) & .084{\scriptsize$\pm$.002} & .052{\scriptsize$\pm$.002} & .164{\scriptsize$\pm$.006}\\
frozen DINOv2 ViT-B & image JEPA, good views & .132{\scriptsize$\pm$.002} & .219{\scriptsize$\pm$.010} & .462{\scriptsize$\pm$.011}\\
frozen DINOv2 ViT-B & image JEPA, bad views & .144{\scriptsize$\pm$.001} & .285{\scriptsize$\pm$.005} & .559{\scriptsize$\pm$.008}\\
\bottomrule
\end{tabular}
\end{table}

\paragraph{Text-only quality.} Good views also make better text encoders on their own. On the STS-Benchmark test set \citep{cer2017semeval} (Spearman correlation between the cosine similarity of two sentences' embeddings and human similarity, no fine-tuning), the good-view JEPA, the bad-view JEPA and next-token prediction score .579 / .528 / .430 when trained on Flickr captions and .612 / .501 / .377 when trained on 2M Recap images, and the 50M-image good-view JEPA reaches .632, above BERT-base (.551) and Qwen2.5-0.5B (.352, mean-pooled) though below all-mpnet-base-v2 (.834), which is trained on a billion supervised sentence pairs. The ordering good $>$ bad $>$ next-token is the same as for alignment on Flickr (on WIT, next-token prediction is above the bad-view JEPA).

\paragraph{Uncertainty on WIT.} WIT-1024 is small. We compute each score on 200 random 800-pair subsets drawn without replacement (the same subsets for every model; each model at its best layers on all pairs; DINOv2 ViT-B) and turn their spread into a 95\% interval for the 1{,}024-pair score: for average-like statistics, the spread of 800-pair subsets around the full score is $\sqrt{224/800}$ times the full score's standard error, so we rescale it by $\sqrt{800/224}$ (\texttt{prh.metrics.subsample\_interval}). Against language models our text JEPAs are level: the 50M-image JEPA minus Qwen2.5-7B is $+.010$ ($[-.005, +.025]$) and minus Qwen2.5-3B $+.003$ ($[-.011, +.018]$), the 100M-image JEPA minus Mistral-7B $-.011$ ($[-.026, +.004]$), and the 2M-image JEPA minus Qwen2.5-0.5B $+.005$ ($[-.008, +.018]$). Between views the differences are clear: on 2M images, the good-view JEPA minus next-token prediction is $+.037$ ($[+.025, +.049]$) and minus the bad-view JEPA $+.054$ ($[+.041, +.067]$); on 50M images, $+.050$ ($[+.037, +.062]$) and $+.063$ ($[+.050, +.075]$). Training seeds matter as much as test sampling: the three seeds of the 50M-image JEPA score .129, .118 and .127 with DINOv2 ViT-B (seed 0 is used above), so the comparisons with language models are within noise, while the gaps between views, at least $+.040$ over next-token prediction and $+.056$ over the bad-view JEPA for every seed, are far outside it.

\paragraph{Image JEPAs trained from scratch.} Behind \cref{fig:scaling} (right): a ViT-S on the 2.16M Recap images (10 epochs), a ViT-B on 10M Recap images (8 epochs) and a ViT-S on ImageNet-1k (300 epochs), all trained without texts. The ImageNet ViT-S, which sees the most samples, agrees most with DINOv2 (.407, against .315 and .317) and with the 50M-image text JEPA (.095 over three text seeds, against .079 and .080), and its alignment with the 109M good-view JEPA reaches .130 (\cref{tab:wit}); two more ImageNet ViT-S trained from other seeds reproduce it (.414 and .405 with DINOv2; with the good-view, next-token and bad-view text JEPAs .098, .061 and .055, and .100, .062 and .053, against .098, .062 and .055). With it, two JEPAs trained separately from scratch agree as much as the image JEPA agrees with the strongest language models: for seed 0, our text JEPA minus Mistral-7B is $-.003$ ($[-.016, +.011]$) and minus Qwen2.5-7B $+.011$ ($[-.001, +.024]$), while it beats next-token prediction on the same texts by $+.037$ ($[+.026, +.048]$; intervals as in the previous paragraph).

\paragraph{Scaling to the texts of 940M images.} \Cref{tab:bigscale} follows our largest good-view JEPAs along training. With the same BERT-B encoder, the texts of 200M images and a global batch of 256 reach .125--.126 on WIT after 89M--160M samples (the end of training), while the texts of all 940M images and a global batch of 2{,}048 reach .182 after 430M samples and .191 after 850M and at the end of training (900M). These runs change unique texts, samples seen and batch size together, and we do not separate their effects. The SIGReg weight also differs: in stable-pretraining as we ran it, the Epps--Pulley statistic is computed on each GPU's batch and scaled by its size, so its effective weight grows with the per-GPU batch, and the first 940M run (16 GPUs of 128) trains with the equivalent of $\lambda=.01$ on a single GPU, half the weight of our single-GPU runs, as do the 200M-image BERT-B (2 GPUs of 128) and the ImageNet and ViT-B image JEPAs (8 GPUs of 128). A second seed on 8 GPUs of 256, i.e., at $\lambda=.02$, reaches .121 on WIT and .477 Flickr R@1: at this scale WIT is sensitive to the SIGReg weight, which our ablations on Flickr captions did not reveal. We chose the weight of the first run for every other 940M run after seeing this result, so that weight was selected on WIT: at $\lambda=.01$ on 8 GPUs of 256, the effective weight of the first run, two more seeds reach .181 and .179 on WIT and .484 and .485 Flickr R@1 (three seeds: .184 and .478), the bad-view control .088 and .318, and three next-token seeds .109, .111 and .112 on WIT and .346, .332 and .328 Flickr R@1. Across DINOv2 sizes, the three good-view seeds average .166--.197 (\cref{tab:wit}), the bad-view control .082--.097 and next-token prediction .101--.122. Averaged over all four good-view seeds, WIT is .168 with DINOv2 ViT-B, above every language model, and .198, .202 and .212 with CLIP ViT-B, -L and -H, level with the best language model (.193, .203 and .210).

\begin{table}[h]
\centering
\small
\caption{\textbf{Alignment keeps growing with the texts of 940M images.} Checkpoints of our largest good-view JEPAs (seed 0 unless named; $^\dagger$: mean of three seeds at the end of training; $^\ddagger$: another seed with twice the effective SIGReg weight, see text; next-token: the same encoder trained to predict the next token; bad views: two masked copies of one text, at the effective SIGReg weight of the first 940M run), scored on WIT-1024 with DINOv2 ViT-B and on Flickr through one rotation fitted on 30k pairs. We observe that \emph{WIT alignment jumps between the texts of 200M and 940M images and rises again by 850M samples}, while Flickr retrieval rises more slowly. Bold: best per column.}
\label{tab:bigscale}
\begin{tabular}{lccccc}
\toprule
Text JEPA & texts of & global batch & samples seen & WIT (DINOv2-B) & Flickr R@1 \\
\midrule
BERT-S & 100M images & 256 & 80M & .125$^\dagger$ & .402$^\dagger$\\
BERT-B & 200M images & 256 & 89M & .125 & .447\\
BERT-B & 200M images & 256 & 117M & .126 & .447\\
BERT-B & 200M images & 256 & 160M & .126 & .451\\
BERT-B & 940M images & 2{,}048 & 430M & .182 & .410\\
BERT-B & 940M images & 2{,}048 & 580M & .182 & .463\\
BERT-B & 940M images & 2{,}048 & 850M & \textbf{.191} & .468\\
BERT-B & 940M images & 2{,}048 & 900M & \textbf{.191} & .464\\
BERT-B, seed 1 & 940M images & 2{,}048 & 900M & .181 & .484\\
BERT-B, seed 2 & 940M images & 2{,}048 & 900M & .179 & \textbf{.485}\\
BERT-B$^\ddagger$ & 940M images & 2{,}048 & 900M & .121 & .477\\
BERT-B, next-token, seed 0 & 940M images & 2{,}048 & 900M & .109 & .346\\
BERT-B, next-token, seed 1 & 940M images & 2{,}048 & 900M & .111 & .332\\
BERT-B, next-token, seed 2 & 940M images & 2{,}048 & 900M & .112 & .328\\
BERT-B, bad views & 940M images & 2{,}048 & 900M & .088 & .318\\
\bottomrule
\end{tabular}
\end{table}

\paragraph{WIT captions in the training texts.} DataComp is crawled from the web, so some WIT captions may appear among the Recap-DataComp alt-texts. We lower-case each WIT caption and each of the 940M alt-texts, keep letters, digits and spaces, and match them exactly or by their first 40 characters: 26 WIT captions match exactly and 17 more by prefix, at least 43 of 1{,}024 in all (captions shorter than 20 characters are not checked; 16 within the texts of the first 100M images, 20 within the first 200M); 3.3M of the 940M images are hosted on Wikimedia. Removing the 43 pairs changes none of the nine text models we checked with DINOv2 ViT-B by more than .002 (\cref{sec:scaling}), and the 109M model scores .193 on the 981 remaining pairs. The WIT-1024 release does not identify its images, so we cannot match them to DataComp images. Flickr30k images can be: their file names are Flickr photo ids, which Flickr-hosted DataComp URLs contain, and at least 178 of the 31{,}014 Flickr30k photos appear among the 940M Recap-DataComp images (166 train, 6 validation and 6 of the 1{,}000 test photos, each with its web alt-text and recaptions, never its Flickr captions), so these can move Flickr test retrieval by at most .006 (\texttt{python -m prh.data overlap}). COCO photos, also hosted on Flickr, match the same way: at least 1{,}207 COCO 2014 photos are among the 940M images, 393 of them from val2014, where \cref{tab:pairfree} scores, and 50 of the 5{,}000 COCO test photos of \cref{tab:coco}, again with their web texts and never their COCO captions; removing these 50 changes the COCO score of the 109M JEPA, and of every baseline we checked, by at most .001 (\texttt{evaluate.py wit --bench coco --drop}).

\paragraph{WIT, PRH protocol.} \Cref{tab:wit} applies the exact protocol of \citet{huh2024platonic} on their WIT-1024 pairs, which none of our models is trained on (DataComp images may overlap with Wikipedia's). With DINOv2, pretrained language models reach .09--.16; alignment mostly grows with the size of the vision model and, for Qwen2.5, with the language model up to 3B, the trend reported by \citet{huh2024platonic}. Our Flickr-trained text encoders, trained on 145k short captions, do not transfer to Wikipedia captions (.029--.037 for every objective, three seeds each, including next-token prediction on caption pairs, and every DINOv2 size; best text layer among the first few, mostly the embeddings). Trained on the texts of 2.16M Recap-DataComp images instead, the same recipe reaches .099--.114 depending on the DINOv2 size (means of three seeds), with its last layer as the best, level with Qwen2.5-0.5B.

\begin{table}[h]
\centering
\small
\setlength{\tabcolsep}{4pt}
\caption{\textbf{A 109M text JEPA beats language models on the PRH benchmark.} PRH protocol of \citet{huh2024platonic}: WIT-1024 pairs, per-layer features, clipping, normalization, maximum over layer pairs. For reference, DINOv2 ViT-B $\times$ BERT-base scores .111, all-mpnet-base-v2 .118 and CLIP's text encoder .124 (.113--.137 across DINOv2 sizes for these two); our Flickr-trained text encoders (good / bad / next-token) score .029--.036 with every DINOv2 size and .023--.024 with our Flickr image JEPA; and our Flickr image JEPA $\times$ DINOv2 ViT-B (image vs.\ image) .241. nt: next-token prediction. Our 33M text JEPAs' entries are means of three seeds, which differ by at most $\pm$.009 with DINOv2 ViT-B; the 109M column is the mean of three seeds (each above every language model). At least 43 of the 1{,}024 WIT captions appear among the Recap-DataComp alt-texts; on the other 981 pairs, none of the nine text models we checked with DINOv2 ViT-B (ours, Qwen2.5-3B/7B, Mistral-7B) changes by more than .002. This demonstrates that \emph{the 109M good-view JEPA surpasses every language model we evaluated with every vision encoder it was scored with}, including our image JEPAs trained from scratch.}
\label{tab:wit}
\begin{tabular}{lcccccccccc}
\toprule
& \multicolumn{4}{c}{Qwen2.5} & Mistral & \multicolumn{4}{c}{ours (33M), on the texts of Recap images} & ours (109M)\\
& & & & & & 2.16M & 10M & 50M & 100M & 940M\\
Vision & 0.5B & 1.5B & 3B & 7B & 7B & good / bad / nt & good & good / bad / nt & good & good\\
\midrule
DINOv2 ViT-S/14 & .092 & .101 & .110 & .110 & .126 & .099 / .052 / .065 & .102 & .114 / .062 / .074 & .114 & \textbf{.166}\\
DINOv2 ViT-B/14 & .102 & .108 & .125 & .118 & .140 & .108 / .053 / .071 & .111 & .125 / .063 / .079 & .125 & \textbf{.184}\\
DINOv2 ViT-L/14 & .114 & .123 & .139 & .137 & .157 & .111 / .057 / .075 & .115 & .130 / .068 / .085 & .131 & \textbf{.196}\\
DINOv2 ViT-g/14 & .115 & .124 & .141 & .133 & .157 & .114 / .059 / .076 & .117 & .132 / .070 / .086 & .133 & \textbf{.197}\\
\midrule
ours, ViT-S, Recap 2M & .062 & .068 & .078 & .073 & .086 & .070 / .042 / .053 & .071 & .079 / .048 / .057 & .080 & \textbf{.106}\\
ours, ViT-B, Recap 10M & .064 & .070 & .079 & .075 & .087 & .071 / .043 / .054 & .072 & .080 / .048 / .058 & .081 & \textbf{.106}\\
ours, ViT-S, ImageNet & .074 & .078 & .091 & .087 & .101 & .082 / .046 / .057 & .085 & .095 / .053 / .062 & .094 & \textbf{.130}\\
ours, ViT-S, Flickr & .050 & .051 & .057 & .055 & .063 & .055 / .034 / .043 & .055 & .060 / .038 / .047 & .060 & \textbf{.078}\\
\bottomrule
\end{tabular}
\end{table}

\paragraph{A second alignment metric.} Mutual 10-NN only counts shared neighbors. We also score every layer pair with linear CKA and the unbiased HSIC estimator \citep{song2012feature}, as in the released code of \citet{huh2024platonic}, with the same features, preprocessing and maximum over layer pairs (\texttt{prh.metrics.unbiased\_cka}). With DINOv2 ViT-B, the 109M good-view JEPA reaches .507 (mean of three seeds, .495--.518), against .390 for all-mpnet-base-v2, .363 for Qwen2.5-3B, .353 for Qwen2.5-7B, .349 for Mistral-7B and .339 for CLIP's text encoder; on the texts of 50M images, the good, bad-view and next-token JEPAs score .385, .192 and .261 (seed 0), and on those of 2.16M images .301, .185 and .225. Across the 16 text models we scored, the two metrics rank them alike (Spearman .88 with DINOv2 ViT-B and .86 with ViT-L). That is, \emph{the ranking of views and the lead of the 109M good-view JEPA do not depend on the alignment metric}.

\paragraph{Larger language models.} \Cref{tab:llms} extends the comparison to Qwen2.5 up to 72B parameters and Llama-3 at 8B and 70B (base models, the same per-layer protocol; the 14B and larger ones split across two GPUs in bfloat16). Alignment grows little with size beyond a few billion parameters--Qwen2.5 reaches .118--.130 from 7B to 32B with DINOv2 ViT-B, and .154 at 72B--and no language model reaches the 109M good-view JEPA with any DINOv2 size, either metric or either benchmark. Llama-3-8B is the language model of the image captioner that wrote Recap-DataComp's recaptions; it scores .145.

\begin{table}[h]
\centering
\small
\setlength{\tabcolsep}{4pt}
\caption{\textbf{No language model up to 72B parameters reaches the 109M good-view JEPA.} PRH protocol (per-layer features, maximum over layer pairs) on WIT-1024 and on the 5{,}000 COCO test pairs, for base language models from 0.5B to 72B parameters and our 109M good-view text JEPA trained on the texts of 940M images (mean of three seeds). We clearly observe that \emph{scaling next-token prediction by 660 times in parameters does not reach the alignment that good views give}.}
\label{tab:llms}
\begin{tabular}{lccccccc}
\toprule
& \multicolumn{5}{c}{WIT-1024} & \multicolumn{2}{c}{COCO-5k}\\
& \multicolumn{4}{c}{mutual 10-NN, DINOv2 S / B / L / g} & CKA, B & 10-NN, B & CKA, B\\
\midrule
Qwen2.5-0.5B & .092 & .102 & .114 & .115 & .318 & .100 & .571\\
Qwen2.5-1.5B & .101 & .108 & .123 & .124 & .324 & .089 & .538\\
Qwen2.5-3B & .110 & .125 & .139 & .141 & .363 & .099 & .550\\
Qwen2.5-7B & .110 & .118 & .137 & .133 & .353 & .101 & .535\\
Qwen2.5-14B & .119 & .130 & .147 & .148 & .333 & .106 & .539\\
Qwen2.5-32B & .116 & .128 & .145 & .144 & .336 & .104 & .533\\
Qwen2.5-72B & .138 & .154 & .166 & .169 & .418 & .108 & .534\\
Mistral-7B & .126 & .140 & .157 & .157 & .349 & .112 & .539\\
Llama-3-8B & .130 & .145 & .165 & .164 & .377 & .109 & .599\\
Llama-3-70B & .136 & .151 & .166 & .166 & .380 & .113 & .587\\
ours: good-view text JEPA, 109M & \textbf{.166} & \textbf{.184} & \textbf{.196} & \textbf{.197} & \textbf{.507} & \textbf{.161} & \textbf{.622}\\
\bottomrule
\end{tabular}
\end{table}

\paragraph{A second benchmark.} WIT-1024 is small, so we repeat the protocol of \citet{huh2024platonic} (per-layer features, clipping, normalization, maximum over layer pairs) on the 5{,}000 COCO Karpathy test images and their first captions (\texttt{evaluate.py wit-embed --bench coco}), whose captions no text encoder of ours is trained on (\cref{tab:coco}; 50 of its photos are among the 940M web images, with their web texts, and removing them moves every text encoder we checked by at most .001). The ranking is that of WIT: the 109M good-view JEPA leads every text encoder with both DINOv2 sizes and both metrics, including all-mpnet-base-v2, whose billion training pairs include COCO captions, and CLIP's text encoder, and at fixed data the good-view JEPA is ahead of next-token prediction and the bad-view JEPA.

\begin{table}[h]
\centering
\small
\caption{\textbf{The ranking holds on COCO.} PRH protocol on the 5{,}000 COCO Karpathy test pairs (first caption), seed 0 for our encoders except the 940M row, the mean of three seeds (each at or above all-mpnet-base-v2); chance mutual 10-NN is .002. We clearly observe that \emph{the 109M good-view JEPA aligns best on a second, five times larger benchmark, ahead of language models up to 72B parameters and of text encoders trained on pairs}.}
\label{tab:coco}
\begin{tabular}{lcccc}
\toprule
& \multicolumn{2}{c}{mutual 10-NN} & \multicolumn{2}{c}{unbiased CKA}\\
Text encoder (COCO-5k) & DINOv2-B & DINOv2-g & DINOv2-B & DINOv2-g\\
\midrule
ours, good, BERT-B, 940M images & \textbf{.161} & \textbf{.169} & \textbf{.622} & \textbf{.667}\\
ours, good, 100M images & .134 & .140 & .587 & .598\\
ours, good, 50M images & .126 & .135 & .578 & .576\\
ours, bad, 50M images & .085 & .090 & .447 & .423\\
ours, next-token, 50M images & .091 & .096 & .522 & .487\\
Qwen2.5-0.5B & .100 & .105 & .571 & .529\\
Qwen2.5-1.5B & .089 & .095 & .538 & .505\\
Qwen2.5-3B & .099 & .104 & .550 & .525\\
Qwen2.5-7B & .101 & .107 & .535 & .510\\
Qwen2.5-14B & .106 & .112 & .539 & .526\\
Qwen2.5-32B & .104 & .109 & .533 & .521\\
Qwen2.5-72B & .108 & .113 & .534 & .521\\
Mistral-7B & .112 & .118 & .539 & .526\\
Llama-3-8B & .109 & .115 & .599 & .560\\
Llama-3-70B & .113 & .118 & .587 & .529\\
BERT-base & .104 & .109 & .485 & .475\\
all-mpnet-base-v2 & .156 & .162 & .608 & .577\\
CLIP text encoder & .132 & .138 & .576 & .565\\
\bottomrule
\end{tabular}
\end{table}

\paragraph{Other vision families.} \Cref{tab:vision_families} repeats the WIT protocol with frozen vision encoders trained by pixel reconstruction (MAE, \citealp{he2022masked}) and by latent prediction from image blocks (I-JEPA, \citealp{assran2023ijepa}; its per-layer feature is the mean token, as it has no CLS token). The ranking of the text side does not depend on the vision encoder: with each of the six, the good-view text JEPA beats next-token prediction and the bad-view JEPA trained on the same texts, and the 109M good-view JEPA beats every language model.

\begin{table}[h]
\centering
\small
\setlength{\tabcolsep}{4pt}
\caption{\textbf{Good views win with every vision family.} WIT-1024 (PRH protocol) for frozen vision encoders trained by pixel reconstruction (MAE), latent prediction (I-JEPA), crops (DINOv2) or image--text contrast (CLIP's image encoders, the vision models with which language models align best in \citet{huh2024platonic}), against our text JEPAs (good / bad / next-token on the texts of 50M images, seed 0; the 109M good-view JEPA on the texts of 940M images, seed 0) and pretrained language models. We clearly observe that \emph{good views beat next-token prediction and bad views with every vision encoder, and that the 109M good-view JEPA beats the three language models with each of them.} MAE and I-JEPA align alike, both below DINOv2, which saw far more curated images; we draw no conclusion on the vision objective. With CLIP's image encoders, the 109M good-view JEPA's three seeds average .215, .219 and .230, against .193, .203 and .210 for the best of all our language models up to 72B (Qwen2.5-72B), and its next-token and bad-view controls on the same texts .137--.146 and .108--.116 (\texttt{runs/results/wit\_clip.txt}).}
\label{tab:vision_families}
\begin{tabular}{lccccccc}
\toprule
& \multicolumn{3}{c}{ours (33M), texts of 50M images} & ours (109M), 940M & Qwen2.5 & Qwen2.5 & Mistral\\
Vision encoder & good & bad & next-token & good & 0.5B & 7B & 7B\\
\midrule
MAE ViT-B/16 (IN-1k) & .074 & .045 & .053 & \textbf{.100} & .060 & .067 & .077\\
MAE ViT-L/16 (IN-1k) & .093 & .050 & .059 & \textbf{.126} & .074 & .084 & .097\\
MAE ViT-H/14 (IN-1k) & .092 & .050 & .058 & \textbf{.125} & .071 & .084 & .096\\
I-JEPA ViT-H/14 (IN-22k) & .095 & .051 & .059 & \textbf{.129} & .068 & .081 & .096\\
DINOv2 ViT-B/14 & .129 & .066 & .079 & \textbf{.191} & .102 & .118 & .140\\
DINOv2 ViT-L/14 & .133 & .069 & .084 & \textbf{.199} & .114 & .137 & .157\\
\midrule
CLIP ViT-B/16 (LAION-2B) & .147 & .076 & .093 & \textbf{.221} & .135 & .156 & .180\\
CLIP ViT-L/14 (LAION-2B) & .148 & .078 & .098 & \textbf{.224} & .144 & .164 & .194\\
CLIP ViT-H/14 (LAION-2B) & .156 & .080 & .099 & \textbf{.233} & .148 & .169 & .196\\
\bottomrule
\end{tabular}
\end{table}

\end{document}

%% file: fig_positives.tex
% What counts as a positive pair for each objective, and which observations its positives connect (Thm. 4).
% Each panel: an example pair (top), then eight captions of two scenes (dots in two ellipses) joined as the
% objective's positives join them, then the reading of Thm. 4.
\begin{tikzpicture}[
  font=\footnotesize,
  ex/.style={draw=black!25, fill=black!3, rounded corners=1.5pt, text width=40mm, align=center, inner sep=2pt,
             minimum height=13mm, font=\footnotesize},
  obs/.style={circle, fill=black!65, inner sep=0pt, minimum size=3.2pt},
  scene/.style={draw=black!25, fill=black!4, ellipse, minimum width=15mm, minimum height=11mm},
  same/.style={aligned, line width=0.7pt},
  cross/.style={bad, line width=0.7pt},
  loop/.style={ntp, line width=0.6pt},
  weak/.style={aligned, line width=0.5pt, densely dotted},
  read/.style={text width=41mm, align=center, font=\footnotesize, black!80, anchor=north,
              execute at begin node={\hyphenpenalty=10000\exhyphenpenalty=10000}},
  head/.style={font=\small\bfseries},
]
% Two scenes with four captions each; A1..A4 and B1..B4 placed identically in every panel.
\newcommand{\scenes}{%
  \node[scene] (SA) at (0.95, 0) {};
  \node[scene] (SB) at (3.05, 0) {};
  \node[font=\footnotesize, black!55, above=-1pt of SA] {scene 1};
  \node[font=\footnotesize, black!55, above=-1pt of SB] {scene 2};
  \node[obs] (A1) at (0.62, 0.20) {}; \node[obs] (A2) at (1.22, 0.25) {};
  \node[obs] (A3) at (0.70, -0.24) {}; \node[obs] (A4) at (1.28, -0.20) {};
  \node[obs] (B1) at (2.72, 0.20) {}; \node[obs] (B2) at (3.32, 0.25) {};
  \node[obs] (B3) at (2.80, -0.24) {}; \node[obs] (B4) at (3.38, -0.20) {};
}
\newcommand{\selfloops}{%
  \foreach \n in {A1,A2,A3,A4,B1,B2,B3,B4}{\draw[loop] (\n) to[out=60, in=120, looseness=14] (\n);}
}
% ---------------- (a) JEPA, good views ----------------
\begin{scope}[xshift=0cm]
  \node[head] at (2.0, 2.6) {(a) Good views};
  \node[ex] at (2.0, 1.6) {\textit{a dog runs on the beach} $\leftrightarrow$ \textit{a puppy plays by the sea}\\two captions of one image};
  \scenes
  \foreach \a/\b in {A1/A2,A1/A3,A1/A4,A2/A3,A2/A4,A3/A4,B1/B2,B1/B3,B1/B4,B2/B3,B2/B4,B3/B4}{\draw[same] (\a) -- (\b);}
  \node[read] at (2.0, -0.68) {every description of a scene, never another: the world's own kernel, by design};
\end{scope}
% ---------------- (b) JEPA, weak bad views ----------------
\begin{scope}[xshift=4.35cm]
  \node[head] at (2.0, 2.6) {(b) Bad views, weak};
  \node[ex] at (2.0, 1.6) {\textit{a dog [M] on the beach} $\leftrightarrow$ \textit{a dog runs [M] the beach}\\two maskings of one caption};
  \scenes
  \selfloops
  \draw[weak] (A1) -- (A3); \draw[weak] (B2) -- (B4);
  \node[read] at (2.0, -0.68) {a caption with its own wording (its nuisance); others only through overlapping corruptions};
\end{scope}
% ---------------- (c) JEPA, strong bad views ----------------
\begin{scope}[xshift=8.7cm]
  \node[head] at (2.0, 2.6) {(c) Bad views, strong};
  \node[ex] at (2.0, 1.6) {\textit{a dog runs on the beach} $\to$ \textit{a [M] [M] on [M] [M]} $\leftarrow$ \textit{a bus parked on a street}\\one masking, two scenes};
  \scenes
  \foreach \a/\b in {A1/A2,A1/A3,A2/A4,A3/A4,B1/B2,B1/B3,B2/B4,B3/B4}{\draw[same] (\a) -- (\b);}
  \foreach \a/\b in {A2/B1,A4/B3,A2/B3,A4/B1}{\draw[cross] (\a) -- (\b);}
  \node[read] at (2.0, -0.68) {corruptions shared across scenes: a kernel that mixes scenes at any data size};
\end{scope}
% ---------------- (d) next-token prediction ----------------
\begin{scope}[xshift=13.05cm]
  \node[head] at (2.0, 2.6) {(d) Next-token prediction};
  \node[ex] at (2.0, 1.6) {\textit{a dog digs in the} $\to$ \textit{sand} $\leftarrow$ \textit{a puppy rolls in the}\\a prefix and its continuation; contexts with one next token};
  \scenes
  \selfloops
  \draw[weak] (A1) -- (A2); \draw[weak] (A3) -- (A4); \draw[weak] (B1) -- (B3);
  \draw[cross, densely dashed, line width=0.5pt] (A4) -- (B3);
  \node[read] at (2.0, -0.68) {its own document (nuisance), contexts sharing one token (narrow), frequent tokens (spill)};
\end{scope}
% legend, one entry under each panel
\begin{scope}[yshift=-2.35cm]
  \draw[same] (0.25, 0) -- (0.7, 0); \node[anchor=west] at (0.75, 0) {within a scene};
  \draw[weak] (4.6, 0) -- (5.05, 0); \node[anchor=west] at (5.1, 0) {sparse: dense data needed};
  \draw[cross] (8.95, 0) -- (9.4, 0); \node[anchor=west] at (9.45, 0) {across scenes};
  \node[obs] (L) at (13.45, -0.06) {}; \draw[loop] (L) to[out=60, in=120, looseness=14] (L);
  \node[anchor=west] at (13.6, 0) {a caption with itself};
\end{scope}
\end{tikzpicture}